# Combining Spatial and Temporal Logics: Expressiveness vs. Complexity

**David Gabelaia** GABELAIA@DCS.KCL.AC.UK
**Roman Kontchakov** ROMANVK@DCS.KCL.AC.UK
**Agi Kurucz** KUAG@DCS.KCL.AC.UK
*Department of Computer Science, King's College London*
*Strand, London WC2R 2LS, U.K.*

**Frank Wolter** FRANK@CSC.LIV.AC.UK
*Department of Computer Science, University of Liverpool*
*Liverpool L69 7ZF, U.K.*

**Michael Zakharyaschev** MZ@DCS.KCL.AC.UK
*Department of Computer Science, King's College London*
*Strand, London WC2R 2LS, U.K.*

## Abstract

In this paper, we construct and investigate a hierarchy of spatio-temporal formalisms that result from various combinations of propositional spatial and temporal logics such as the propositional temporal logic $\mathcal{PTL}$, the spatial logics $\mathcal{RCC}$-8, $\mathcal{BRCC}$-8, $\mathcal{S}4_u$ and their fragments. The obtained results give a clear picture of the trade-off between expressiveness and 'computational realisability' within the hierarchy. We demonstrate how different combining principles as well as spatial and temporal primitives can produce NP-, PSPACE-, EXPSPACE-, 2EXPSPACE-complete, and even undecidable spatio-temporal logics out of components that are at most NP- or PSPACE-complete.

## 1. Introduction

Qualitative representation and reasoning has been quite successful in dealing with both time and space. There exists a wide spectrum of temporal logics (see, e.g., Allen, 1983; Clarke & Emerson, 1981; Manna & Pnueli, 1992; Gabbay, Hodkinson, & Reynolds, 1994; van Benthem, 1995). There is a variety of spatial formalisms (e.g., Clarke, 1981; Egenhofer & Franzosa, 1991; Randell, Cui, & Cohn, 1992; Asher & Vieu, 1995; Lemon & Pratt, 1998). In both cases determining the computational complexity of the respective reasoning problems has been one of the most important research issues. For example, Renz and Nebel (1999) analysed the complexity of $\mathcal{RCC}$-8, a fragment of the *region connection calculus* $\mathcal{RCC}$ with eight jointly exhaustive and pairwise disjoint base relations between spatial regions introduced by Egenhofer and Franzosa (1991) and Randell and his colleagues (1992); Nebel and Bürckert (1995) investigated the complexity of Allen's interval algebra; numerous results on the computational complexity of the point-based propositional linear temporal logic $\mathcal{PTL}$ over various flows of time were obtained by Sistla and Clarke (1985) and Reynolds (2003, 2004). In many cases these investigations resulted in the development and implementation of effective reasoning algorithms (see, e.g., Wolper, 1985; Smith & Park, 1992; Egenhofer & Sharma, 1993; Schwendimann, 1998; Fisher, Dixon, & Peim, 2001; Renz & Nebel, 2001; Hustadt & Konev, 2003).

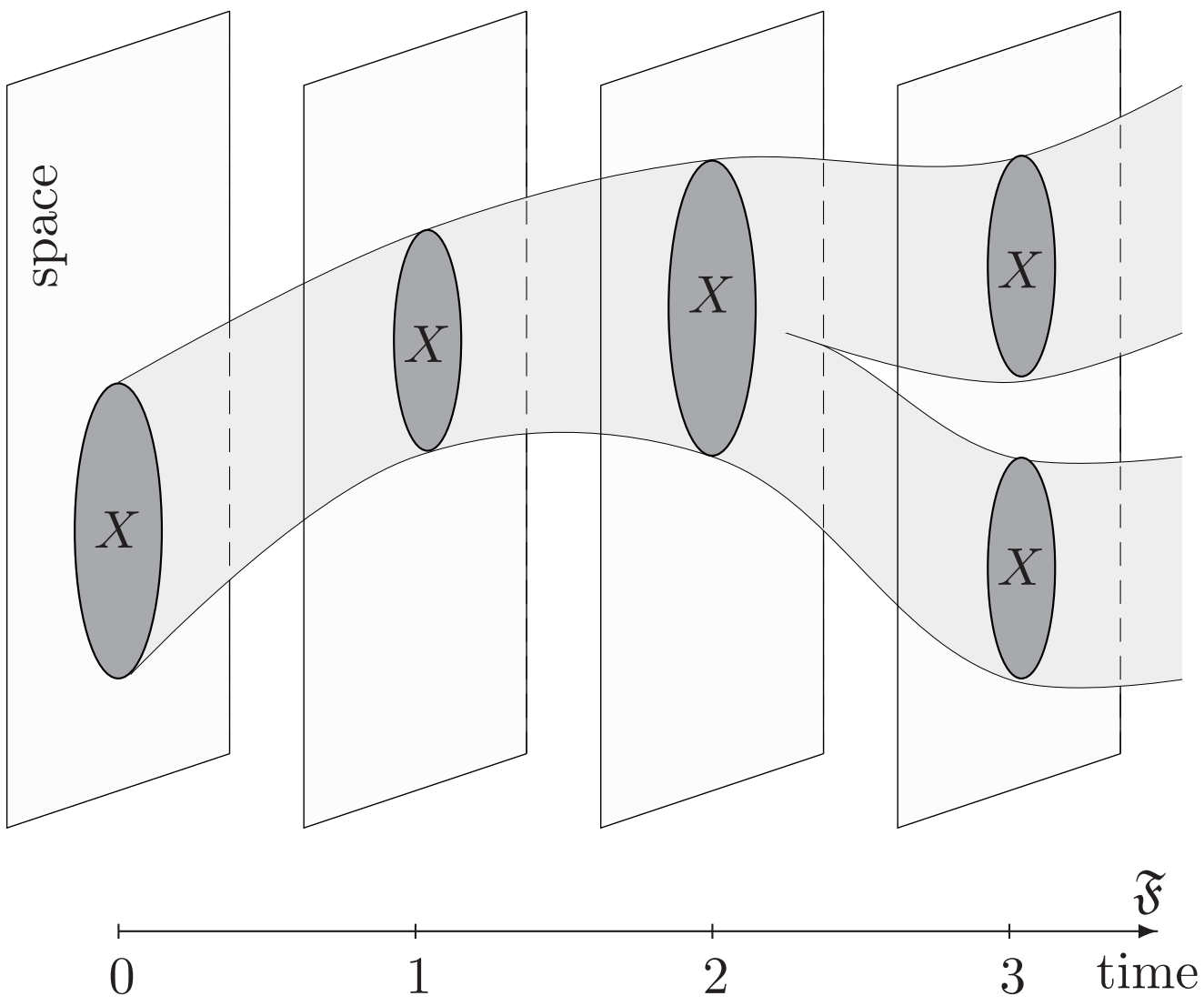

Figure 1: Topological temporal model.

The next apparent and natural step is to *combine* these two kinds of reasoning. Of course, there have been attempts to construct spatio-temporal hybrids. For example, the intended interpretation of Clarke's (1981, 1985) region-based calculus was spatio-temporal. Region connection calculus $\mathcal{RCC}$ (Randell et al., 1992) contained a function $\mathsf{space}(X, t)$ for representing the space occupied by object $X$ at moment of time $t$. Muller (1998a) developed a first-order theory for reasoning about motion of spatial entities. However, all of these formalisms turn out to be 'too expressive' from the computational point of view: they are *undecidable*. Moreover, as far as we know, no serious attempts to investigate and implement partial (say, incomplete) algorithms capable of spatio-temporal reasoning with these logics have been made.

The problem of constructing spatio-temporal logics with better algorithmic properties and analysing their computational complexity was first attacked by Wolter and Zakharyaschev (2000b); see also the 'popular' and extended version (Wolter & Zakharyaschev, 2002) of that conference paper, as well as (Bennett & Cohn, 1999; Bennett, Cohn, Wolter, & Zakharyschev, 2002; Gerevini & Nebel, 2002).

The main idea underlying all these papers is to consider various combinations of 'well-behaved' spatial and temporal logics. The intended spatio-temporal structures can be regarded then as the Cartesian products of the intended time-line and topological (or some other) spaces that are used to model the spatial dimension. Figure 1 shows such a product (of the flow of time $\mathfrak{F} = \langle \mathbb{N}, < \rangle$ and the two-dimensional Euclidean space $\mathfrak{T}$) with a moving spatial object $X$. The moving object can be viewed either as a 3D spatio-temporal entity (in this particular case) or as the collection of the 'snapshots' or slices of this entity at each moment of time; for a discussion see, e.g., (Muller, 1998b) and references therein. In this paper, we use the snapshot terminology and understand by a *moving spatial object* (or, more precisely, interpret such an object as) any set of pairs $\langle X, t \rangle$ where, for each point $t$ of the flow of time, $X$ is a subset of the topological space—the *state of the spatial object at moment $t$*.

The expressive power (and consequently the computational complexity) of the combined spatio-temporal formalisms obviously depends on three parameters:

1. the expressivity of the spatial component,
2. the expressivity of the temporal component, and
3. the *interaction* between the two components allowed in the combined logic.

Regardless of the chosen component languages, the minimal requirement for a spatio-temporal combination to be useful is its ability to

*express changes in time of the truth-values of purely spatial propositions.* **(PC)**

Typical examples of logics meeting this *spatial propositions' truth change principle* are the combinations of $\mathcal{RCC}$-8 and Allen's interval calculus (Bennett et al., 2002; Gerevini & Nebel, 2002) and those combinations of $\mathcal{RCC}$-8 and $\mathcal{PTL}$ introduced by Wolter and Zakharyaschev (2000b) that allow applications of temporal operators to Boolean combinations of $\mathcal{RCC}$-8 relations. Languages satisfying (PC) can capture, for instance, some aspects of the *continuity of change principle* (see, e.g., Cohn, 1997) such as

(A) if two images on the computer screen are disconnected now, then they either remain disconnected or become externally connected in one quantum of the computer's time.

Another example is the following statement about the geography of Europe:

(B) Kaliningrad is disconnected from the EU until the moment when Poland becomes a tangential proper part of the EU, after which Kaliningrad and the EU will be externally connected forever.

However, languages meeting (PC) do not necessarily satisfy our second fundamental *spatial object change principle* according to which we should be able to

*express changes or evolutions of spatial objects in time.* **(OC)**

In logical terms, (PC) refers to the change of truth-values of propositions, while (OC) to the change of extensions of predicates; see Fig. 2 where $\bigcirc X$ at moment $t$ denotes the state of $X$ at moment $t+1$. Here are some examples motivating (OC):

(C) Continuity of change: 'the cyclone's current position overlaps its position in an hour.'

(D) Two physical objects cannot occupy the same space: 'if tomorrow object $X$ is at the place where object $Y$ is today, then $Y$ will have to move by tomorrow.'

(E) Geographic regions change: 'the space occupied by Europe never changes.'

(F) Geographic regions change: 'in two years the EU will be extended with Romania and Bulgaria.'

(G) Fairness conditions on regions: 'it will be raining over every part of England ever and ever again.'

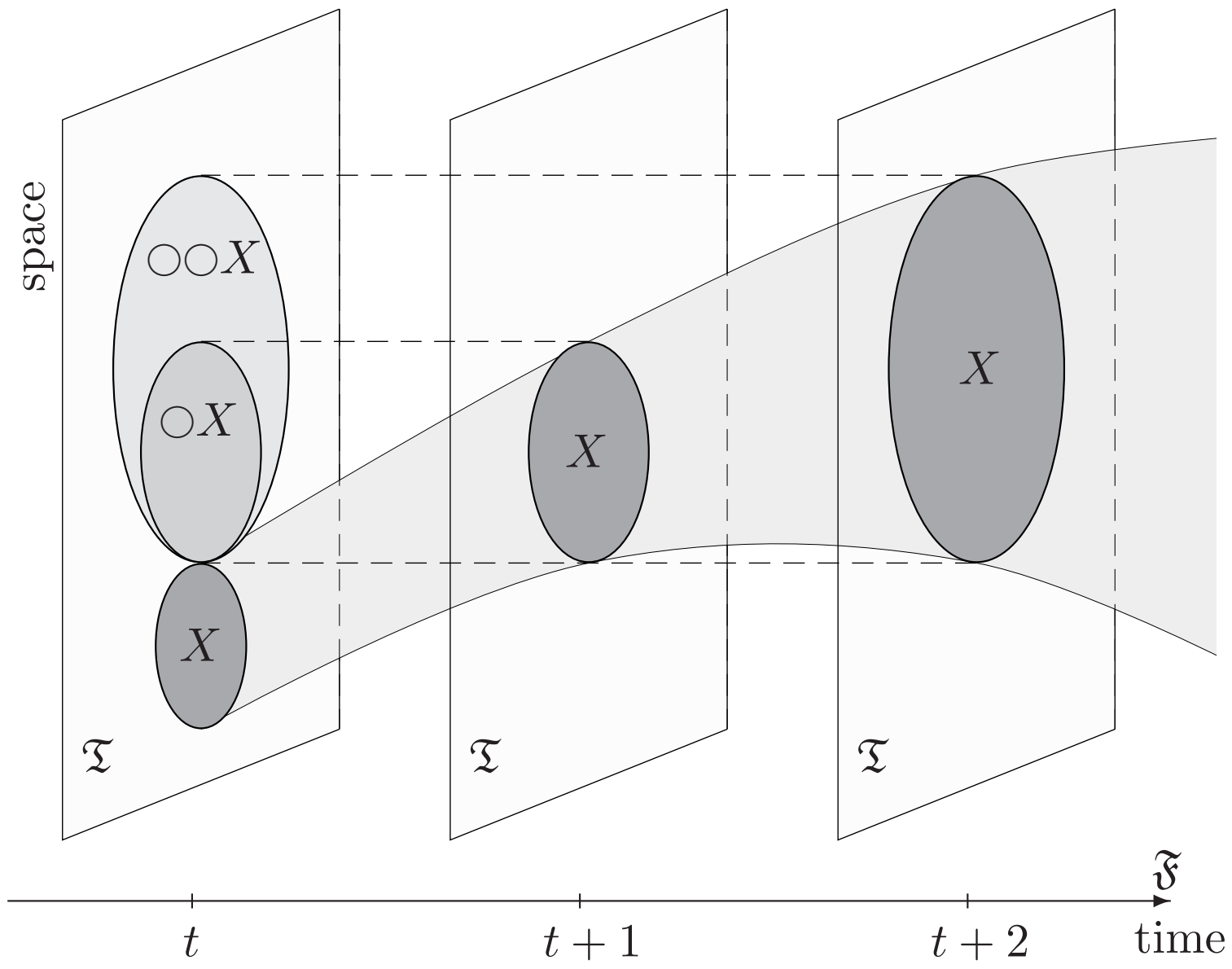

Figure 2: Temporal operators on regions.

(H) Mutual exclusion: ‘if Earth consists of water and land, and the space occupied by water expands, then the space occupied by land shrinks.’

It should be clear that to represent these statements we have to refer to the evolution of spatial objects in time (say, to compare objects $X$ and $\bigcirc\bigcirc X$)—it is not enough to only take into account the change of the truth-values of propositions speaking about spatial objects.

*The main aim of this paper is to investigate the trade-off between the expressive power and the computational behaviour of spatio-temporal hybrids satisfying the* (PC) *and* (OC) *principles and interpreted in various spatio-temporal structures. Our purpose is to show what computational obstacles one can expect if the application domain requires this or that kind of interactions between temporal and spatial operators.*

The spatio-temporal logics we consider below are combinations of fragments of $\mathcal{PTL}$ interpreted over different flows of time with fragments of the propositional spatial logic $\mathcal{S}4_u$ (equipped with the interior and closure operators, the universal and existential quantifiers over points in space as well as the Booleans) interpreted in topological spaces. This choice is motivated by the following reasons:

- The component logics are well understood and established in temporal and spatial knowledge representation; all of them are supported by reasonably effective reasoning procedures.

- By definition, implicit or explicit temporal quantification is necessary to capture (OC), and fragments of $\mathcal{PTL}$ are the weakest languages with such quantification we know of.

Allen's interval calculus, for example, does not provide means for any quantification over intervals. It is certainly suitable for spatio-temporal hybrids satisfying (PC) (see Bennett et al., 2002; Gerevini & Nebel, 2002) but there is no natural conservative way of combining it with spatial formalisms to meet (OC). On the other hand, it is embedded in $\mathcal{PTL}$ (Blackburn, 1992). A natural alternative to $\mathcal{PTL}$ would be the extension of Allen's calculus by means of quantification over intervals introduced by Halpern and Shoham (1986), but unfortunately this temporal logic turns out to be highly undecidable.

- Although the logic $\mathcal{S}4_u$ was originally introduced in the realm of modal logic (see below for details), the work of Bennett (1994), Nutt (1999), Renz (2002) and Wolter and Zakharyaschev (2000a) showed that it can be regarded as a unifying language that contains many spatial formalisms like $\mathcal{RCC}$-8, $\mathcal{BRCC}$-8 or the 9-intersections of Egenhofer and Herring (1991) as fragments.

Apart from the choice of component languages and the level of their interaction, the expressive power and the computational complexity of spatio-temporal logics strongly depend on the restrictions we may want to impose on the intended spatio-temporal structures and the interpretations of spatial objects.

- We can choose among different flows of time (say, discrete or dense, infinite or finite)
- and among different topological spaces (say, arbitrary, Euclidean or Aleksandrov).
- At each time point we can interpret spatial objects as arbitrary subsets of the topological space, as regular closed (or open) ones, as polygons, etc.
- To represent the assumption that everything eventually comes to an end, we only do not know when, one can restrict the class of intended models by imposing the *finite change assumption* which states that no spatial object can change its spatial configuration infinitely often, or the more 'liberal' *finite state assumption* according to which every spatial object can have only finitely many possible states (although it may change its states infinitely often).

The paper is organised as follows. In Section 2 we introduce in full detail the component spatial and temporal logics to be combined later on. In particular, besides the standard spatial logics like $\mathcal{RCC}$-8 or the 9-intersections of Egenhofer and Herring (1991), we consider their generalisations in the framework of $\mathcal{S}4_u$ and investigate the computational complexity. For example, we show that the maximal fragment of $\mathcal{S}4_u$ dealing with regular closed spatial objects turns out to be PSPACE-complete, while a natural generalisation of the 9-intersections is still in NP. In Section 3 we introduce a hierarchy of spatio-temporal logics outlined above, provide them with a topological-temporal semantics, and analyse their computational properties. First we show that spatio-temporal logics satisfying only the (PC) principle are not more complex than their components. Then we consider 'maximal' combinations of $\mathcal{S}4_u$ with (fragments of) $\mathcal{PTL}$ meeting both (PC) and (OC) and see that this straightforward approach does not work: the resulting logics turn out to be undecidable. Finally, we systematically investigate the trade-off between expressivity and complexity of spatio-temporal formalisms and construct a hierarchy of decidable logics satisfying (PC) and (OC) whose complexity ranges from PSPACE to 2EXPSPACE. These and other results, possible implementations as well as open problems are discussed in Section 4. For the reader's convenience most important (un)decidability and complexity results obtained in this paper are summarised in Table 1 on page 193. All technical definitions and detailed proofs can be found in the appendices.

## 2. Propositional Logics of Space and Time

We begin by introducing and discussing the spatial and the temporal formalisms we are going to combine later on in this paper.

### 2.1 Logics of Space

We will be dealing with a number of logics suitable for qualitative spatial representation and reasoning: the well-known $\mathcal{RCC}$-8, $\mathcal{BRCC}$-8 and $\mathcal{S}4_u$, as well as certain fragments of the last one. The intended interpretations for all of these logics are topological spaces.

A *topological space* is a pair $\mathfrak{T} = \langle U, \mathbb{I} \rangle$ in which $U$ is a nonempty set, the *universe* of the space, and $\mathbb{I}$ is the *interior operator* on $U$ satisfying the standard *Kuratowski axioms*: for all $X, Y \subseteq U$,

$$\mathbb{I}(X \cap Y) = \mathbb{I}X \cap \mathbb{I}Y, \quad \mathbb{I}X \subseteq \mathbb{I}\mathbb{I}X, \quad \mathbb{I}X \subseteq X \quad \text{and} \quad \mathbb{I}U = U.$$

The operator dual to $\mathbb{I}$ is called the *closure operator* and denoted by $\mathbb{C}$: for every $X \subseteq U$, we have $\mathbb{C}X = U - \mathbb{I}(U - X)$. Thus, $\mathbb{I}X$ is the *interior* of a set $X$, while $\mathbb{C}X$ is its *closure.* $X$ is called *open* if $X = \mathbb{I}X$ and *closed* if $X = \mathbb{C}X$. The complement of an open set is closed and vice versa. The *boundary* of a set $X \subseteq U$ is defined as $\mathbb{C}X - \mathbb{I}X$. Note that $X$ and $U - X$ have the same boundary.

#### 2.1.1 $\mathcal{S}4_u$

Our most expressive spatial formalism is $\mathcal{S}4_u$—i.e., the propositional modal logic $\mathcal{S}4$ extended with the universal modalities. The 'pedigree' of this logic is quite unusual. $\mathcal{S}4$ was introduced independently by Orlov (1928), Lewis (in Lewis & Langford, 1932), and Gödel (1933) without any intention to reason about space. Orlov and Gödel understood it as a logic of 'provability' (in order to provide a classical interpretation for the intuitionistic logic of Brouwer and Heyting) and Lewis as a logic of necessity and possibility, that is, as a *modal logic.* Besides the Boolean connectives and propositional variables, the language of $\mathcal{S}4$ contains two modal operators: $\mathbf{I}$ (it is necessary or provable) and $\mathbf{C}$, the dual of $\mathbf{I}$ (it is possible or consistent). In other words, the formulas of $\mathcal{S}4$ can be defined as follows:

$$\tau \quad ::= \quad p \quad | \quad \overline{\tau} \quad | \quad \tau_1 \sqcap \tau_2 \quad | \quad \mathbf{I}\tau, \tag{1}$$

where the $p$ are variables. Set $\mathbf{C}\tau = \overline{\mathbf{I}\overline{\tau}}$. We denote the modal operators by $\mathbf{I}$ and $\mathbf{C}$ (rather than the conventional $\Box$ and $\Diamond$) because we understand, following an observation made by several logicians in the late thirties and early forties (Stone, 1937; Tarski, 1938; Tsao Chen, 1938; McKinsey, 1941), $\mathcal{S}4$ as a *logic of topological spaces*: if we interpret the propositional variables as subsets of a topological space, the Booleans as the standard set-theoretic operations, and $\mathbf{I}$ and $\mathbf{C}$ as, respectively, the interior and the closure operators on the space, then an $\mathcal{S}4$-formula is modally consistent if and only if it is satisfiable in a topological space—i.e., its value is not empty under some interpretation.[1]

More precisely, a *topological model* is a pair of the form $\mathfrak{M} = \langle \mathfrak{T}, \mathfrak{U} \rangle$, where $\mathfrak{T} = \langle U, \mathbb{I} \rangle$ is a topological space and $\mathfrak{U}$, a *valuation*, is a map associating with every variable $p$ a set $\mathfrak{U}(p) \subseteq U$. Then the valuation $\mathfrak{U}$ is inductively extended to arbitrary $\mathcal{S}4$-formulas by taking:

$$\mathfrak{U}(\overline{\tau}) = U - \mathfrak{U}(\tau), \qquad \mathfrak{U}(\tau_1 \sqcap \tau_2) = \mathfrak{U}(\tau_1) \cap \mathfrak{U}(\tau_2), \qquad \mathfrak{U}(\mathbf{I}\tau) = \mathbb{I}\mathfrak{U}(\tau).$$

Expressions $\tau$ of the form (1) are interpreted as subsets of topological spaces; that is why we will call them *spatial terms*. In particular, propositional variables of $\mathcal{S}4$ will be understood as *spatial variables*.

The language of $\mathcal{S}4_u$ extends $\mathcal{S}4$ with the *universal* and the *existential quantifiers* $\boxed{\forall}$ and $\diamondsuit$, respectively (known in modal logic as the *universal modalities*). Given a spatial term $\tau$, we write $\diamondsuit\tau$ to say that the part of space (represented by) $\tau$ is not empty (there is at least one point in $\tau$); $\boxed{\forall}\tau$ means that $\tau$ occupies the whole space (all points belong to $\tau$). By taking Boolean combinations of such expressions we arrive at what will be called *spatial formulas*. A BNF definition looks as follows:[2]

$$\varphi \quad ::= \quad \boxed{\forall}\tau \quad | \quad \neg\varphi \quad | \quad \varphi_1 \wedge \varphi_2,$$

where the $\tau$ are spatial terms. Set $\diamondsuit\tau = \neg\boxed{\forall}\overline{\tau}$. Spatial formulas can be either true or false in topological models. The *truth-relation* $\mathfrak{M} \models \varphi$—a spatial formula $\varphi$ is true in a topological model $\mathfrak{M}$—is defined in the standard way:

- $\mathfrak{M} \models \boxed{\forall}\tau \quad \text{iff} \quad \mathfrak{U}(\tau) = U$,
- $\mathfrak{M} \models \neg\varphi \quad \text{iff} \quad \mathfrak{M} \not\models \varphi$,
- $\mathfrak{M} \models \varphi_1 \wedge \varphi_2 \quad \text{iff} \quad \mathfrak{M} \models \varphi_1$ and $\mathfrak{M} \models \varphi_2$.

Say that a spatial formula $\varphi$ is *satisfiable* if there is a topological model $\mathfrak{M}$ such that $\mathfrak{M} \models \varphi$.

The seemingly simple 'query language' $\mathcal{S}4_u$ can express rather complex relations between sets in topological spaces. For example, the formula

$$\boxed{\forall}(q \sqsubset p) \;\wedge\; \boxed{\forall}(p \sqsubset \mathbf{C}q) \;\wedge\; \diamondsuit p \;\wedge\; \neg\diamondsuit\mathbf{I}q$$

says that a set $q$ is dense in a nonempty set $p$, but has no interior (here $\tau_1 \sqsubset \tau_2$ is an abbreviation for $\overline{\tau_1 \sqcap \overline{\tau_2}}$).

The following 'folklore' complexity result has been proved in different settings (see, e.g., Nutt, 1999; Areces, Blackburn, & Marx, 2000):

**Theorem 2.1.** (i) *$\mathcal{S}4_u$ enjoys the exponential finite model property; i.e., every satisfiable spatial formula $\varphi$ is satisfiable in a topological space whose size is at most exponential in the size of $\varphi$.*

(ii) *Satisfiability of spatial formulas in topological models is* PSPACE*-complete.*

1. Moreover, according to McKinsey (1941) and McKinsey and Tarski (1944), any $n$-dimensional *Euclidean* space, for $n \geq 1$, is enough to satisfy all consistent $\mathcal{S}4$-formulas.
2. Formally, the language of $\mathcal{S}4_u$ as defined above is weaker than the standard one, say, that of Goranko and Passy (1992). However, one can easily show that they have precisely the same expressive power: see, e.g., (Hughes & Cresswell, 1996) or (Aiello & van Benthem, 2002b).

One way of proving this theorem is first to observe that every satisfiable spatial formula is satisfied in an *Aleksandrov model*, i.e., a model based on an Aleksandrov topological space—alias a standard Kripke frame for $\mathcal{S}4$ (see, e.g., McKinsey & Tarski, 1944; Goranko & Passy, 1992).

We remind the reader that a topological space is called an *Aleksandrov space* (Alexandroff, 1937) if arbitrary (not only finite) intersections of open sets are open. A *Kripke frame* (or simply a *frame*) for $\mathcal{S}4$ is a pair the form $\mathfrak{G} = \langle V, R\rangle$, where $V$ is a nonempty set and $R$ a transitive and reflexive relation (i.e., a *quasi-order*) on $V$. Every such frame $\mathfrak{G}$ induces the interior operator $\mathbb{I}_{\mathfrak{G}}$ on $V$: for every $X \subseteq V$,

$$\mathbb{I}_{\mathfrak{G}}X = \{x \in X \mid \forall y \in V\ (xRy \rightarrow y \in X)\}.$$

In other words, the open sets of the topological space $\mathfrak{T}_{\mathfrak{G}} = \langle V, \mathbb{I}_{\mathfrak{G}}\rangle$ are the *upward closed* (or *R-closed*) subsets of $V$. The *minimal neighbourhood* of a point $x$ in $\mathfrak{T}_{\mathfrak{G}}$ (that is the minimal open set to contain $x$) consists of all those points that are $R$-accessible from $x$. It is well-known (see, e.g., Bourbaki, 1966) that $\mathfrak{T}_{\mathfrak{G}}$ is an Aleksandrov space and, conversely, every Aleksandrov space is induced by a quasi-order.

Now, to complete the proof, it suffices to recall that $\mathcal{S}4$ is PSPACE-hard (Ladner, 1977) and use, say, the standard tableau technique to establish the exponential finite model property and construct a PSPACE satisfiability checking algorithm for spatial formulas.

Although being of the same computational complexity as $\mathcal{S}4$, the logic $\mathcal{S}4_u$ is more expressive. A standard example is that spatial formulas can distinguish between arbitrary and connected[3] topological spaces. Consider, for instance, the formula

$$\boxdot(\mathbf{C}p \sqsubset p)\ \wedge\ \boxdot(p \sqsubset \mathbf{I}p)\ \wedge\ \diamonddot p\ \wedge\ \neg\boxdot p \tag{2}$$

saying that $p$ is both closed and open, nonempty and does not coincide with the whole space. It can be satisfied only in a model whose underlying topological space is not connected, while all satisfiable $\mathcal{S}4$-formulas are satisfied in connected (e.g., Euclidean) spaces.

Another example illustrating the expressive power of $\mathcal{S}4_u$ is the formula

$$\diamonddot p \wedge \boxdot(p \sqsubset \mathbf{C}\overline{p}) \wedge \boxdot(\overline{p} \sqsubset \mathbf{C}p) \tag{3}$$

defining a nonempty set $p$ such that both $p$ and $\overline{p}$ have empty interiors. In fact, the second and the third conjuncts say that $p$ and $\overline{p}$ consist of boundary points only.

### 2.1.2 $\mathcal{RCC}$-8 as a Fragment of $\mathcal{S}4_u$

In qualitative spatial representation and reasoning, it is quite often assumed that spatial terms can only be interpreted by regular closed (or open) sets of topological spaces (see, e.g., Davis, 1990; Asher & Vieu, 1995; Gotts, 1996). One of the reasons for imposing this restriction is to exclude from consideration such 'pathological' sets as $p$ in (3). Recall that a set $X$ is *regular closed* if $X = \mathbb{C}\mathbb{I}X$, which clearly does not hold for any set $p$ satisfying (3). Another reason is to ensure that the space occupied by a physical body is homogeneous in the sense that it does not contain parts of 'different dimensionality.' For example, the

3. We remind the reader that a topological space is *connected* if its universe cannot be represented as the union of two disjoint nonempty open sets.

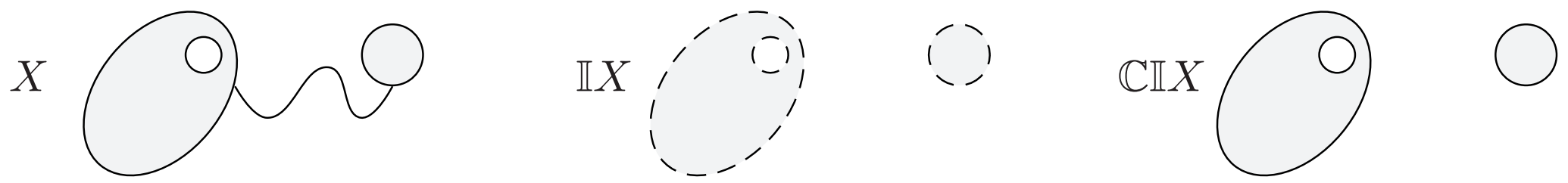

Figure 3: Regular closure.

subset $X$ of the Euclidean plane in Fig. 3 consists of three parts: a 2D ellipse with a hole, a 2D circle, and a 1D curve connecting them. This curve disappears if we form the set $\mathbb{C}\mathbb{I}X$, which is regular closed because $\mathbb{C}\mathbb{I}\mathbb{C}\mathbb{I}X = \mathbb{C}\mathbb{I}X$, for every $X$ and every topological space.

In this paper, we will consider several fragments of $\mathcal{S}4_u$ dealing with *regular closed sets*. From now on we will call such sets *regions*. Perhaps, the best known language devised for speaking about regions is $\mathcal{RCC}$-8 which was introduced in the area of Geographical Information Systems (see Egenhofer & Franzosa, 1991; Smith & Park, 1992) and as a decidable subset of Region Connection Calculus $\mathcal{RCC}$ (Randell et al., 1992). The syntax of $\mathcal{RCC}$-8 contains eight binary predicates,

- $\mathsf{DC}(X, Y)$ — regions $X$ and $Y$ are disconnected,
- $\mathsf{EC}(X, Y)$ — $X$ and $Y$ are externally connected,
- $\mathsf{EQ}(X, Y)$ — $X$ and $Y$ are equal,
- $\mathsf{PO}(X, Y)$ — $X$ and $Y$ partially overlap,
- $\mathsf{TPP}(X, Y)$ — $X$ is a tangential proper part of $Y$,
- $\mathsf{NTPP}(X, Y)$ — $X$ is a nontangential proper part of $Y$,
- the inverses of the last two—$\mathsf{TPPi}(X, Y)$ and $\mathsf{NTPPi}(X, Y)$,

which can be combined using the Boolean connectives. For example, given a spatial database describing the geography of Europe, we can query whether the United Kingdom and the Republic of Ireland share a common border. The answer can be found by checking whether the $\mathcal{RCC}$-8 formula $\mathsf{EC}(\mathit{UK}, \mathit{RoI})$ follows from the database.

The arguments of the $\mathcal{RCC}$-8 predicates are called *region variables*; they are interpreted by regular closed sets—i.e., regions—of topological spaces. The satisfiability problem for $\mathcal{RCC}$-8 formulas under such interpretations is NP-complete (Renz & Nebel, 1999).

The expressive power of $\mathcal{RCC}$-8 is rather limited. It only operates with 'simple' regions and does not distinguish between connected and disconnected ones, regions with and without holes, etc. (Egenhofer & Herring, 1991). Nor can $\mathcal{RCC}$-8 represent complex relations between more than two regions. Consider, for example, three countries (say, Russia, Lithuania and Poland) such that not only each one of them is adjacent to the others, but there is a point where all the three meet. To express this fact we may need a ternary predicate like

$$\mathsf{EC3}(\mathit{Russia}, \mathit{Lithuania}, \mathit{Poland}). \tag{4}$$

To analyse possible ways of extending the expressive power of $\mathcal{RCC}$-8, it will be convenient to view it as a fragment of $\mathcal{S}4_u$ (that $\mathcal{RCC}$-8 can be embedded into $\mathcal{S}4_u$ was first shown by Bennett, 1994). Observe first that, for every spatial variable $p$, the spatial term

$$\mathbf{CI}p \tag{5}$$

is interpreted as a regular closed set in every topological model. So, with every region variable $X$ of $\mathcal{RCC}$-8 we can associate the spatial term $\varrho_X = \mathbf{CI}p_X$, where $p_X$ is a spatial variable, and then translate the $\mathcal{RCC}$-8 predicates into spatial formulas by taking:

$$\begin{aligned}
\mathsf{EC}(X,Y) &= \Diamond_\exists(\varrho_X \sqcap \varrho_Y) \wedge \neg\Diamond_\exists(\mathbf{I}\varrho_X \sqcap \mathbf{I}\varrho_Y),\\
\mathsf{DC}(X,Y) &= \neg\Diamond_\exists(\varrho_X \sqcap \varrho_Y),\\
\mathsf{EQ}(X,Y) &= \Box_\forall(\varrho_X \sqsubset \varrho_Y) \wedge \Box_\forall(\varrho_Y \sqsubset \varrho_X),\\
\mathsf{PO}(X,Y) &= \Diamond_\exists(\mathbf{I}\varrho_X \sqcap \mathbf{I}\varrho_Y) \wedge \neg\Box_\forall(\varrho_X \sqsubset \varrho_Y) \wedge \neg\Box_\forall(\varrho_Y \sqsubset \varrho_X),\\
\mathsf{TPP}(X,Y) &= \Box_\forall(\varrho_X \sqsubset \varrho_Y) \wedge \neg\Box_\forall(\varrho_Y \sqsubset \varrho_X) \wedge \neg\Box_\forall(\varrho_X \sqsubset \mathbf{I}\varrho_Y),\\
\mathsf{NTPP}(X,Y) &= \Box_\forall(\varrho_X \sqsubset \mathbf{I}\varrho_Y) \wedge \neg\Box_\forall(\varrho_Y \sqsubset \varrho_X)
\end{aligned}$$

($\mathsf{TPPi}$ and $\mathsf{NTPPi}$ are the mirror images of $\mathsf{TPP}$ and $\mathsf{NTPP}$, respectively). The first of these formulas, for instance, says that two regions are externally connected iff the intersection of the regions is not empty, whereas the intersection of their interiors is. It should be clear that an $\mathcal{RCC}$-8 formula is satisfiable in a topological space if and only if its translation into $\mathcal{S}4_u$ defined above is satisfiable in a topological model.

This translation also shows that in $\mathcal{RCC}$-8 any two regions can be related in terms of truth/falsity of atomic spatial formulas of the form

$$\Box_\forall(\overline{\varrho_1 \sqcap \varrho_2}), \qquad \Box_\forall(\overline{\mathbf{I}\varrho_1 \sqcap \mathbf{I}\varrho_2}), \qquad \Box_\forall(\varrho_1 \sqsubset \varrho_2) \quad \text{and} \quad \Box_\forall(\varrho_1 \sqsubset \mathbf{I}\varrho_2),$$

where $\varrho_1$ and $\varrho_2$ are spatial terms of the form (5). For example, the first of these formulas says that the intersection of two regions is empty, whereas the last one states that one region is contained in the interior of another one. In other words, $\mathcal{RCC}$-8 can be regarded as part of the following fragment of $\mathcal{S}4_u$:

$$\begin{aligned}
\varrho \quad &::= \quad \mathbf{CI}p,\\
\tau \quad &::= \quad \overline{\varrho_1 \sqcap \varrho_2} \quad | \quad \overline{\mathbf{I}\varrho_1 \sqcap \mathbf{I}\varrho_2} \quad | \quad \varrho_1 \sqsubset \varrho_2 \quad | \quad \varrho_1 \sqsubset \mathbf{I}\varrho_2,\\
\varphi \quad &::= \quad \Box_\forall\tau \quad | \quad \neg\varphi \quad | \quad \varphi_1 \wedge \varphi_2.
\end{aligned}$$

Here we distinguish between two types of spatial terms. Those of the form $\varrho$ will be called *atomic region terms*—they represent the (regular closed) regions we want to compare. Spatial terms of the form $\tau$ are used to relate regions to each other (note that their extensions are not necessarily regular closed).

Actually, the fragment introduced above is a bit more expressive than $\mathcal{RCC}$-8: for example, it contains (appropriately modified) formula (2) which can be satisfied only in disconnected topological spaces, while all satisfiable $\mathcal{RCC}$-8 formulas are satisfiable in any Euclidean space (Renz, 1998). However, it will be convenient for us *not to distinguish* between these two spatial logics. First, it will turn out that the same technical results regarding their computational complexity hold for them even when combined with temporal logics. And second, the more intuitive and concise language of $\mathcal{RCC}$-8 is more suitable for illustrations. For instance, we do not distinguish between the region variable $X$ and the region term $\varrho_X$ and use $\mathsf{DC}(\varrho_1, \varrho_2)$ as an abbreviation for $\neg\Diamond(\varrho_1 \sqcap \varrho_2)$.

The definition above suggests two ways of increasing the expressive power of $\mathcal{RCC}$-8 (while keeping all regions regular closed):

(i) by allowing more complex region terms $\varrho$, and

(ii) by allowing more ways of relating them (i.e., more complex terms $\tau$).

#### 2.1.3 $\mathcal{BRCC}$-8 as a Fragment of $\mathcal{S}4_u$

The language $\mathcal{BRCC}$-8 of Wolter and Zakharyaschev (2000a) (see also Balbiani, Tinchev, & Vakarelov, 2004) extends $\mathcal{RCC}$-8 in direction (i). It uses the same eight binary predicates as $\mathcal{RCC}$-8 and allows not only atomic regions but also their intersections, unions and complements. For instance, in $\mathcal{BRCC}$-8 we can express the fact that a region (say, the Swiss Alps) is the intersection of two other regions (Switzerland and the Alps in this case):

$$\mathsf{EQ}(SwissAlps, Switzerland \sqcap Alps). \tag{6}$$

We can embed $\mathcal{BRCC}$-8 to $\mathcal{S}4_u$ by using almost the same translation as in the case of $\mathcal{RCC}$-8. The only difference is that now, since Boolean combinations of regular closed sets are not necessarily regular closed, we should prefix compound spatial terms with **CI**. This way we can obtain, for example, the spatial term

$$\mathbf{CI}\,(Switzerland \sqcap Alps)$$

representing the Swiss Alps. In the same manner we can treat other set-theoretic operations, which leads us to the following definition of *Boolean region terms*:

$$\varrho \quad ::= \quad \mathbf{CI}p \quad | \quad \mathbf{CI}\overline{\varrho} \quad | \quad \mathbf{CI}(\varrho_1 \sqcap \varrho_2).$$

In other words, Boolean region terms denote precisely the members of the well-known Boolean algebra of regular closed sets. (The union $\sqcup$ is expressible via intersection and complement in the usual way.) To simplify notation, given a spatial term $\tau$, we write $\lceil\tau\rceil$ to denote the result of prefixing **CI** to every subterm of $\tau$; in particular,

$$\lceil p \rceil = \mathbf{CI}p, \qquad \lceil\overline{\tau}\rceil = \mathbf{CI}\overline{\lceil\tau\rceil} \quad \text{and} \quad \lceil\tau_1 \sqcap \tau_2\rceil = \mathbf{CI}(\lceil\tau_1\rceil \sqcap \lceil\tau_2\rceil).$$

Note that $\lceil\tau\rceil$ is (equivalent to) a Boolean region term, for every spatial term $\tau$. Now the Swiss Alps from the example above can be represented as $\lceil Switzerland \sqcap Alps\rceil$.

It is of interest to note that Boolean region terms do not increase the complexity of reasoning in arbitrary topological models: the satisfiability problem for $\mathcal{BRCC}$-8 formulas is still NP-complete (however, it becomes PSPACE-complete if all intended models are based on connected spaces). On the other hand, $\mathcal{BRCC}$-8 allows some restricted comparisons of more than two regions as, e.g., in (6). Nevertheless, as we shall see below, ternary relations like (4) are still unavailable in $\mathcal{BRCC}$-8: they require different ways of comparing regions; cf. (ii).

### 2.1.4 $\mathcal{RC}$

Egenhofer and Herring (1991) proposed to relate any *two* regions in terms of the 9-intersections—3 × 3-matrix specifying emptiness/nonemptiness of all (nine) possible intersections of the interiors, boundaries and exteriors of the regions. Recall that, for a region $X$, these three disjoint parts of the space $\langle U, \mathbb{I}\rangle$ can be represented as

$$\mathbb{I}X, \qquad X \cap (U - \mathbb{I}X) \qquad \text{and} \qquad U - X,$$

respectively. By generalising this approach to any finite number of regions, we obtain the following fragment $\mathcal{RC}$ of $\mathcal{S}4_u$:

$$\begin{aligned} \varrho \quad &::= \quad \text{Boolean region terms}, \\ \tau \quad &::= \quad \varrho \quad | \quad \mathbf{I}\varrho \quad | \quad \overline{\tau} \quad | \quad \tau_1 \sqcap \tau_2, \\ \varphi \quad &::= \quad \boxdot\tau \quad | \quad \neg\varphi \quad | \quad \varphi_1 \wedge \varphi_2. \end{aligned}$$

In other words, in $\mathcal{RC}$ we can define relations between regions in terms of emptiness/nonemptiness of sets formed by using arbitrary set-theoretic operations on regions and their interiors. However, nested applications of the topological operators are not allowed (an example where such applications are required can be found in the next section).

Clearly, both $\mathcal{RCC}$-8 and $\mathcal{BRCC}$-8 are fragments of $\mathcal{RC}$. Moreover, unlike $\mathcal{BRCC}$-8, the language of $\mathcal{RC}$ allows us to consider more complex relations between regions. For instance, the ternary relation required in (4) can now be defined as follows:

$$\mathsf{EC3}(X, Y, Z) = \diamond(\varrho_X \sqcap \varrho_Y \sqcap \varrho_Z) \wedge \neg\diamond(\mathbf{I}\varrho_X \sqcap \mathbf{I}\varrho_Y) \wedge \neg\diamond(\mathbf{I}\varrho_Y \sqcap \mathbf{I}\varrho_Z) \wedge \neg\diamond(\mathbf{I}\varrho_Z \sqcap \mathbf{I}\varrho_X).$$

Another, more abstract, example is the formula

$$\diamond\big(\varrho_1 \sqcap \cdots \sqcap \varrho_i \wedge \mathbf{I}\varrho_{i+1} \sqcap \cdots \sqcap \mathbf{I}\varrho_j \sqcap \overline{\varrho_{j+1}} \sqcap \cdots \sqcap \overline{\varrho_k} \sqcap \overline{\mathbf{I}\varrho_{k+1}} \sqcap \cdots \sqcap \overline{\mathbf{I}\varrho_n}\big)$$

which says that

> *regions $\varrho_1, \ldots, \varrho_i$ meet somewhere inside the region occupied jointly by all $\varrho_{i+1}, \ldots, \varrho_j$, but outside the regions $\varrho_{j+1}, \ldots, \varrho_k$ and not inside $\varrho_{k+1}, \ldots, \varrho_n$.*

Although $\mathcal{RC}$ is more expressive than both $\mathcal{RCC}$-8 and $\mathcal{BRCC}$-8, reasoning in this language is still of the same computational complexity:

**Theorem 2.2.** *The satisfiability problem for $\mathcal{RC}$-formulas in arbitrary topological models is* NP*-complete.*

This result will be proved in Appendix A. Lemma A.1 shows that every satisfiable $\mathcal{RC}$-formula can be satisfied in a model based on the Aleksandrov space that is induced by a disjoint union of *n-brooms*—i.e., quasi-orders of the form depicted in Fig. 4. Topological spaces of this kind have a rather primitive structure satisfying the following property:

(rc) only the roots of $n$-brooms can be boundary points, and the minimal neighbourhood of every boundary point—i.e., the $n$-broom containing this point—must contain *at least one* internal point and *at least one* external point.

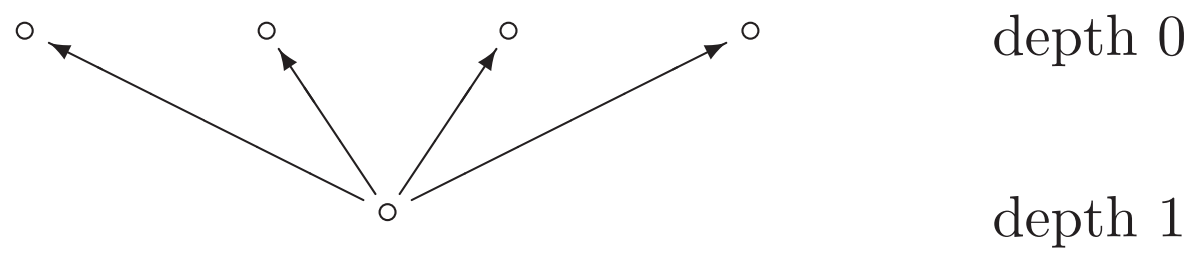

Figure 4: $n$-broom (for $n = 4$).

For example, spatial formula (3) cannot be satisfied in a model with this property, and so it is not in $\mathcal{RC}$.

By Lemma A.2, the size of such a satisfying model is polynomial (in fact, quadratical) in the length of the input $\mathcal{RC}$-formula, and so we have a nondeterministic polynomial time algorithm. Actually, the proof is a straightforward generalisation of the complexity proof for $\mathcal{BRCC}$-8 given by Wolter and Zakharyaschev (2000a): the only difference is that in the case of $\mathcal{BRCC}$-8 it is sufficient to consider only 2-brooms (which were called *forks*). This means, in particular, that ternary relation (4)—which is satisfiable only in a model with an $n$-broom, for $n \geq 3$—is indeed not expressible in $\mathcal{BRCC}$-8.

*Remark* 2.3. In topological terms, $n$-brooms are examples of so-called *door spaces* where every subset is either open or closed. However, the modal theory of $n$-brooms defines a wider and more interesting topological class known as *submaximal spaces* in which every dense subset is open. Submaximal spaces have been around since the early 1960s and have generated interesting and challenging problems in topology. For a survey and a systematic study of these spaces see (Arhangel'skii & Collins, 1995) and references therein.

#### 2.1.5 $\mathcal{RC}^{max}$

One could go even further in direction (ii) and impose no restrictions whatsoever on the ways of relating Boolean region terms. This leads us to the *maximal* fragment $\mathcal{RC}^{max}$ of $\mathcal{S}4_u$ in which spatial terms are interpreted by regular closed sets. Its syntax is defined as follows:

$$\begin{array}{lcl} \varrho & ::= & \text{Boolean region terms}, \\ \tau & ::= & \varrho \quad | \quad \overline{\tau} \quad | \quad \tau_1 \sqcap \tau_2 \quad | \quad \mathbf{I}\tau, \\ \varphi & ::= & \boxdot\tau \quad | \quad \neg\varphi \quad | \quad \varphi_1 \wedge \varphi_2, \end{array}$$

To understand the difference between $\mathcal{RC}$ and $\mathcal{RC}^{max}$, consider the $\mathcal{RC}^{max}$-formula

$$\diamondplus(\lceil q_1\rceil \sqcap \overline{\mathbf{I}\lceil q_1\rceil}) \;\wedge\; \boxdot\Big((\lceil q_1\rceil \sqcap \overline{\mathbf{I}\lceil q_1\rceil}) \sqsubset \mathbf{C}\big(\mathbf{I}\lceil q_1\rceil \sqcap \lceil q_2\rceil \sqcap \overline{\mathbf{I}\lceil q_2\rceil}\big)\Big). \tag{7}$$

It says that the boundary of $\lceil q_1\rceil$ is not empty and that every neighbourhood of every point in this boundary contains an internal point of $\lceil q_1\rceil$ that belongs to the boundary of $\lceil q_2\rceil$ (compare with property (rc) above). The simplest Aleksandrov model satisfying this formula is of depth 2; it is shown in Fig. 5.

The price we have to pay for this expressivity is that the complexity of $\mathcal{RC}^{max}$ is the same as that of full $\mathcal{S}4_u$:

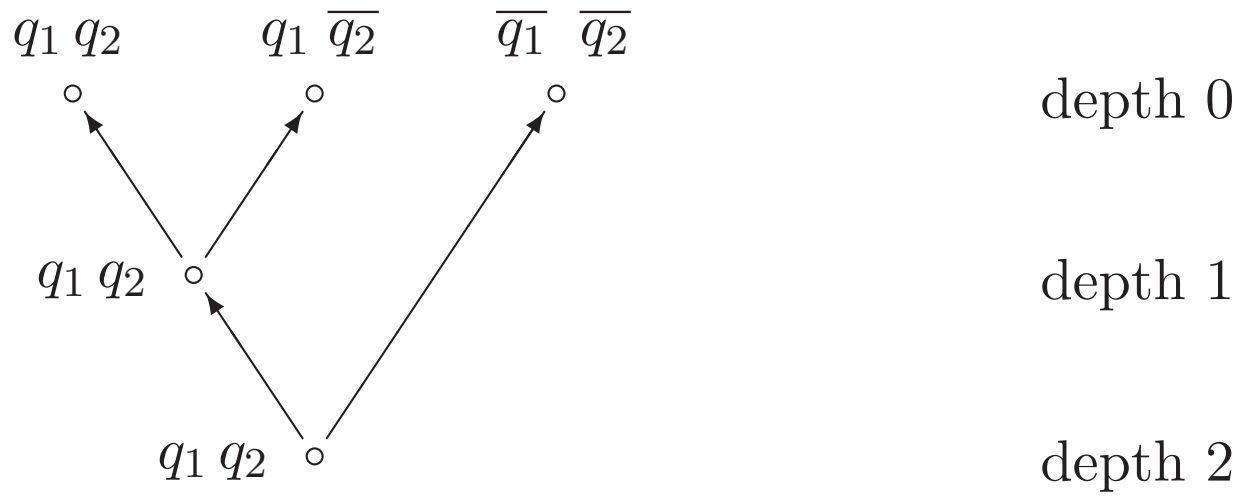

Figure 5: Model satisfying formula (7).

**Theorem 2.4.** *The satisfiability problem for $\mathcal{RC}^{max}$-formulas is* PSPACE-*complete.*

The upper bound follows from Theorem 2.1 and the lower bound is proved in Appendix A, where we construct a sequence of $\mathcal{RC}^{max}$-formulas such that each of them is satisfiable in an Aleksandrov space of cardinality at least exponential in the length of the formula. The first formula of the sequence is similar to (7) above.

It is of interest to note, however, that $\mathcal{RC}^{max}$ is still not expressive enough to define such 'pathological' sets as $p$ in (3) which is clearly not regular closed.

To conclude this section, we summarise the inclusions between the spatial languages introduced above:

$$\mathcal{RCC}\text{-}8 \quad \subsetneqq \quad \mathcal{BRCC}\text{-}8 \quad \subsetneqq \quad \mathcal{RC} \quad \subsetneqq \quad \mathcal{RC}^{max} \quad \subsetneqq \quad \mathcal{S}4_u.$$

For more discussions of spatial logics of this kind we refer the reader to the paper (Pratt-Hartmann, 2002).

### 2.2 Temporal Logics

As was said in the introduction, the temporal components of our spatio-temporal hybrids are (fragments of) the *propositional temporal logic $\mathcal{PTL}$* interpreted in various *flows of time* which are modelled by strict linear orders $\mathfrak{F} = \langle W, < \rangle$, where $W$ is a nonempty set of *time points* and $<$ is a (connected, transitive and irreflexive) *precedence* relation on $W$.

The language $\mathcal{PTL}$ is based on the following alphabet:

- propositional variables $p_0, p_1, \dots,$
- the Booleans $\neg$ and $\wedge$, and
- the binary temporal operators $\mathcal{U}$ ('until') and $\mathcal{S}$ ('since').

The set of *$\mathcal{PTL}$-formulas* is defined in the standard way:

$$\varphi \quad ::= \quad p \quad | \quad \neg\varphi \quad | \quad \varphi_1 \wedge \varphi_2 \quad | \quad \varphi_1\, \mathcal{U}\, \varphi_2 \quad | \quad \varphi_1\, \mathcal{S}\, \varphi_2.$$

*$\mathcal{PTL}$-models* are pairs of the form $\mathfrak{M} = \langle \mathfrak{F}, \mathfrak{V} \rangle$ such that $\mathfrak{F} = \langle W, < \rangle$ is a flow of time and $\mathfrak{V}$, a *valuation*, is a map associating with each variable $p$ a set $\mathfrak{V}(p) \subseteq W$ of time points (where $p$ is supposed to be true). The *truth-relation* $(\mathfrak{M}, w) \models \varphi$, for an arbitrary $\mathcal{PTL}$-formula $\varphi$ and $w \in W$, is defined inductively as follows, where $(u, v)$ denotes the open interval $\{w \in W \mid u < w < v\}$:

- $(\mathfrak{M}, w) \models p$ iff $w \in \mathfrak{V}(p)$,
- $(\mathfrak{M}, w) \models \neg\varphi$ iff $(\mathfrak{M}, w) \not\models \varphi$,
- $(\mathfrak{M}, w) \models \varphi_1 \wedge \varphi_2$ iff $(\mathfrak{M}, w) \models \varphi_1$ and $(\mathfrak{M}, w) \models \varphi_2$,
- $(\mathfrak{M}, w) \models \varphi_1 \,\mathcal{U}\, \varphi_2$ iff there is $v > w$ such that $(\mathfrak{M}, v) \models \varphi_2$ and $(\mathfrak{M}, u) \models \varphi_1$ for all $u \in (w, v)$,
- $(\mathfrak{M}, w) \models \varphi_1 \,\mathcal{S}\, \varphi_2$ iff there is $v < w$ such that $(\mathfrak{M}, v) \models \varphi_2$ and $(\mathfrak{M}, u) \models \varphi_1$ for all $u \in (v, w)$.

A $\mathcal{PTL}$-formula $\varphi$ is *satisfied* in $\mathfrak{M}$ if $(\mathfrak{M}, w) \models \varphi$ for some $w \in W$.

We took the operators $\mathcal{U}$ and $\mathcal{S}$ as primitive simply because all other important temporal operators can be defined via them. For example, $\Diamond_F$ ('sometime in the future') and $\Box_F$ ('always in the future') are expressible via $\mathcal{U}$ as

$$\Diamond_F \varphi = \top \,\mathcal{U}\, \varphi, \qquad \Box_F \varphi = \neg \Diamond_F \neg \varphi,$$

($\top$ is the logical constant 'true') which means that

- $(\mathfrak{M}, w) \models \Diamond_F \varphi$ iff there is $v > w$ such that $(\mathfrak{M}, v) \models \varphi$,
- $(\mathfrak{M}, w) \models \Box_F \varphi$ iff $(\mathfrak{M}, v) \models \varphi$ for all $v > w$.

As our intended flows of time are *strict* linear orders, the 'next-time' operator $\bigcirc$ is also definable via $\mathcal{U}$ by taking

$$\bigcirc \varphi = \bot \,\mathcal{U}\, \varphi$$

($\bot$ is the logical constant 'false') which perfectly reflects our intuition: if $\mathfrak{F}$ is discrete then

- $(\mathfrak{M}, w) \models \bigcirc \psi$ iff $(\mathfrak{M}, w + 1) \models \psi$,

where $w + 1$ is the immediate successor of $w$ in $\mathfrak{F}$. The reader should not have problems in defining the 'past' versions of $\Diamond_F$, $\Box_F$ and $\bigcirc$.

The following results are due to Sistla and Clarke (1985) and Reynolds (2003, 2004):

**Theorem 2.5.** *The satisfiability problem for $\mathcal{PTL}$-formulas is* PSPACE*-complete for each of the following classes of flows of time: all strict linear orders, all finite strict linear orders, $\langle \mathbb{N}, < \rangle$, $\langle \mathbb{Z}, < \rangle$, $\langle \mathbb{Q}, < \rangle$, $\langle \mathbb{R}, < \rangle$.*

Note, however, that reasoning becomes somewhat simpler if we take $\Diamond_F$, $\Box_F$ and their past counterparts (but no $\bigcirc$, $\mathcal{U}$ and $\mathcal{S}$) as the only temporal primitives. Denote by $\mathcal{PTL}_\Box$ the corresponding fragment of $\mathcal{PTL}$. Then, according to the results of Ono and Nakamura (1980), Sistla and Clarke (1985), and Wolter (1996), we have:

**Theorem 2.6.** *The satisfiability problem for $\mathcal{PTL}_\Box$-formulas is* NP*-complete for each of the classes of flows of time mentioned in Theorem 2.5.*

## 3. Combinations of Spatial and Temporal Logics

In this section we introduce and discuss various ways of combining logics of space and time. First we construct spatio-temporal logics satisfying only the (PC) principle (see the introduction) and show that they inherit good computational properties of their components. Being encouraged by these results, we then consider 'maximal' combinations of $\mathcal{S}4_u$ with (fragments of) $\mathcal{PTL}$ meeting both (PC) and (OC) and see that such a straightforward approach does not work: we end up with undecidable logics. This leads us to a systematic investigation of the trade-off between expressivity and computational complexity of spatio-temporal formalisms. The result is a hierarchy of decidable logics satisfying (PC) and (OC) whose complexity ranges from PSPACE to 2EXPSPACE.

### 3.1 Spatio-Temporal Logics with (PC)

We begin our investigation of combinations of the spatial and temporal logics introduced above by considering the language $\mathcal{PTL}[\mathcal{S}4_u]$ in which the temporal operators can be applied to spatial formulas but not to spatial terms (this way of 'temporalising' a logic was first introduced by Finger and Gabbay, 1992). A precise syntactic definition of $\mathcal{PTL}[\mathcal{S}4_u]$-*terms* $\tau$ and $\mathcal{PTL}[\mathcal{S}4_u]$-*formulas* $\varphi$ is as follows:

$$\begin{aligned}
\tau \quad &::= \quad p \quad | \quad \overline{\tau} \quad | \quad \tau_1 \sqcap \tau_2 \quad | \quad \mathbf{I}\tau,\\
\varphi \quad &::= \quad \boxdot\tau \quad | \quad \neg\varphi \quad | \quad \varphi_1 \wedge \varphi_2 \quad | \quad \varphi_1 \,\mathcal{U}\, \varphi_2 \quad | \quad \varphi_1 \,\mathcal{S}\, \varphi_2.
\end{aligned}$$

Note that the definition of $\mathcal{PTL}[\mathcal{S}4_u]$-terms coincides with the definition of spatial terms in $\mathcal{S}4_u$ which reflects the fact that $\mathcal{PTL}[\mathcal{S}4_u]$ cannot capture the change of spatial objects in time. We have imposed no restrictions upon the temporal operators in formulas—so the combined language still has the full expressive power of $\mathcal{PTL}$. (Clearly, $\mathcal{S}4_u$ is a fragment of $\mathcal{PTL}[\mathcal{S}4_u]$.)

In a similar way we can introduce spatio-temporal logics based on all other spatial languages we are dealing with: $\mathcal{RCC}$-8, $\mathcal{BRCC}$-8, $\mathcal{RC}$ and $\mathcal{RC}^{max}$. For example, the temporalisation $\mathcal{PTL}[\mathcal{BRCC}\text{-}8]$ of $\mathcal{BRCC}$-8 (denoted by $\mathcal{ST}_0$ in the hierarchy of Wolter and Zakharyaschev 2002) allows applications of the temporal operators to $\mathcal{RCC}$-8 predicates but not to Boolean region terms. These languages can be regarded as fragments of $\mathcal{PTL}[\mathcal{S}4_u]$ in precisely the same way as their spatial components were treated as fragments of $\mathcal{S}4_u$.

We illustrate the expressive power of $\mathcal{PTL}[\mathcal{RCC}\text{-}8]$ by formalising sentences (A) and (B) from the introduction:

$$\mathsf{DC}(\mathit{Image}_1, \mathit{Image}_2) \to \bigcirc\mathsf{DC}(\mathit{Image}_1, \mathit{Image}_2) \vee \bigcirc\mathsf{EC}(\mathit{Image}_1, \mathit{Image}_2), \tag{A}$$

$$\big(\mathsf{DC}(\mathit{Kaliningrad}, \mathit{EU}) \,\mathcal{U}\, \mathsf{TPP}(\mathit{Poland}, \mathit{EU})\big) \wedge \Box_F\big(\mathsf{TPP}(\mathit{Poland}, \mathit{EU}) \to \mathsf{EC}(\mathit{Kaliningrad}, \mathit{EU})\big). \tag{B}$$

Sentences (C)–(H) cannot be expressed in this language (or even in $\mathcal{PTL}[\mathcal{S}4_u]$): they require comparisons of states of spatial objects at different time instants.

The intended semantics of $\mathcal{PTL}[\mathcal{S}4_u]$ (and all other spatio-temporal logics considered in this paper) is rather straightforward. A *topological temporal model* (a *tt-model*, for short) is a triple of the form $\mathfrak{M} = \langle \mathfrak{F}, \mathfrak{T}, \mathfrak{U} \rangle$, where $\mathfrak{F} = \langle W, < \rangle$ is a flow of time, $\mathfrak{T} = \langle U, \mathbb{I} \rangle$ a topological space, and $\mathfrak{U}$, a *valuation*, is a map associating with every spatial variable $p$ and every time point $w \in W$ a set $\mathfrak{U}(p,w) \subseteq U$—the 'space' occupied by $p$ at moment $w$; see Fig. 1. The valuation $\mathfrak{U}$ is inductively extended to arbitrary $\mathcal{PTL}[\mathcal{S}4_u]$-terms (i.e., spatial terms) in precisely the same way as for $\mathcal{S}4_u$, we only have to add a time point as a parameter:

$$\mathfrak{U}(\overline{\tau}, w) = U - \mathfrak{U}(\tau, w), \qquad \mathfrak{U}(\tau_1 \sqcap \tau_2, w) = \mathfrak{U}(\tau_1, w) \cap \mathfrak{U}(\tau_2, w), \qquad \mathfrak{U}(\mathbf{I}\tau, w) = \mathbb{I}\mathfrak{U}(\tau, w).$$

The truth-values of $\mathcal{PTL}[\mathcal{S}4_u]$-formulas are defined in the same way as for $\mathcal{PTL}$:

- $(\mathfrak{M}, w) \models \boxtimes\tau$ iff $\mathfrak{U}(\tau, w) = U$,
- $(\mathfrak{M}, w) \models \neg\varphi$ iff $(\mathfrak{M}, w) \not\models \varphi$,
- $(\mathfrak{M}, w) \models \varphi_1 \land \varphi_2$ iff $(\mathfrak{M}, w) \models \varphi_1$ and $(\mathfrak{M}, w) \models \varphi_2$,
- $(\mathfrak{M}, w) \models \varphi_1 \,\mathcal{U}\, \varphi_2$ iff there is $v > w$ such that $(\mathfrak{M}, v) \models \varphi_2$ and $(\mathfrak{M}, u) \models \varphi_1$ for all $u \in (w, v)$,
- $(\mathfrak{M}, w) \models \varphi_1 \,\mathcal{S}\, \varphi_2$ iff there is $v < w$ such that $(\mathfrak{M}, v) \models \varphi_2$ and $(\mathfrak{M}, u) \models \varphi_1$ for all $u \in (v, w)$.

And as in the pure temporal case, the operators $\Box_F$, $\Diamond_F$, $\bigcirc$ as well as their past counterparts can be defined in terms of $\mathcal{U}$ and $\mathcal{S}$.

A $\mathcal{PTL}[\mathcal{S}4_u]$-formula $\varphi$ is said to be *satisfiable* if there exists a tt-model $\mathfrak{M}$ such that $(\mathfrak{M}, w) \models \varphi$ for some time point $w$.

The following optimal complexity result will be obtained in Appendix B.1:

**Theorem 3.1.** *The satisfiability problem for $\mathcal{PTL}[\mathcal{S}4_u]$-formulas in tt-models based on arbitrary flows of time,* (*arbitrary*) *finite flows of time, $\langle \mathbb{N}, < \rangle$, $\langle \mathbb{Z}, < \rangle$, $\langle \mathbb{Q}, < \rangle$, or $\langle \mathbb{R}, < \rangle$, is* PSPACE*-complete.*

The proof of this theorem is based on the fact that the interaction between spatial and temporal components of $\mathcal{PTL}[\mathcal{S}4_u]$ is very restricted. In fact, for every $\mathcal{PTL}[\mathcal{S}4_u]$-formula $\varphi$ one can construct a $\mathcal{PTL}$-formula $\varphi^*$ by replacing every occurrence of a (spatial) subformula $\boxtimes\tau$ in $\varphi$ with a fresh propositional variable $p_\tau$. Then, given a $\mathcal{PTL}$-model $\mathfrak{N} = \langle \mathfrak{F}, \mathfrak{V} \rangle$ for $\varphi^*$ and a moment of time $w$, we take the set

$$\Phi_w = \{\boxtimes\tau \mid (\mathfrak{N}, w) \models p_\tau\} \;\cup\; \{\neg\boxtimes\tau \mid (\mathfrak{N}, w) \models \neg p_\tau\}$$

of spatial formulas. It is not hard to see that if $\Phi_w$ is satisfiable for every $w$ in $\mathfrak{F}$, then there is a tt-model satisfying $\varphi$ and based on the flow $\mathfrak{F}$. Now, to check whether $\varphi$ is satisfiable, it suffices to use a suitable nondeterministic algorithm (see, e.g., Sistla & Clarke, 1985; Reynolds, 2003, 2004) which guesses a $\mathcal{PTL}$-model for $\varphi^*$ and then, for each time point $w$, to check satisfiability of $\Phi_w$. This can be done using polynomial space in the length of $\varphi$.

Theorem 3.1 (together with Theorem 2.5) shows that all spatio-temporal logics of the form $\mathcal{PTL}[\mathcal{L}]$, for $\mathcal{L} \in \{\mathcal{RCC}\text{-}8, \mathcal{BRCC}\text{-}8, \mathcal{RC}, \mathcal{RC}^{max}\}$, are also PSPACE-complete over the standard flows of time.

Now let us consider temporalisations of spatial logics with the (NP-complete) fragment $\mathcal{PTL}_\Box$ of $\mathcal{PTL}$. By Theorems 2.4 and 3.1, both $\mathcal{PTL}_\Box[\mathcal{S}4_u]$ and $\mathcal{PTL}_\Box[\mathcal{RC}^{max}]$ are PSPACE-complete. However, for simpler (NP-complete) spatial components we obtain a better result:

**Theorem 3.2.** *The satisfiability problem for $\mathcal{PTL}_\Box[\mathcal{RC}]$-formulas in tt-models based on each of the classes of flows of time mentioned in Theorem 3.1 is* NP*-complete.*

The proof is essentially the same as that of Theorem 3.1, but now nondeterministic polynomial-time algorithms for the component logics are available. It follows from Theorem 3.2 that $\mathcal{PTL}_\Box[\mathcal{RCC}\text{-}8]$ and $\mathcal{PTL}_\Box[\mathcal{BRCC}\text{-}8]$ are NP-complete as well.

### 3.2 'Maximal' Combinations with (PC) and (OC)

As we saw in the previous section, the computational complexity of spatio-temporal logics without (OC) is the maximum of the complexity of their components, which reflects the very limited interaction between spatial and temporal operators in languages without any means of expressing (OC).

A 'maximalist' approach to constructing spatio-temporal logics capable of capturing both (PC) and (OC) is to allow unrestricted applications of the Booleans, the topological and the temporal operators to form spatio-temporal terms.

Denote by $\mathcal{PTL} \times \mathcal{S}4_u$ the spatio-temporal language given by the following definition:

$$\begin{array}{lcl} \tau & ::= & p \quad | \quad \overline{\tau} \quad | \quad \tau_1 \sqcap \tau_2 \quad | \quad \mathbf{I}\tau \quad | \quad \tau_1 \,\mathcal{U}\, \tau_2 \quad | \quad \tau_1 \,\mathcal{S}\, \tau_2, \\ \varphi & ::= & \boxdot\tau \quad | \quad \neg\varphi \quad | \quad \varphi_1 \wedge \varphi_2 \quad | \quad \varphi_1 \,\mathcal{U}\, \varphi_2 \quad | \quad \varphi_1 \,\mathcal{S}\, \varphi_2. \end{array}$$

Expressions of the form $\tau$ will be called *spatio-temporal terms*. Unlike the previous section, these terms can be time-dependent. The definition of expressions of the form $\varphi$ is the same as for $\mathcal{PTL}[\mathcal{S}4_u]$; they will be called *$\mathcal{PTL} \times \mathcal{S}4_u$-formulas*. All of the languages from Section 3.1, including $\mathcal{PTL}[\mathcal{S}4_u]$, are clearly fragments of $\mathcal{PTL} \times \mathcal{S}4_u$.

As before, we can introduce the temporal operators $\Box_F$, $\Diamond_F$, $\bigcirc$ as well as their past counterparts applicable to formulas. Moreover, these operators can now be used to form spatio-temporal terms: for example,

$$\Diamond_F\tau = \top \,\mathcal{U}\, \tau, \qquad \Box_F\tau = \neg\Diamond_F\overline{\tau} \quad \text{and} \quad \bigcirc\tau = \bot \,\mathcal{U}\, \tau,$$

where $\bot$ denotes the empty set and $\top$ the whole space.

Spatio-temporal formulas are supposed to represent propositions speaking about moving spatial objects represented by spatio-temporal terms. The truth-values of propositions in spatio-temporal structures can vary in time, but do not depend on points of spaces—they are defined in precisely the same way as in the case of $\mathcal{PTL}[\mathcal{S}4_u]$. But how to understand temporalised terms?

The meaning of $\bigcirc\tau$ should be clear: at moment $w$, it denotes the space occupied by $\tau$ at the next moment $w+1$ (see Fig. 2). For example, we can write

$$\diamondsuit(\mathbf{I}\lceil \mathit{Cyclone}\rceil \sqcap \mathbf{I}\lceil \bigcirc \mathit{Cyclone}\rceil) \tag{C}$$

to say that regions *Cyclone* and $\bigcirc$*Cyclone* overlap (thereby formalising sentence (C) from the introduction). The formula

$$\mathsf{EQ}(\bigcirc\bigcirc EU, EU \sqcup Romania \sqcup Bulgaria) \tag{F}$$

says that in two years the EU (as it is today) will be extended with Romania and Bulgaria. Note that $\bigcirc\bigcirc\mathsf{EQ}(EU, EU \sqcup Romania \sqcup Bulgaria)$ has a different meaning because the EU may expand or shrink in a year. It is also not hard to formalise sentences (D), (E) and (H) from the introduction:

$$\mathsf{EQ}(\bigcirc X, Y) \to \neg\mathsf{EQ}(Y, \bigcirc Y), \tag{D}$$

$$\Box_F \mathsf{EQ}(\bigcirc Europe, Europe), \tag{E}$$

$$\mathsf{EQ}(Earth, W \sqcup L) \wedge \mathsf{EC}(W, L) \wedge \mathsf{P}(W, \bigcirc W) \to \mathsf{P}(\bigcirc L, L), \tag{H}$$

where $\mathsf{P}(X, Y)$—'$X$ is a part of $Y$'—denotes the disjunction of $\mathsf{EQ}(X, Y)$, $\mathsf{TPP}(X, Y)$ and $\mathsf{NTPP}(X, Y)$.

The intended interpretation of terms of the form $\Diamond_F \tau$, $\Box_F \tau$ (and their past counterparts) is a bit more sophisticated. It reflects the standard temporal meanings of propositions '$\Diamond_F\, x \in \tau$' and '$\Box_F\, x \in \tau$,' for all points $x$ in the topological space:

- at moment $w$, term $\Diamond_F \tau$ is interpreted as the union of all spatial extensions of $\tau$ at moments $v > w$;
- at moment $w$, term $\Box_F \tau$ is interpreted as the intersection of all spatial extensions of $\tau$ at moments $v > w$.

For example, consider Fig. 2 with moving spatial object $X$ depicted on it at three consecutive moments of time (it does not change after $t + 2$). Then $\Diamond_F X$ at $t$ is the union of $\bigcirc X$ and $\bigcirc\bigcirc X$ at $t$ and $\Box_F X$ at $t$ is the intersection of $\bigcirc X$ and $\bigcirc\bigcirc X$ at $t$ (i.e., $\bigcirc X$).

As another example, take the spatial object *Rain*. Then

- $\Diamond_F$*Rain* at moment $w$ occupies the space where it will be raining at *some* time points $v > w$ (which may be different for different places). $\Box_F$*Rain* at $w$ occupies the space where it will *always* be raining after $w$.
- $\Box_F \Diamond_F$*Rain* at $w$ is the space where it will be raining ever and ever again after $w$, while $\Diamond_F \Box_F$*Rain* comprises all places where it will always be raining starting from some future moments of time.

This interpretation shows how to formalise sentence (G) from the introduction:

$$\mathsf{P}(England, \Box_F \Diamond_F Rain). \tag{G}$$

Now, what can be the meaning of $Rain\,\mathcal{U}\,Snow$? Similarly to the readings of $\Box_F \tau$ and $\Diamond_F \tau$ above, we adopt the following definition:

- at moment $w$, the spatial extension of $\tau_1 \mathcal{U} \tau_2$ consists of those points $x$ of the topological space for which there is $v > w$ such that $x$ belongs to $\tau_2$ at moment $v$ and $x$ is in $\tau_1$ at all $u$ whenever $w < u < v$.

The past counterpart of $\mathcal{U}$—i.e., the operator 'since' $\mathcal{S}$—can be used to say that the part of Russia that has been remaining Russian since 1917 is not connected to the part of Germany (Königsberg) that became Russian after the Second World War (Kaliningrad):

$$\mathsf{DC}(\mathit{Russia}\,\mathcal{S}\,\mathit{Russian_Empire}, \mathit{Russia}\,\mathcal{S}\,\mathit{Germany}).$$

The models $\mathfrak{M} = \langle \mathfrak{F}, \mathfrak{T}, \mathfrak{U} \rangle$ for $\mathcal{PTL} \times \mathcal{S}4_u$ are precisely the same topological temporal models we introduced for $\mathcal{PTL}[\mathcal{S}4_u]$. However, now we need additional clauses defining extensions of spatio-temporal terms:

- $\mathfrak{U}(\tau_1 \,\mathcal{U}\, \tau_2, w) = \bigcup\limits_{v>w} \Big(\mathfrak{U}(\tau_2, v) \cap \bigcap\limits_{u\in(w,v)} \mathfrak{U}(\tau_1, u)\Big),$
- $\mathfrak{U}(\tau_1 \,\mathcal{S}\, \tau_2, w) = \bigcup\limits_{v<w} \Big(\mathfrak{U}(\tau_2, v) \cap \bigcap\limits_{u\in(v,w)} \mathfrak{U}(\tau_1, u)\Big).$

Then we also have:

$$\mathfrak{U}(\Diamond_F \tau, w) = \bigcup_{v>w} \mathfrak{U}(\tau, v) \qquad \text{and} \qquad \mathfrak{U}(\Box_F \tau, w) = \bigcap_{v>w} \mathfrak{U}(\tau, v),$$

and, for discrete $\mathfrak{F}$,

$$\mathfrak{U}(\bigcirc\tau, w) = \mathfrak{U}(\tau, w+1).$$

The truth-values of $\mathcal{PTL} \times \mathcal{S}4_u$-formulas are computed in precisely the same way as in the case of $\mathcal{PTL}[\mathcal{S}4_u]$. A $\mathcal{PTL} \times \mathcal{S}4_u$-formula $\varphi$ is called *satisfiable* if there exists a tt-model $\mathfrak{M}$ such that $(\mathfrak{M}, w) \models \varphi$ for some time point $w$.

At first sight it may appear that the computational properties of the constructed logic should not be too bad—after all, its spatial and temporal components are PSPACE-complete. It turns out, however, that this is not the case:

**Theorem 3.3.** *The satisfiability problem for $\mathcal{PTL} \times \mathcal{S}4_u$-formulas in tt-models based on the flows of time $\langle \mathbb{N}, < \rangle$ or $\langle \mathbb{Z}, < \rangle$ is undecidable.*

Without going into details of the proof of this theorem, one might immediately conjecture that it is the use of the *infinitary* operators $\mathcal{U}$, $\Box_F$ and $\Diamond_F$ in the construction of spatio-temporal terms that makes the logic 'over-expressive.' Moreover, the whole idea of topological temporal models based on *infinite* flows of time may look counterintuitive in the context of spatio-temporal representation and reasoning (unlike, say, models used to represent the behaviour of reactive computer systems).

There are different approaches to avoid infinity in tt-models. The most radical one is to allow only finite flows of time. A more cautious approach is to impose the following *finite change assumption* on models (based on infinite flows of time):

**FCA** *No term can change its spatial extension infinitely often.*

This means that under **FCA** we consider only those valuations $\mathfrak{U}$ in tt-models $\langle \mathfrak{F}, \mathfrak{T}, \mathfrak{U} \rangle$ that satisfy the following condition: for every spatio-temporal term $\tau$, there are pairwise disjoint intervals $I_1, \dots, I_n$ of $\mathfrak{F} = \langle W, < \rangle$ such that $W = I_1 \cup \dots \cup I_n$ and the state of $\tau$ remains constant on each $I_j$, i.e., $\mathfrak{U}(\tau, u) = \mathfrak{U}(\tau, v)$ for any $u, v \in I_j$. It turns out, however, that in the case of *discrete* flows of time **FCA** does not give us anything new as compared to arbitrary finite flows of time. More precisely, one can easily show that the satisfiability problem for $\mathcal{PTL} \times \mathcal{S}4_u$-formulas in tt-models satisfying **FCA** and based on $\langle \mathbb{N}, < \rangle$ or $\langle \mathbb{Z}, < \rangle$ is polynomially reducible to satisfiability in tt-models based on finite flows of time, and the other way round. Note also that for the flows of time mentioned above, **FCA** can be captured by the formulas $\Diamond_F \Box_F \mathsf{EQ}(\tau, \bigcirc_F \tau)$ (and its past counterpart for $\langle \mathbb{Z}, < \rangle$), for every spatio-temporal term $\tau$.

A more 'liberal' way of reducing infinite unions and intersections to finite ones is to adopt the *finite state assumption*:

**FSA** *Every term may have only finitely many possible states* (*although it may change its states infinitely often*).

Say that a tt-model $\langle \mathfrak{F}, \mathfrak{T}, \mathfrak{U} \rangle$ satisfies **FSA** if, for every spatio-temporal term $\tau$, there are finitely many sets $A_1, \dots, A_m$ in the space $\mathfrak{T}$ such that $\{\mathfrak{U}(\tau, w) \mid w \in W\} = \{A_1, \dots, A_m\}$. Such models can be used, for instance, to capture periodic fluctuations due to season or climate changes, say, a daily tide. Similarly to **FCA** finitising the flow of time, **FSA** virtually makes the underlying topological space finite. The following proposition will be proved in Appendix B:

**Proposition 3.4.** *A $\mathcal{PTL} \times \mathcal{S}4_u$-formula is satisfiable in a tt-model with* **FSA** *and based a flow of time $\mathfrak{F}$ iff it is satisfiable in a tt-model based on $\mathfrak{F}$ and a finite* (*Aleksandrov*) *topological space.*

Unfortunately, none of these approaches works for $\mathcal{PTL} \times \mathcal{S}4_u$—we still have:

**Theorem 3.5.** (i) *The satisfiability problem for $\mathcal{PTL} \times \mathcal{S}4_u$-formulas in tt-models based on* (*arbitrary*) *finite flows of time is undecidable.*

(ii) *The satisfiability problem for $\mathcal{PTL} \times \mathcal{S}4_u$-formulas in tt-models based on the flows of time $\langle \mathbb{N}, < \rangle$ or $\langle \mathbb{Z}, < \rangle$ and satisfying* **FSA** *is undecidable.*

The next-time operator $\bigcirc$ does not look so 'harmful' as the infinitary $\mathcal{U}$, $\Box_F$, $\Diamond_F$, and still can capture some aspects of (OC) (see formulas (C), (D), (F) and (G) above). So let us consider the fragment $\mathcal{PTL} \circ \mathcal{S}4_u$ of $\mathcal{PTL} \times \mathcal{S}4_u$ with spatio-temporal terms of the form:

$$\tau \quad ::= \quad p \quad | \quad \overline{\tau} \quad | \quad \tau_1 \sqcap \tau_2 \quad | \quad \mathbf{I}\tau \quad | \quad \bigcirc\tau.$$

In other words, $\mathcal{PTL} \circ \mathcal{S}4_u$ does not allow applications of temporal operators different from $\bigcirc$ to form spatio-temporal terms (but they are still available as formula constructors). This means that we can compare the states of a spatial object $X$ over a bounded set of time points only: for any time point $t$ and any natural numbers $n, m \geq 0$, we can compare at $t$ the state of $X$ at $t + n$ with its state at $t + m$.

This fragment is definitely less expressive than full $\mathcal{PTL} \times \mathcal{S}4_u$. For instance, according to Lemma B.1, $\mathcal{PTL} \circ \mathcal{S}4_u$-formulas do not distinguish between arbitrary tt-models and those based on Aleksandrov topological spaces—we will call them *Aleksandrov* tt-models. On the other hand, the set of $\mathcal{PTL} \times \mathcal{S}4_u$-formulas satisfiable in Aleksandrov models is a proper subset of those satisfiable in arbitrary tt-models. Consider, for example, the $\mathcal{PTL} \times \mathcal{S}4_u$-formula

$$\boxtimes(\Box_F \mathbf{I}p \ \sqsubset \ \mathbf{I}\Box_F p).$$

One can readily see that it is true in every Aleksandrov tt-model, but its negation can be satisfied in a topological model. For it suffices to take the flow $\mathfrak{F} = \langle \mathbb{N}, < \rangle$ and the topology $\mathfrak{T} = \langle \mathbb{R}, \mathbb{I} \rangle$ with the standard interior operator $\mathbb{I}$ on the real line, select a sequence $X_n$ of open sets such that $\bigcap_{n \in \mathbb{N}} X_n$ is not open, e.g., $X_n = (-1/n, 1/n)$, and put $\mathfrak{U}(p, n) = X_n$.

However, even this seemingly weak interaction between topological and temporal operators turns out to be dangerous:

**Theorem 3.6.** *The satisfiability problem for $\mathcal{PTL} \circ \mathcal{S}4_u$-formulas in tt-models based on the flows of time $\langle \mathbb{N}, < \rangle$ or $\langle \mathbb{Z}, < \rangle$ is undecidable. It is undecidable as well for tt-models satisfying* **FSA** *or based on (arbitrary) finite flows of time.*

Theorem 2.6 might suggest considering the fragment $\mathcal{PTL}_\Box \times \mathcal{S}4_u$ with $\Box_F$ and its past counterpart $\Box_P$ as the only temporal primitives applicable both to formulas and terms:

$$\tau \quad ::= \quad p \quad | \quad \overline{\tau} \quad | \quad \tau_1 \sqcap \tau_2 \quad | \quad \mathbf{I}\tau \quad | \quad \Box_F \tau \quad | \quad \Box_P \tau,$$

$$\varphi \quad ::= \quad \boxed{\forall}\tau \quad | \quad \neg\varphi \quad | \quad \varphi_1 \land \varphi_2 \quad | \quad \Box_F \varphi \quad | \quad \Box_P \varphi.$$

Yet again the result is 'negative:'

**Theorem 3.7.** *The satisfiability problem for $\mathcal{PTL}_\Box \times \mathcal{S}4_u$-formulas in tt-models (with or without* **FSA***) based on the flows of time $\langle \mathbb{N}, < \rangle$ or $\langle \mathbb{Z}, < \rangle$ is undecidable. It is undecidable as well for tt-models based on (arbitrary) finite flows of time.*

These undecidability results (the strongest ones, Theorems 3.6 and 3.7, to be more precise) will be proved in Appendix B.2 by a reduction of Post's correspondence problem which is known to be undecidable (Post, 1946). As we will see from the proofs, these theorems actually hold for the 'future fragments' of the corresponding languages.

### 3.3 Decidable Spatio-Temporal Logics with (PC) and (OC)

An important lesson we learn from (the proofs of) the 'negative' results of Section 3.2 is that full $\mathcal{S}4_u$ is too expressive for computationally well-behaved combinations with fragments of $\mathcal{PTL}$. On the other hand, as was said in Section 2.1.2, qualitative spatial representation and reasoning often requires extensions of spatial variables to be regular closed (i.e., regions). This restriction is very important for constructing decidable spatio-temporal logics with (PC) and (OC). First, the undecidability proofs from Appendix B.2 do not go through in this case. And second, as will be shown below, decidable combinations of $\mathcal{PTL}$ and some of the fragments of $\mathcal{S}4_u$ introduced in Section 2.1 do exist. In fact, we will construct a hierarchy of decidable spatio-temporal logics of different computational complexity by imposing various restrictions on regions themselves, the ways they can be compared, and the interactions between spatial and temporal constructors.

We begin by considering the simplest combination of $\mathcal{PTL}$ and $\mathcal{RCC}$-8 capturing (PC) and (OC). This logic called $\mathcal{PTL} \circ \mathcal{RCC}$-8 (it was introduced under the name $\mathcal{ST}_1^-$ by Wolter and Zakharyaschev, 2002) operates with *spatio-temporal region terms* of the form

$$\varrho \quad ::= \quad \mathbf{CI}p \quad | \quad \mathbf{CI}\bigcirc\varrho.$$

To relate these terms, we are allowed to use the eight binary predicates of $\mathcal{RCC}$-8; then arbitrary temporal operators and Boolean connectives can be applied to produce $\mathcal{PTL} \circ \mathcal{RCC}$-8 *formulas.* Typical examples of such formulas are (A), (B), (D) and (E) above. Note that (C) can be regarded as a $\mathcal{PTL} \circ \mathcal{RCC}$-8 formula as well (two regions overlap iff they are neither disconnected nor externally connected). On the other hand, (F), (H) and (G) are not $\mathcal{PTL} \circ \mathcal{RCC}$-8 formulas because the first two use the $\sqcup$ operation on region terms and (G) uses temporal operators $\Box_F$ and $\Diamond_F$ on region terms.

As before, $\mathcal{PTL} \circ \mathcal{RCC}$-8 formulas are interpreted in topological temporal models (or tt-models). However, only discrete flows of time do make sense for this language. Although the interaction between topological and temporal operators is similar to that in $\mathcal{PTL} \circ \mathcal{S}4_u$ (clearly, $\mathcal{PTL} \circ \mathcal{RCC}$-8 is a fragment of $\mathcal{PTL} \circ \mathcal{S}4_u$), we have the following rather unexpected and encouraging result:

**Theorem 3.8.** *The satisfiability problem for $\mathcal{PTL} \circ \mathcal{RCC}$-8 formulas in tt-models based on $\langle \mathbb{N}, < \rangle$, $\langle \mathbb{Z}, < \rangle$ or (arbitrary) finite flows of time is* PSPACE*-complete.*

This theorem will be proved in Appendix C.5. The idea of the proof is similar to that of Theorem 3.1: we consider the spatial and the temporal parts of a given formula separately. However, to take into account the interaction between these parts, we use the so-called 'completion property' of $\mathcal{RCC}$-8 (cf. Balbiani & Condotta, 2002) with respect to a certain class $\mathfrak{C}$ of models: given a satisfiable set $\Phi$ of $\mathcal{RCC}$-8 formulas and a model in $\mathfrak{C}$ satisfying a subset of $\Phi$, one can extend this 'partial' model to a model in $\mathfrak{C}$ satisfying the whole $\Phi$.

What happens if we extend the expressive power of the spatial component by allowing Boolean operators on spatio-temporal region terms, i.e., jump from $\mathcal{RCC}$-8 to $\mathcal{BRCC}$-8? Define *spatio-temporal Boolean region terms* by taking

$$\varrho \quad ::= \quad \mathbf{CI}p \quad | \quad \mathbf{CI}\overline{\varrho} \quad | \quad \mathbf{CI}(\varrho_1 \sqcap \varrho_2) \quad | \quad \mathbf{CI}\bigcirc\varrho.$$

Denote by $\mathcal{PTL} \circ \mathcal{BRCC}$-8 the language obtained from $\mathcal{PTL} \circ \mathcal{RCC}$-8 by allowing spatio-temporal Boolean region terms as arguments of the $\mathcal{RCC}$-8 predicates (this language was called $\mathcal{ST}_1$ by Wolter and Zakharyaschev, 2002). Formulas (A)–(F) and (H) belong to $\mathcal{PTL} \circ \mathcal{BRCC}$-8, but (G) uses the $\Box_F$ and $\Diamond_F$ operators on regions and so is not in $\mathcal{PTL} \circ \mathcal{BRCC}$-8.

Now, another surprise is that the replacement of $\mathcal{RCC}$-8 with $\mathcal{BRCC}$-8 in our temporal context results in an exponential jump of the computational complexity (remember that both $\mathcal{RCC}$-8 and $\mathcal{BRCC}$-8 are NP-complete):

**Theorem 3.9.** *The satisfiability problem for $\mathcal{PTL} \circ \mathcal{BRCC}$-8 formulas in tt-models based on the flows of time $\langle \mathbb{N}, < \rangle$ or $\langle \mathbb{Z}, < \rangle$ is* EXPSPACE*-complete. It is* EXPSPACE*-complete as well for models satisfying* **FSA** *or based on (arbitrary) finite flows of time.*

The EXPSPACE upper bound (see Appendix C.3) is proved by a polynomial embedding of $\mathcal{PTL} \circ \mathcal{BRCC}$-8 into the one-variable fragment $\mathcal{QTL}^1$ of first-order temporal logic, which is known to be EXPSPACE-complete (Hodkinson, Kontchakov, Kurucz, Wolter, & Zakharyaschev, 2003). To construct this embedding, we first show that $\mathcal{PTL} \circ \mathcal{BRCC}$-8 is complete with respect to Aleksandrov tt-models. In fact, we prove that every satisfiable formula of the more expressive logic $\mathcal{PTL} \circ \mathcal{S}4_u$ introduced in Section 3.2 can be satisfied in an Aleksandrov tt-model (see Lemma B.1 and the discussion above). Lemma C.1 then shows that to satisfy a $\mathcal{PTL} \circ \mathcal{BRCC}$-8 formula, it suffices to take an Aleksandrov tt-model based on a partial order of depth 1. By Lemma C.2, the width of the partial order can be bounded by 2 (just as in the case of $\mathcal{BRCC}$-8), and therefore unions of forks (or 2-brooms) are enough to satisfy $\mathcal{PTL} \circ \mathcal{BRCC}$-8 formulas. These Aleksandrov tt-models based on unions of forks can be encoded by means of unary predicates of $\mathcal{QTL}^1$.

The EXPSPACE lower bound is proved in Appendix C.1 by encoding the corridor tiling problem. It can also be established by a direct polynomial embedding of $\mathcal{QTL}^1$ into $\mathcal{PTL} \circ \mathcal{BRCC}$-8. To illustrate the idea, consider the $\mathcal{QTL}^1$-formula $\forall x\,(P(x) \lor \bigcirc P(x))$ saying that, for every point of the space, either it is in $P$ now or will be there tomorrow. The same statement can be expressed in $\mathcal{PTL} \circ \mathcal{BRCC}$-8 by the formula $\mathsf{EQ}(P \sqcup \bigcirc P, \overline{E}) \land \mathsf{DC}(E, E)$, where the last conjunct makes $E$ empty.

Now let us make one more step in space and extend $\mathcal{BRCC}$-8 to $\mathcal{RC}$, thus obtaining the spatio-temporal language $\mathcal{PTL} \circ \mathcal{RC}$ with the following syntax:

$$\begin{aligned}
\varrho \quad &::= \quad \mathbf{CI}p \quad | \quad \mathbf{CI}\overline{\varrho} \quad | \quad \mathbf{CI}(\varrho_1 \sqcap \varrho_2) \quad | \quad \mathbf{CI}\bigcirc\varrho,\\
\tau \quad &::= \quad \varrho \quad | \quad \mathbf{I}\varrho \quad | \quad \overline{\tau} \quad | \quad \tau_1 \sqcap \tau_2,\\
\varphi \quad &::= \quad \boxdot\tau \quad | \quad \neg\varphi \quad | \quad \varphi_1 \land \varphi_2 \quad | \quad \varphi_1 \,\mathcal{U}\, \varphi_2 \quad | \quad \varphi_2 \,\mathcal{S}\, \varphi_2.
\end{aligned}$$

The reader should not be surprised now (although the authors were) that the extra expressivity results in one more exponential gap:

**Theorem 3.10.** *The satisfiability problem for $\mathcal{PTL} \circ \mathcal{RC}$-formulas in tt-models based on the flows of time $\langle \mathbb{N}, < \rangle$ or $\langle \mathbb{Z}, < \rangle$ is* 2EXPSPACE*-complete. It is* 2EXPSPACE*-complete as well for models satisfying* **FSA** *or based on (arbitrary) finite flows of time.*

The lower bound is established in Appendix C.1 and the upper bound in Appendix C.2.

Perhaps, it is proper time now to have a closer look at the emerging landscape. What exactly causes these exponential 'jumps'? Can we locate more precise borders in the ladder PSPACE–EXPSPACE–2EXPSPACE?

By analysing the proof of Theorem 3.8 (see Appendix C.5), we note that not so much can be added to $\mathcal{RCC}$-8. In fact, the maximal spatio-temporal logic (denoted by $\mathcal{PTL} \circ \mathcal{RC}_2$) for which this proof goes through is based on spatio-temporal terms of the form

$$\begin{aligned}
\varrho \quad &::= \quad \mathbf{CI}p \quad | \quad \mathbf{CI}\bigcirc\varrho,\\
\theta \quad &::= \quad \varrho \quad | \quad \mathbf{I}\varrho \quad | \quad \overline{\varrho} \quad | \quad \overline{\mathbf{I}\varrho},\\
\tau \quad &::= \quad \overline{\theta_1 \sqcap \theta_2}.
\end{aligned}$$

On the other hand, even the addition of predicates of the form $\mathsf{EQ}(X, Y \sqcup Z)$ is enough to make the logic EXPSPACE-hard (see Remark C.3). Thus, $\mathcal{PTL} \circ \mathcal{RCC}$-8 (or rather its extension $\mathcal{PTL} \circ \mathcal{RC}_2$) is located pretty close to the border between PSPACE and EXPSPACE spatio-temporal logics.

The following fragment $\mathcal{RC}^-$ of $\mathcal{RC}$ indicates where the border between EXPSPACE and 2EXPSPACE may lie:

$$\begin{aligned}
\varrho \quad &::= \quad \text{Boolean region terms},\\
\delta \quad &::= \quad \varrho \quad | \quad \overline{\sigma},\\
\sigma \quad &::= \quad \mathbf{I}\varrho \quad | \quad \overline{\delta} \quad | \quad \sigma_1 \sqcap \sigma_2,\\
\tau \quad &::= \quad \overline{\delta_1 \sqcap \cdots \sqcap \delta_m} \quad | \quad \overline{\delta \sqcap \sigma} \quad | \quad \overline{\sigma}.
\end{aligned}$$

Intuitively, the $\delta$ and the $\sigma$ are spatial terms interpreted by regular closed and regular open[4] sets, respectively (the interior of a region is regular open, the complement of a regular closed set is regular open (and vice versa), regular closed sets are closed under unions and regular open ones are closed under intersections). Thus, $\delta$ can be regarded as a generalisation of region terms and $\sigma$ as a generalisation of the interiors of regions. In other words, $\mathcal{RC}^-$ is the fragment of $\mathcal{RC}$ in which only the following ways of relating regions are available:

- there is a point where some regions meet;
- a region intersects the interior of another one;
- the interior of a region is not empty.

It is readily checked that $\mathcal{BRCC}$-8 is a fragment of $\mathcal{RC}^-$. Moreover, it is a proper fragment because (4) belongs to the latter but not to the former. The formula

$$\boxtimes\big((\lceil NorthKorea\rceil \sqcap \lceil SouthKorea\rceil) \sqsubset \lceil DmZone\rceil\big) \tag{8}$$

(saying that the demilitarised zone between the North Korea and the South Korea consists of the border between them along with some adjacent territories) shows that $\mathcal{RC}^-$ is a proper subset of $\mathcal{RC}$:

$$\mathcal{BRCC}\text{-}8 \quad \subsetneqq \quad \mathcal{RC}^- \quad \subsetneqq \quad \mathcal{RC}.$$

Although $\mathcal{RC}^-$ extends $\mathcal{BRCC}$-8, it gives rise to the spatio-temporal logic of the same computational complexity:

**Theorem 3.11.** *The satisfiability problem for $\mathcal{PTL} \circ \mathcal{RC}^-$-formulas in tt-models based on the flows of time $\langle \mathbb{N}, < \rangle$ or $\langle \mathbb{Z}, < \rangle$ is* EXPSPACE-*complete. It is* EXPSPACE-*complete as well for models satisfying* **FSA** *or based on* (*arbitrary*) *finite flows of time.*

The lower bound follows immediately from Theorem 3.9 and the proof of the upper bound is similar to that of Theorem 3.9 (see Appendix C.3). Again, due to the restriction on possible ways of relating regions, we can polynomially bound the width $n$ of $n$-brooms that are required to satisfy $\mathcal{PTL}\circ\mathcal{RC}^-$-formulas (cf. Lemma C.2). In fact, we need formulas similar to (8) in order to increase complexity to 2EXPSPACE.

The constructed hierarchy of decidable spatio-temporal logics still leaves at least one important question: do there exist decidable spatio-temporal logics that allow applications of the temporal operators $\mathcal{U}$, $\Box_F$, $\Diamond_F$ to region terms and what is their complexity? Consider the languages $\mathcal{PTL} \times \mathcal{L}$, for $\mathcal{L} \in \{\mathcal{BRCC}\text{-}8, \mathcal{RC}^-, \mathcal{RC}\}$, which differ from $\mathcal{PTL} \circ \mathcal{L}$ only in the definition of spatio-temporal *region* terms:

$$\varrho \quad ::= \quad \mathbf{CI}p \quad | \quad \mathbf{CI}\overline{\varrho} \quad | \quad \mathbf{CI}(\varrho_1 \sqcap \varrho_2) \quad | \quad \mathbf{CI}(\varrho_1 \,\mathcal{U}\, \varrho_2) \quad | \quad \mathbf{CI}(\varrho_1 \,\mathcal{S}\, \varrho_2).$$

The following two theorems provide a positive (though partial) answer to this question:

**Theorem 3.12.** *The satisfiability problem for $\mathcal{PTL} \times \mathcal{BRCC}$-8 and $\mathcal{PTL} \times \mathcal{RC}^-$-formulas in tt-models based on $\langle \mathbb{N}, < \rangle$ or $\langle \mathbb{Z}, < \rangle$ and satisfying* **FSA**, *or based on* (*arbitrary*) *finite flows of time is* EXPSPACE-*complete.*

4. Remember that a set $X$ is *regular open* if $\mathbb{IC}X = X$.

**Theorem 3.13.** *The satisfiability problem for $\mathcal{PTL} \times \mathcal{RC}$-formulas in tt-models based on $\langle \mathbb{N}, < \rangle$ or $\langle \mathbb{Z}, < \rangle$ and satisfying* **FSA***, or based on (arbitrary) finite flows of time is* 2EXPSPACE-*complete.*

The upper bounds mentioned in these two theorems are proved in Appendices C.3 and C.2, respectively. The lower bounds follow from the results for $\mathcal{PTL} \circ \mathcal{BRCC}$-8 (Theorem 3.9) and $\mathcal{PTL} \circ \mathcal{RC}^-$ (Theorem 3.10).

To appreciate the following theorem, the reader should recall that both $\mathcal{PTL}_\Box$ and $\mathcal{RC}^-$ are NP-complete:

**Theorem 3.14.** *The satisfiability problem for $\mathcal{PTL}_\Box \times \mathcal{BRCC}$-8 and $\mathcal{PTL}_\Box \times \mathcal{RC}^-$-formulas in tt-models based on $\langle \mathbb{N}, < \rangle$ or $\langle \mathbb{Z}, < \rangle$ and satisfying* **FSA***, or based on (arbitrary) finite flows of time is* EXPSPACE-*complete.*

Actually it is a consequence of the EXPSPACE-hardness of $\mathcal{QTL}^1$ with the sole temporal operator $\Box_F$ (see Hodkinson et al., 2003).

Unfortunately, very little is known about the complexity of our spatio-temporal languages interpreted in tt-models based on dense or arbitrary flows of time. In fact, the only result we know of can be proved using the recent work (Hodkinson, 2004; Hodkinson et al., 2003):

**Theorem 3.15.** *The satisfiability problem for $\mathcal{PTL} \times \mathcal{BRCC}$-8 and $\mathcal{PTL} \times \mathcal{RC}^-$-formulas in tt-models satisfying* **FSA** *and based on $\langle \mathbb{Q}, < \rangle$, $\langle \mathbb{R}, < \rangle$ or arbitrary flows of time belongs to* 2EXPTIME *and is* EXPSPACE-*hard.*

## 4. Conclusion

We have provided an in-depth analysis of the computational complexity of various spatio-temporal logics interpreted in Cartesian products of flows of time and topological spaces. Some of these results are collected in Table 1. The design of the languages was driven by the idea to cover the most basic features of spatio-temporal hybrids combining standard logics of time and mereotopology, with the aim being to see how complex reasoning with these hybrids could be. We did not try to fine-tune the languages for real-world applications. On the contrary, we tried to keep them as 'pure' and representative as possible and determine computational challenges which *any* multi-dimensional approach to reasoning about space and time would face. With this research objective in mind, we discuss now some conclusions that can be drawn from Table 1.

The conclusion to be drawn from the undecidability results is easy: do not try to implement a sound, complete and terminating algorithm which is supposed to decide the satisfiability problem for $\mathcal{PTL} \times \mathcal{S}4_u$, $\mathcal{PTL} \circ \mathcal{S}4_u$ or $\mathcal{PTL}_\Box \times \mathcal{S}4_u$—you will never succeed. If decision procedures are required, then alternative languages have to be devised.

The interpretation of the complexity results for decidable logics is not so transparent: it is well-known that such results do not provide us with immediate conclusions regarding the behaviour of implemented systems. For example, sometimes algorithms running in exponential time in the worst-case perform better on *practical problems* than worst-case optimal algorithms that run in polynomial time. Indeed, the complexity results should be analysed *together with their proofs*—if significant conclusions are required (cf. Nebel, 1996).

<table>
<tr><th rowspan="2">language</th><th rowspan="2">flow of time</th><th colspan="5">spatial component $\mathcal{L}$</th></tr>
<tr><th>$\mathcal{RCC}$-8</th><th>$\mathcal{BRCC}$-8</th><th>$\mathcal{RC}$</th><th>$\mathcal{RC}^{max}$</th><th>$\mathcal{S}4_u$</th></tr>
<tr><td>$\mathcal{L}$</td><td>n/a</td><td colspan="2">NP</td><td>NP<br>(Thm. 2.2)</td><td>PSPACE<br>(Thm. 2.4)</td><td>PSPACE<br>(Thm. 2.1)</td></tr>
<tr><td>$\mathcal{PTL}_{\Box}[\mathcal{L}]$</td><td>$\mathbb{N}, \mathbb{Z}, \mathbb{Q}, \mathbb{R}$, finite or arbitrary</td><td colspan="3">NP<br>(Thm. 3.2)</td><td colspan="2">PSPACE<br>(Thm. 3.1)</td></tr>
<tr><td>$\mathcal{PTL}[\mathcal{L}]$</td><td>$\mathbb{N}, \mathbb{Z}, \mathbb{Q}, \mathbb{R}$, finite or arbitrary</td><td colspan="5">PSPACE<br>(Thm. 3.1)</td></tr>
<tr><td rowspan="2">$\mathcal{PTL} \circ \mathcal{L}$</td><td>$\mathbb{N}, \mathbb{Z}$</td><td>PSPACE<br>(Thm. 3.8)</td><td rowspan="2">EXPSPACE<br>(Thm. 3.9)</td><td rowspan="2">2EXPSPACE<br>(Thm. 3.10)</td><td rowspan="2">?</td><td rowspan="2">undecidable<br>(Thm. 3.6)</td></tr>
<tr><td>finite or $\mathbb{N}, \mathbb{Z}$+**FSA**</td><td>$\leq$ EXPSPACE<br>$\geq$ PSPACE</td></tr>
<tr><td rowspan="3">$\mathcal{PTL}_{\Box} \times \mathcal{L}$</td><td>$\mathbb{N}, \mathbb{Z}$</td><td colspan="4">?</td><td rowspan="2">undecidable<br>(Thm. 3.7)</td></tr>
<tr><td>finite or $\mathbb{N}, \mathbb{Z}$+**FSA**</td><td>$\leq$ EXPSPACE<br>$\geq$ NP</td><td>EXPSPACE<br>(Thm. 3.14)</td><td>$\leq$ 2EXPSPACE<br>$\geq$ EXPSPACE</td><td>?</td></tr>
<tr><td>arbitrary or $\mathbb{Q}, \mathbb{R}$ with **FSA**</td><td>$\leq$ 2EXPTIME<br>$\geq$ NP</td><td>$\leq$ 2EXPTIME<br>$\geq$ EXPSPACE</td><td>?</td><td>?</td><td>?</td></tr>
<tr><td rowspan="3">$\mathcal{PTL} \times \mathcal{L}$</td><td>$\mathbb{N}, \mathbb{Z}$</td><td colspan="4">?</td><td>undecidable<br>(Thm. 3.3)</td></tr>
<tr><td>finite or $\mathbb{N}, \mathbb{Z}$+**FSA**</td><td>$\leq$ EXPSPACE<br>$\geq$ PSPACE</td><td>EXPSPACE<br>(Thm. 3.12)</td><td>2EXPSPACE<br>(Thm. 3.13)</td><td>?</td><td>undecidable<br>(Thm. 3.5)</td></tr>
<tr><td>arbitrary or $\mathbb{Q}, \mathbb{R}$ with **FSA**</td><td>$\leq$ 2EXPTIME<br>$\geq$ PSPACE</td><td>$\leq$ 2EXPTIME<br>$\geq$ EXPSPACE<br>(Thm. 3.15)</td><td>?</td><td>?</td><td>?</td></tr>
</table>

Table 1: Complexity of the satisfiability problem for spatial and spatio-temporal logics.

Only the proofs show where the sources of the complexity are and whether they could be relevant for practical problems and the implementation of algorithms.

In this respect our proofs are actually rather informative. The decidability proof for $\mathcal{PTL}[\mathcal{S}4_u]$ immediately provides us with a modular algorithm combining any known procedures for the components. The EXPSPACE-completeness results for $\mathcal{PTL} \times \mathcal{BRCC}$-8 (with **FSA**) and $\mathcal{PTL} \circ \mathcal{BRCC}$-8 show an extremely close link between the spatio-temporal languages and the *one-variable fragment of first-order temporal logic*. The algorithmic problems investigated in the context of first-order temporal logic are, therefore, of the same character as those we deal with in the spatio-temporal context. Thus, the experience of working with algorithms for (fragments of) first-order temporal logics (Hodkinson, Wolter, & Zakharyaschev, 2000; Degtyarev, Fisher, & Konev, 2003; Kontchakov, Lutz, Wolter, & Zakharyaschev, 2004) about which we have a pretty good knowledge by now almost directly translates to insights into possible algorithms for spatio-temporal logics. The PSPACE-completeness result for $\mathcal{PTL} \circ \mathcal{RCC}$-8 is obtained by means of a reduction (modulo $\mathcal{RCC}$-8 reasoning) to $\mathcal{PTL}$. So we can conclude from the proof that it will be sufficient to have good solvers for $\mathcal{RCC}$-8 *and* $\mathcal{PTL}$ to obtain a reasonable prover for $\mathcal{PTL} \circ \mathcal{RCC}$-8. The interaction between the two components turned out to be rather weak.

In conclusion, the complexity proofs clearly show the algorithmic problems to be solved when dealing with the spatio-temporal logics presented in this paper. In particular, devising algorithms for these logics should be conceived as part of the more general enterprise of developing algorithms for propositional and the one-variable fragment of first-order temporal logic.

Here are some comments on and explanations of the most important results in Table 1:

1. The undecidability result for $\mathcal{PTL} \times \mathcal{S}4_u$, $\mathcal{PTL} \circ \mathcal{S}4_u$ and $\mathcal{PTL}_\Box \times \mathcal{S}4_u$ solves a major open problem of Wolter and Zakharyaschev (2002). It shows that, while $\mathcal{S}4_u$ is a suitable candidate for efficient pure spatial reasoning (Bennett, 1996; Renz & Nebel, 1998; Aiello & van Benthem, 2002a), its temporal extensions satisfying both (PC) and (OC) are not suitable for practical *spatio-temporal* representation and reasoning.

2. Logics like $\mathcal{PTL} \times \mathcal{BRCC}$-8 may turn out to be undecidable when interpreted in *arbitrary topological temporal models*. One of the main origins of their expressive power is a possibility to form infinite intersections and unions of regions. However, we can 'tame' the computational behaviour of these logics by imposing natural restrictions on the classes of admissible models such as **FSA**.

3. The PSPACE upper bound for $\mathcal{PTL} \circ \mathcal{RCC}$-8 and the EXPSPACE lower bound for $\mathcal{PTL} \circ \mathcal{BRCC}$-8 solve two other major open problems of Wolter and Zakharyaschev (2002). It is of interest to note that the spatial fragments of $\mathcal{PTL} \circ \mathcal{RCC}$-8 and $\mathcal{PTL} \circ \mathcal{BRCC}$-8 have the same computational complexity: both are NP-complete over arbitrary topological spaces. Thus the additional Boolean connectives on spatial regions interacting with the next-time operator $\bigcirc$ can make the logic substantially more complex.

4. The 2EXPSPACE-completeness result for $\mathcal{PTL} \times \mathcal{RC}$ with **FSA** and $\mathcal{PTL} \circ \mathcal{RC}$ is another example when a seemingly tiny increase of expressiveness results in a significant jump of complexity.

5. PSPACE-completeness of $\mathcal{PTL} \circ \mathcal{RCC}$-8 is a particularly 'good news,' since it shows that this combination of $\mathcal{PTL}$ and $\mathcal{RCC}$-8 has the same computational complexity as $\mathcal{PTL}$ itself, for which surprisingly fast systems have been implemented (Schwendimann, 1998; Hustadt & Konev, 2003). This gives us hope that 'practical' algorithms for $\mathcal{PTL} \circ \mathcal{RCC}$-8 can be implemented. Indeed, the proof shows that it may be possible to encode the satisfiability problem for $\mathcal{PTL} \circ \mathcal{RCC}$-8 into the satisfiability problem for $\mathcal{PTL}$ and then use $\mathcal{PTL}$ provers. We note that this complexity result has been conjectured by Demri and D'Souza (2002) and that our proof uses some ideas of Balbiani and Condotta (2002).

6. On the other hand, the EXPSPACE lower bounds for $\mathcal{PTL} \times \mathcal{BRCC}$-8 with **FSA** and $\mathcal{PTL} \circ \mathcal{BRCC}$-8 do not necessarily mean that reasoning with these logics is hopeless. In fact, we show that both of them can be regarded as fragments of the one-variable first-order temporal logic, for which tableau- and resolution-based decision procedures have been developed and implemented (Degtyarev et al., 2003; Kontchakov et al., 2004).

Of course, there are many directions of further research in spatio-temporal knowledge representation and reasoning. Here we mention only some of them that are closely related to the logics we have considered above.

- In this paper, we confined ourselves to considering linear flows of time. It may be of interest, however, to investigate the computational properties of spatio-temporal logics based on the branching time paradigm (see, e.g., Clarke & Emerson, 1981; Emerson & Halpern, 1985) in order to model uncertainty about the future. Recent results by Hodkinson, Wolter and Zakharyaschev (2001, 2002) give hope that such logics can be decidable.

- We confined ourselves to considering only mereotopological formalisms for the spatial dimension. It would be also of interest to consider spatial logics of directions (Ligozat, 1998), shape (Galton & Meathrel, 1999), size (Zimmermann, 1995), position (Clementini, Di Felice, & Hernández, 1997), or even their hybrids (Gerevini & Renz, 2002). We note that some results in this direction have been recently obtained by Balbiani and Condotta (2002) and Demri and D'Souza (2002).

- Another interesting and important perspective in both spatial and spatio-temporal representation and reasoning is to move from arbitrary topological spaces to those induced by *metric spaces* and introduce explicit and/or implicit numerical parameters. First encouraging steps in this direction have been made in the work (Kutz, Sturm, Suzuki, Wolter, & Zakharyaschev, 2003).

We conclude the paper with a number of open problems:

1. What is the precise computational complexity of $\mathcal{PTL} \times \mathcal{BRCC}$-8 with **FSA** over dense flows of time and arbitrary strict linear orders?

2. Are logics of the form $\mathcal{PTL} \times \mathcal{L}$ and $\mathcal{PTL}_{\Box} \times \mathcal{L}$, for $\mathcal{L} \in \{\mathcal{RC}, \mathcal{BRCC}\text{-}8, \mathcal{RCC}\text{-}8\}$, decidable without **FSA**?

3. Are combinations of $\mathcal{PTL}$ and $\mathcal{PTL}_\Box$ with $\mathcal{RC}^{max}$ (satisfying both (PC) and (OC)) decidable?

4. Is $\mathcal{PTL} \times \mathcal{S}4_u$ undecidable over dense flows of time and arbitrary strict linear orders?

5. Is $\mathcal{PTL} \times \mathcal{RCC}$-8 with **FSA** decidable in PSPACE?

## Acknowledgments

The work on this paper was partially supported by U.K. EPSRC grants no. GR/R45369/01, GR/R42474/01, GR/S61966/01 and GR/S63182/01. The work of the third author was also partially supported by Hungarian Foundation for Scientific Research grants T30314 and 035192.

Special thanks are due to the referees of the first version of this paper whose remarks, criticism and constructive suggestions have led to many days of intensive and exciting research, new results and, hopefully, a better paper.

## Appendix A. Complexity of Spatial Logics

In this appendix we prove Theorems 2.2 and 2.4. In these proofs we use the fact that $\mathcal{S}4_u$ (as well as its fragments) is complete with respect to (finite) Aleksandrov topological spaces (McKinsey & Tarski, 1944; Goranko & Passy, 1992). Recall from p. 174 that an *Aleksandrov* (*topological*) *model* is a pair of the form $\mathfrak{M} = \langle \mathfrak{G}, \mathfrak{V} \rangle$, where $\mathfrak{G} = \langle V, R \rangle$ is a quasi-order and $\mathfrak{V}$ is a map from the set of spatial variables into $2^V$. It will be more convenient for us to unify notation for spatial formulas and spatial terms and write $(\mathfrak{M}, x) \models \tau$ instead of $x \in \mathfrak{V}(\tau)$, for $\tau$ a spatial term and $x$ a point in $V$. In particular, by the definition of the interior and closure operators in Aleksandrov spaces,

$$(\mathfrak{M}, x) \models \mathbf{I}\tau \quad \text{iff} \quad \forall y \in V\big(xRy \to (\mathfrak{M}, y) \models \tau\big),$$
$$(\mathfrak{M}, x) \models \mathbf{C}\tau \quad \text{iff} \quad \exists y \in V\big(xRy \wedge (\mathfrak{M}, y) \models \tau\big).$$

By the *length* $\ell(\varphi)$ of a formula $\varphi$ we understand the number of subformulas and subterms occurring in $\varphi$.

**Proof of Theorem 2.2.** The proof follows from Lemmas A.1 and A.2 below which show together that every satisfiable $\mathcal{RC}$-formula can be satisfied in an Aleksandrov model of size polynomial (in fact, quadratical) in the length of the input formula (in other words, $\mathcal{RC}$ has the polynomial finite model property). Thus, we have a nondeterministic polynomial time algorithm for the satisfiability problem. ❑

In fact, Lemma A.1 shows that $\mathcal{RC}$ is complete with respect to a subclass of Aleksandrov spaces, namely, finite disjoint unions of finite brooms. Recall from p. 179 that a *broom* is a partial order $\mathfrak{b}$ of the form $\langle \{r\} \cup V_0, R \rangle$, where $R$ is the reflexive closure of $\{r\} \times V_0$ (see Fig. 4). We call $r$ the *root* of $\mathfrak{b}$ and points in $V_0$ the *leaves* of $\mathfrak{b}$; they are also referred to as *points of depth* 1 and 0, respectively. A broom $\mathfrak{b}$ is said to be a *$\kappa$-broom*, $\kappa \leq \omega$, if $|V_0| \leq \kappa$. In particular, we call a broom *finite* if it is an $n$-broom, for some $n < \omega$.

**Lemma A.1.** *Every satisfiable $\mathcal{RC}$-formula is satisfied in an Aleksandrov model based on a finite disjoint union of finite brooms.*

**Proof.** As is well-known, if an $\mathcal{RC}$-formula $\varphi$ is satisfiable then it can be satisfied in a finite Aleksandrov model $\mathfrak{M} = \langle \mathfrak{G}, \mathfrak{V} \rangle$, $\mathfrak{G} = \langle V, R \rangle$. Define a new relation $R'$ on $V$ by taking $R'$ to be the reflexive closure of $R \cap (V_1 \times V_0)$, where

$$V_0 = \{x \in V \mid \neg \exists y\, (xRy \wedge \neg\, yRx)\} \quad \text{and} \quad V_1 = V - V_0.$$

(Without loss of generality we may assume that $V_1 \neq \emptyset$ and no $y \in V_0$ has more than one proper $R$-predecessor.) Let $\mathfrak{G}' = \langle V, R' \rangle$ and $\mathfrak{M}' = \langle \mathfrak{G}', \mathfrak{V} \rangle$. Clearly, $\mathfrak{G}'$ is a partial order as required. We prove that, for every $\mathcal{RC}$-formula $\psi$,

$$\mathfrak{M} \models \psi \qquad \text{iff} \qquad \mathfrak{M}' \models \psi. \tag{9}$$

First we show that, for every Boolean region term $\varrho$ and every $x \in V$,

$$(\mathfrak{M}', x) \models \varrho \qquad \text{iff} \qquad (\mathfrak{M}, x) \models \varrho. \tag{10}$$

By definition, $(\mathfrak{M}', x) \models p$ iff $(\mathfrak{M}, x) \models p$, for every spatial variable $p$. It is readily seen that for every $y \in V_0$ and every spatial term $\tau$, we have $(\mathfrak{M}', y) \models \tau$ iff $(\mathfrak{M}, y) \models \tau$. Now, if $\varrho$ is a Boolean region term then $\varrho = \mathbf{CI}\tau$ for some spatial term $\tau$, and we clearly have:

$$\begin{aligned}
(\mathfrak{M}, x) \models \mathbf{CI}\tau \quad & \text{iff} \quad \exists y \in V \ \big(xRy \text{ and } \forall z \in V\ (yRz \to (\mathfrak{M}, z) \models \tau)\big) \\
& \text{iff} \quad \exists y \in V_0 \ \big(xR'y \text{ and } (\mathfrak{M}, y) \models \tau\big) \\
& \text{iff} \quad \exists y \in V_0 \ \big(xR'y \text{ and } (\mathfrak{M}', y) \models \tau\big) \\
& \text{iff} \quad \exists y \in V_0 \ \big(xR'y \text{ and } (\mathfrak{M}', y) \models \mathbf{I}\tau\big) \\
& \text{iff} \quad (\mathfrak{M}', x) \models \mathbf{CI}\tau.
\end{aligned}$$

Next, we extend (10) to spatial terms of the form $\mathbf{I}\varrho$ where $\varrho$ is a Boolean region term. If $(\mathfrak{M}, x) \models \mathbf{I}\varrho$ then $(\mathfrak{M}, y) \models \varrho$ whenever $xRy$, and so, by $R' \subseteq R$, we have $(\mathfrak{M}', x) \models \mathbf{I}\varrho$. Conversely, suppose $(\mathfrak{M}', x) \models \mathbf{I}\varrho$. Take any $y$ with $xRy$ and any $z \in V_0$ with $yRz$. We claim that $(\mathfrak{M}, z) \models \varrho$. Indeed, if $x \in V_1$ then this follows by IH from $xR'z$. If $x \in V_0$ then $zRx$. Since $(\mathfrak{M}', x) \models \varrho$, by IH and $\varrho = \mathbf{CI}\tau$, we obtain $(\mathfrak{M}, z) \models \varrho$. Now $(\mathfrak{M}, y) \models \varrho$ follows by $yRz$ and $\varrho = \mathbf{CI}\tau$. Thus, $(\mathfrak{M}, x) \models \mathbf{I}\varrho$.

Finally, we can easily extend (10) to arbitrary spatial terms and formulas of $\mathcal{RC}$ because both are constructed from spatial terms of the form $\varrho$ and $\mathbf{I}\varrho$, with $\varrho$ a Boolean region term, using operators that do not depend on the structure of the underlying partial order. Thus we have (9). ❑

**Lemma A.2.** *Every satisfiable $\mathcal{RC}$-formula $\varphi$ is satisfied in an Aleksandrov model based on a disjoint union of at most $\ell(\varphi)$ many $2\ell(\varphi)$-brooms.*

**Proof.** Remember that every $\mathcal{RC}$-formula $\varphi$ is (equivalent to) a Boolean combination of spatial formulas from some set $\Sigma_\varphi = \{\diamondsuit\tau_1, \dots, \diamondsuit\tau_m\}$.[5] For each $\diamondsuit\tau \in \Sigma_\varphi$, the spatial term

5. In the following proof we consider $\diamondsuit$ as primary and $\boxdot\tau$ as an abbreviation for $\neg\diamondsuit\overline{\tau}$.

$\tau$ is also a Boolean (or rather set-theoretic) combination of some $\varrho_1, \dots, \varrho_k$, $\mathbf{I}\varrho'_1, \dots, \mathbf{I}\varrho'_m$, where the $\varrho_i$ and the $\varrho'_i$ are Boolean region terms.

It follows from Lemma A.1 that $\varphi$ is satisfied in an Aleksandrov model $\mathfrak{M} = \langle \mathfrak{G}, \mathfrak{V} \rangle$, where $\mathfrak{G} = \langle V, R \rangle$ is a finite disjoint union of finite brooms. For every $\diamond\tau \in \Sigma_\varphi$ with $\mathfrak{M} \models \diamond\tau$, fix a point $x_\tau \in V$ such that $(\mathfrak{M}, x_\tau) \models \tau$. We may assume that the $x_\tau$ are pairwise distinct and that the roots of all brooms are the points of the form $x_\tau$ for $\diamond\tau \in \Sigma_\varphi$. Therefore, $\mathfrak{G}$ is a disjoint union of $\leq \ell(\varphi)$ many finite brooms $\mathfrak{b}_\tau$, for $\diamond\tau \in \Sigma_\varphi$.

Let us construct a new model $\mathfrak{M}'$ as follows. For each broom $\mathfrak{b}_\tau$, $\diamond\tau \in \Sigma_\varphi$, and each $\varrho \in \Xi_\tau$, we pick

- a leaf $y_{\tau,\varrho}$ of $\mathfrak{b}_\tau$ (if any) such that $(\mathfrak{M}, y_{\tau,\varrho}) \models \varrho$,
- a leaf $y_{\tau,\overline{\varrho}}$ of $\mathfrak{b}_\tau$ (if any) such that $(\mathfrak{M}, y_{\tau,\overline{\varrho}}) \models \overline{\varrho}$ and remove the other leaves of $\mathfrak{b}_\tau$. Denote by $\mathfrak{b}'_\tau$ the resulting broom. Clearly, it is a $2\ell(\varphi)$-broom. Let $\mathfrak{G}' = \langle V', R' \rangle$ be the disjoint union of all $\mathfrak{b}'_\tau$, for $\diamond\tau \in \Sigma_\varphi$, and $\mathfrak{M}' = \langle \mathfrak{G}', \mathfrak{V} \rangle$. It is easy to see that $\mathfrak{G}'$ is as required.

Now, to show that $\varphi$ is satisfied in $\mathfrak{M}'$, it suffices to prove that, for all $\diamond\tau \in \Sigma_\varphi$,

$$\mathfrak{M}' \models \diamond\tau \qquad \text{iff} \qquad \mathfrak{M} \models \diamond\tau. \tag{11}$$

By definition of $\mathfrak{M}'$, for all leaves $y$ of $\mathfrak{G}'$ and all spatial terms $\tau$,

$$(\mathfrak{M}', y) \models \tau \qquad \text{iff} \qquad (\mathfrak{M}, y) \models \tau.$$

Next, for every root $x_\tau$ of $\mathfrak{b}_\tau$, every $\diamond\tau \in \Sigma_\varphi$ and every $\varrho \in \Xi_\tau$, we have $(\mathfrak{M}, x_\tau) \models \varrho$ iff there is a leaf $y$ such that $x_\tau R y$ and $(\mathfrak{M}, y) \models \varrho$ (simply because $\varrho = \mathbf{CI}\delta$, for some $\delta$). It follows from the construction of $\mathfrak{M}'$ that $(\mathfrak{M}, x_\tau) \models \varrho$ iff $(\mathfrak{M}', x_\tau) \models \varrho$, for every $\varrho \in \Xi_\tau$. It also follows that $(\mathfrak{M}, x_\tau) \models \mathbf{I}\varrho$ implies $(\mathfrak{M}', x_\tau) \models \mathbf{I}\varrho$. Conversely, if $(\mathfrak{M}', x_\tau) \models \mathbf{I}\varrho$, but $(\mathfrak{M}, x_\tau) \not\models \mathbf{I}\varrho$ then there is a leaf $y$ such that $x_\tau R y$ and $(\mathfrak{M}, y) \not\models \varrho$ which is a contradiction. Since intersection and complement do not depend on the structure of the underlying frame, we have $(\mathfrak{M}', x_\tau) \models \tau$ iff $(\mathfrak{M}, x_\tau) \models \tau$, for every root $x_\tau$ of $\mathfrak{b}_\tau$, which proves (11). ❑

**Proof of Theorem 2.4.** The PSPACE upper bound follows from Theorem 2.1. The proof of PSPACE-hardness is by reduction of the validity problem for quantified Boolean formulas which is known to be PSPACE-complete (Stockmeyer, 1987). We will slightly modify the proof of Ladner (1977) (that shows the PSPACE-hardness of $\mathcal{S}4$), in order to take into account that the variables in $\mathcal{RC}^{max}$-formulas are always prefixed by **CI**.

We may assume that quantified Boolean formulas are of the form

$$\varphi = Q_1 p_1 \dots Q_n p_n \, \varphi',$$

where $Q_i \in \{\forall, \exists\}$ and $\varphi'$ is a Boolean formula with variables $p_1, \dots, p_n$. As is well known, all possible truth assignments to $p_1, \dots, p_n$ can be arranged as the leaves of a full binary tree of depth $n$. The left subtree of the root contains all truth assignments with $p_1$ true and the right subtree those with $p_1$ false; then we branch on $p_2$, then $p_3$, and so on. We can determine whether $\varphi$ is valid by pruning this full binary tree: whenever $Q_i$ is $\forall$, then we keep both subtrees at the $i$th level, and whenever $Q_i$ is $\exists$ then only one of them. If this way we can end up with a tree such that all its leaves evaluate $\varphi'$ to true, then $\varphi$ is valid, otherwise not.

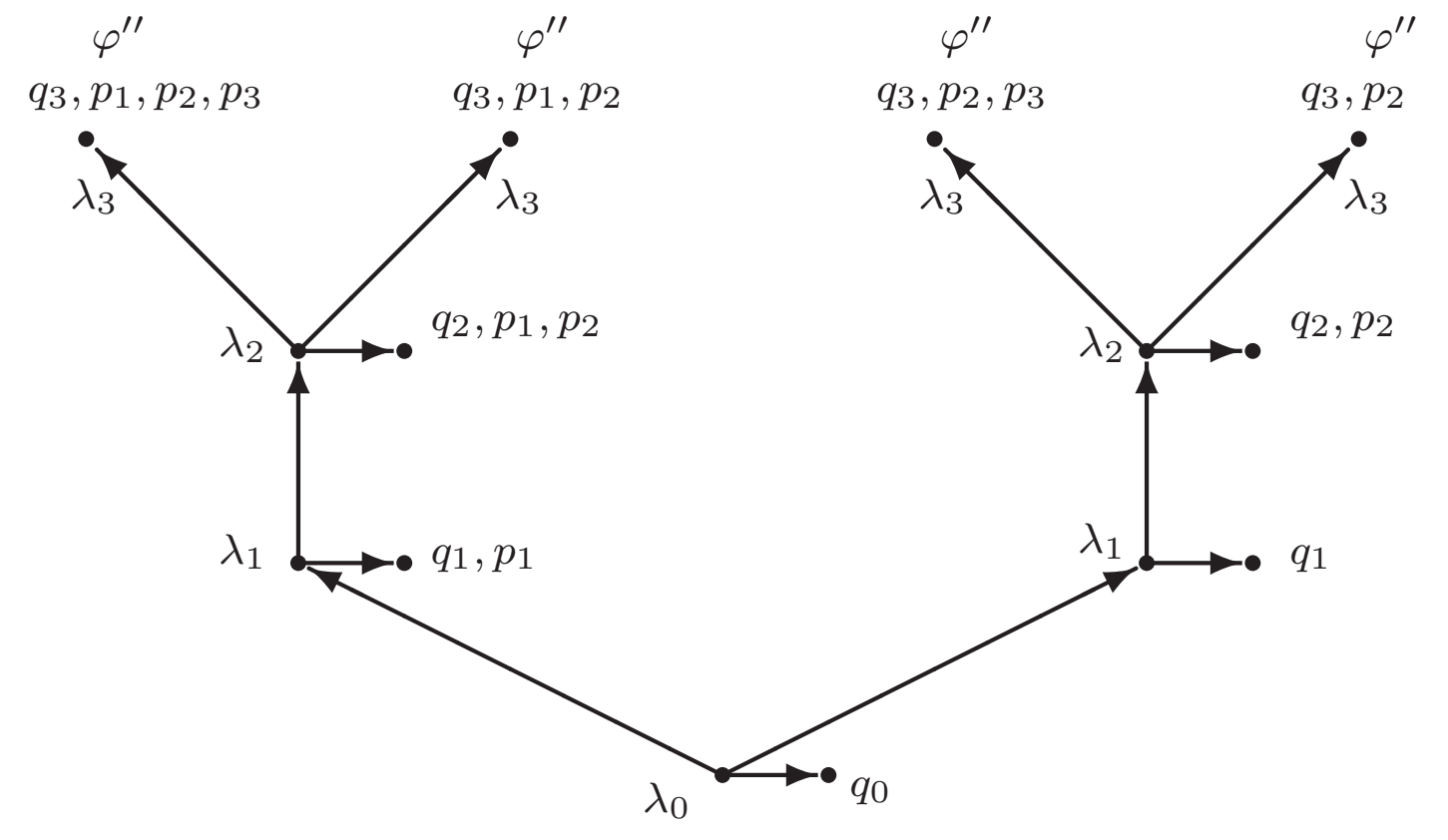

Figure 6: An Aleksandrov model that may satisfy $\varphi^*$, for $\varphi = \forall p_1 \exists p_2 \forall p_3\ \varphi'$.

We will 'generate' the leaves of this binary tree in Aleksandrov models with the help of an $\mathcal{RC}^{max}$-formula. More precisely, we will construct an $\mathcal{RC}^{max}$-formula $\varphi^*$ such that

- its length is polynomial in the length of $\varphi$, and
- $\varphi^*$ is satisfied in an Aleksandrov model iff $\varphi$ is valid.

Take fresh spatial variables $q_0, \dots, q_n$, and put, for $i = 0, \dots, n$,

$$\lambda_i = \begin{cases} \lceil q_0 \rceil \sqcap \lceil q_1 \rceil & \text{if} \quad i = 0; \\ \overline{\lceil q_{i-1} \rceil} \sqcap \lceil q_i \rceil \sqcap \lceil q_{i+1} \rceil, & \text{if} \quad 0 < i < n; \\ \overline{\lceil q_{n-1} \rceil} \sqcap \lceil q_n \rceil, & \text{if} \quad i = n. \end{cases}$$

Now consider the variables $p_1 \dots, p_n$ of $\varphi$ as spatial variables, and let $\varphi''$ be the result of replacing every occurrence of $p_i$ with $\lceil p_i \rceil$ in $\varphi'$. Put

$$\varphi^* = \Diamond \lambda_0 \ \wedge \bigwedge_{Q_i = \exists} \Box\big(\lambda_{i-1} \sqsubset (\tau_i^- \sqcup \tau_i^+)\big) \ \wedge \bigwedge_{Q_i = \forall} \Box\big(\lambda_{i-1} \sqsubset (\tau_i^- \sqcap \tau_i^+)\big) \ \wedge \ \Box\big(\lambda_n \sqsubset \varphi''\big),$$

where, for $i = 1, \dots, n$,

$$\tau_i^- = \mathbf{C}\big(\lambda_i \sqcap \overline{\lceil p_i \rceil}\big) \qquad \text{and} \qquad \tau_i^+ = \mathbf{C}\big(\lambda_i \sqcap \mathbf{I}\lceil p_i \rceil\big).$$

Clearly, $\varphi^*$ is an $\mathcal{RC}^{max}$-formula and its length is polynomial in the length of $\varphi$.

Suppose first that $\varphi$ is valid. Then Fig. 6 shows the structure of a possible Aleksandrov model satisfying $\varphi^*$.

The converse direction is similar to that of Ladner's proof (1977). Suppose that $\varphi^*$ is satisfied in an Aleksandrov model $\mathfrak{M}$. Then, for each 'necessary' sequence of truth values for $\lceil p_1 \rceil, \dots, \lceil p_n \rceil$, there is a point in $\mathfrak{M}$ 'reflecting' this sequence (we do not use the 'structure' of the spatial terms $\lambda_i$ here). Since, by the last conjunct of $\varphi^*$, $\varphi''$ holds in $\mathfrak{M}$ at all these points, we obtain that the quantified Boolean formula $\varphi$ must be valid. ❑

## Appendix B. Spatio-Temporal Logics Based on $\mathcal{S}4_u$

In this appendix we prove Theorems 3.1, 3.2, 3.6 and 3.7 as well as Proposition 3.4. Then Theorems 3.3 and 3.5 are immediate corollaries of Theorem 3.6. But first, some general results are established to be used later on.

We remind the reader that by an *Aleksandrov tt-model* we mean a tt-model based on an Aleksandrov (topological) space. Every such model can be regarded as a triple of the form $\mathfrak{K} = \langle \mathfrak{F}, \mathfrak{G}, \mathfrak{V} \rangle$, where $\mathfrak{F} = \langle W, < \rangle$ is a flow of time, $\mathfrak{G} = \langle V, R \rangle$ a quasi-order, and $\mathfrak{V}$ is a map associating with every spatial variable $p$ and every time point $w \in W$ a set $\mathfrak{V}(p, w) \subseteq V$. As in Appendix A, instead of $x \in \mathfrak{V}(\tau, w)$ we write $(\mathfrak{K}, \langle w, x \rangle) \models \tau$ to unify notation for spatio-temporal formulas and terms.

Given a spatio-temporal formula $\varphi$, we denote by $sub\,\varphi$ the set of all its subformulas and by $term\,\varphi$ the set of all spatio-temporal terms occurring in $\varphi$.

**Lemma B.1.** (i) *If a $\mathcal{PTL} \times \mathcal{S}4_u$-formula $\varphi$ is satisfied in a tt-model with* **FSA** *and based on a flow of time $\mathfrak{F}$, then $\varphi$ is satisfied in an Aleksandrov tt-model with* **FSA** *and based on $\mathfrak{F}$.*

(ii) *If a $\mathcal{PTL} \circ \mathcal{S}4_u$-formula $\varphi$ is satisfied in a tt-model based on a flow of time $\mathfrak{F}$, then $\varphi$ is satisfied in an Aleksandrov tt-model based on $\mathfrak{F}$ as well.*

*Moreover, in both cases we can choose an Aleksandrov tt-model $\mathfrak{K} = \langle \mathfrak{F}, \mathfrak{G}, \mathfrak{V} \rangle$ satisfying $\varphi$ (with $\mathfrak{F} = \langle W, < \rangle$ and $\mathfrak{G} = \langle V, R \rangle$) in such a way that for all $w \in W$, $x \in V$ and spatio-temporal terms $\tau$, the set*

$$A_{w,x,\tau} = \{ y \in V \mid xRy \text{ and } (\mathfrak{K}, \langle w, y \rangle) \models \tau \}$$

*contains an $R$-maximal point*[6] *(provided of course that $A_{w,x,\tau} \neq \emptyset$).*

**Proof.** The proof uses the Stone–Jónsson–Tarski representation of topological Boolean algebras (in particular, topological spaces) in the form of general frames (see, e.g., Goldblatt, 1976 or Chagrov & Zakharyaschev, 1997).

(i) Suppose that $\varphi$ is satisfied in a tt-model $\mathfrak{M} = \langle \mathfrak{F}, \mathfrak{T}, \mathfrak{U} \rangle$ with **FSA** and based on a topological space $\mathfrak{T} = \langle U, \mathbb{I} \rangle$. Denote by $V$ the set of all ultrafilters over $U$. For any two ultrafilters $\boldsymbol{x}_1, \boldsymbol{x}_2 \in V$, put $\boldsymbol{x}_1 R \boldsymbol{x}_2$ iff $\forall A \subseteq U\ (\mathbb{I}A \in \boldsymbol{x}_1 \to A \in \boldsymbol{x}_2)$. It is easy to see that $R$ is a quasi-order on $V$. Define an Aleksandrov tt-model $\mathfrak{K} = \langle \mathfrak{F}, \mathfrak{G}, \mathfrak{V} \rangle$ by taking $\mathfrak{G} = \langle V, R \rangle$ and $\mathfrak{V}(p, w) = \{ \boldsymbol{x} \in V \mid \mathfrak{U}(p, w) \in \boldsymbol{x} \}$. We show by induction on the construction of a spatio-temporal term $\tau$ that, for all $w \in W$ and $\boldsymbol{x} \in V$,

$$(\mathfrak{K}, \langle w, \boldsymbol{x} \rangle) \models \tau \qquad \text{iff} \qquad \mathfrak{U}(\tau, w) \in \boldsymbol{x}. \tag{12}$$

The basis of induction and the case of the Booleans are trivial. The case of $\tau = \mathbf{I}\tau'$ is standard (consult Goldblatt, 1976 or Chagrov & Zakharyaschev, 1997).

*Case* $\tau = \tau_1 \,\mathcal{U}\, \tau_2$. Assume that $(\mathfrak{K}, \langle w, \boldsymbol{x} \rangle) \models \tau_1 \,\mathcal{U}\, \tau_2$. Then there is $v > w$ such that $(\mathfrak{K}, \langle v, \boldsymbol{x} \rangle) \models \tau_2$ and $(\mathfrak{K}, \langle u, \boldsymbol{x} \rangle) \models \tau_1$ for all $u$ in the interval $(w, v)$. By IH, $\mathfrak{U}(\tau_2, v) \in \boldsymbol{x}$ and $\mathfrak{U}(\tau_1, u) \in \boldsymbol{x}$ for all $u \in (w, v)$. Since

$$\mathfrak{U}(\tau_1 \,\mathcal{U}\, \tau_2, w) \supseteq \mathfrak{U}(\tau_2, v) \cap \bigcap_{u \in (w,v)} \mathfrak{U}(\tau_1, u),$$

6. A point $z$ is said to be *R-maximal* in $A \subseteq V$ if, for every $z' \in A$, we have $z'Rz$ whenever $zRz'$.

we shall have $\mathfrak{U}(\tau_1 \,\mathcal{U}\, \tau_2, w) \in \boldsymbol{x}$ if we show that

$$\mathfrak{U}(\tau_2, v) \cap \bigcap_{u \in (w,v)} \mathfrak{U}(\tau_1, u) \in \boldsymbol{x}. \tag{13}$$

In view of **FSA**, we can find time points $u_1, \dots, u_l \in (w, v)$ such that

$$\mathfrak{U}(\tau_1, u_1) \cap \dots \cap \mathfrak{U}(\tau_1, u_l) = \bigcap_{u \in (w,v)} \mathfrak{U}(\tau_1, u),$$

which yields (13) because ultrafilters are closed under *finite* intersections.

Conversely, let $\mathfrak{U}(\tau_1 \,\mathcal{U}\, \tau_2, w) \in \boldsymbol{x}$. By **FSA**, there are time points $v_1, \dots v_l$ such that

$$\mathfrak{U}(\tau_1 \,\mathcal{U}\, \tau_2, w) = \bigcup_{1 \le i \le l} \Big( \mathfrak{U}(\tau_2, v_i) \cap \bigcap_{u \in (w, v_i)} \mathfrak{U}(\tau_1, u) \Big).$$

And since $\boldsymbol{x}$ is an ultrafilter,

$$\mathfrak{U}(\tau_2, v_i) \cap \bigcap_{u \in (w, v_i)} \mathfrak{U}(\tau_1, u) \in \boldsymbol{x},$$

for some $i$, $1 \le i \le l$. Therefore, by IH, $(\mathfrak{K}, \langle v_i, \boldsymbol{x} \rangle) \models \tau_2$ and $(\mathfrak{K}, \langle u, \boldsymbol{x} \rangle) \models \tau_1$ for all $u \in (w, v_i)$. Hence $(\mathfrak{K}, \langle w, \boldsymbol{x} \rangle) \models \tau_1 \,\mathcal{U}\, \tau_2$.

*Case* $\tau = \tau_1 \,\mathcal{S}\, \tau_2$ is considered analogously.

Now, we show that, for all $w \in W$ and spatio-temporal terms $\tau$,

$$(\mathfrak{K}, w) \models \boxdot\tau \qquad \text{iff} \qquad \mathfrak{U}(\tau, w) = U.$$

Suppose that $(\mathfrak{K}, w) \models \boxdot\tau$. Then $(\mathfrak{K}, \langle w, \boldsymbol{y} \rangle) \models \tau$ for all $\boldsymbol{y} \in V$, and so, by IH, $\mathfrak{U}(\tau, w) \in \boldsymbol{y}$ for all $\boldsymbol{y} \in V$. But then $\mathfrak{U}(\tau, w) = U$. Conversely, if $\mathfrak{U}(\tau, w) = U$ then $\mathfrak{U}(\tau, w) \in \boldsymbol{y}$ for all $\boldsymbol{y} \in V$, from which, by IH, $(\mathfrak{K}, w) \models \boxdot\tau$.

It follows immediately that $\varphi$ is satisfied in $\mathfrak{K}$. It should be also clear that $\mathfrak{K}$ satisfies **FSA**. This proves (i). The existence of $R$-maximal points in sets of the form $A_{w,\boldsymbol{x},\tau}$ (where $w \in W$, $\boldsymbol{x} \in V$ and $\tau$ is a spatio-temporal term) follows from a result of Fine (1974); see also (Chagrov & Zakharyaschev, 1997, Theorem 10.36).

(ii) The construction is the same as in (i). First we show by induction that, for every spatio-temporal term $\tau$ of $\mathcal{PTL} \circ \mathcal{S}4_u$, $(\mathfrak{K}, \langle w, \boldsymbol{x} \rangle) \models \tau$ iff $\mathfrak{U}(\tau, w) \in \boldsymbol{x}$. This time, however, instead of $\mathcal{U}$ and $\mathcal{S}$ we need the inductive step for $\bigcirc$.

*Case* $\tau = \bigcirc\tau'$. We have $(\mathfrak{K}, \langle w, \boldsymbol{x} \rangle) \models \bigcirc\tau'$ iff there exists an immediate successor $w'$ of $w$ such that $(\mathfrak{K}, \langle w', \boldsymbol{x} \rangle) \models \tau'$ iff, by IH, there is an immediate successor $w'$ of $w$ such that $\mathfrak{U}(\tau', w') \in \boldsymbol{x}$. It remains to recall that $\mathfrak{U}(\bigcirc\tau', w) = \mathfrak{U}(\tau', w')$ whenever $w'$ is the immediate successor of $w$ and $\mathfrak{U}(\bigcirc\tau', w) = \emptyset$ whenever $w$ has no immediate successor.

The remaining part of the proof is the same as in (i). ❑

**Proof of Proposition 3.4.** The implication ($\Leftarrow$) follows immediately from the definition.

($\Rightarrow$) Suppose that a $\mathcal{PTL} \times \mathcal{S}4_u$-formula $\varphi$ is satisfied in a tt-model with **FSA** and a flow of time $\mathfrak{F} = \langle W, < \rangle$. Then, by Lemma B.1 (i), $\varphi$ is satisfiable in an Aleksandrov tt-model $\mathfrak{M} = \langle \mathfrak{F}, \mathfrak{G}, \mathfrak{V} \rangle$ with **FSA** and based on a quasi-order $\mathfrak{G} = \langle V, R \rangle$. In view of **FSA**, for every $\tau \in term\varphi$, there are finitely many sets $A_1, \dots, A_k \subseteq V$ such that $\{\mathfrak{V}(\tau, w) \mid w \in W\} = \{A_1, \dots, A_k\}$. Therefore, there are finitely many time points $w_1, \dots, w_m \in W$ such that, for every $w \in W$, there is $w_i$, $1 \le i \le m$, with $\mathfrak{V}(\tau, w) = \mathfrak{V}(\tau, w_i)$ for *all* $\tau \in term\varphi$. Now we use the Lemmon filtration (see, e.g., Chagrov & Zakharyaschev, 1997) to construct a tt-model based on a *finite* Aleksandrov topological space. First, define an equivalence relation $\sim$ on $V$ by taking $x \sim y$ if

$$(\mathfrak{M}, \langle w_i, x\rangle) \models \tau \quad \text{iff} \quad (\mathfrak{M}, \langle w_i, y\rangle) \models \tau, \quad \text{for all } i,\ 1 \le i \le m, \text{ and } \tau \in term\varphi.$$

Denote by $[x]$ the equivalence class of $x \in V$. The set $V/_\sim$ of pairwise distinct equivalence classes is clearly finite. Define a binary relation $S$ on $V/_\sim$ by taking $[x]S[y]$ if

$$(\mathfrak{M}, \langle w_i, y\rangle) \models \mathbf{I}\tau \quad \text{whenever} \quad (\mathfrak{M}, \langle w_i, x\rangle) \models \mathbf{I}\tau, \quad \text{for all } i,\ 1 \le i \le m, \text{ and } \tau \in term\varphi.$$

Clearly, $S$ is well-defined, reflexive and transitive, and so $\mathfrak{G}' = \langle V/_\sim, S\rangle$ is a finite quasi-order. Let $\mathfrak{V}'(p, w) = \{[x] \mid x \in \mathfrak{V}(p, w)\}$, for every spatial variable $p$ and every $w \in W$.

Consider the tt-model $\mathfrak{M}' = \langle \mathfrak{F}, \mathfrak{G}', \mathfrak{V}'\rangle$. First we show that for all $\tau \in term\varphi$, $x \in V$ and $w \in W$,

$$(\mathfrak{M}, \langle w, x\rangle) \models \tau \qquad \text{iff} \qquad (\mathfrak{M}', \langle w, [x]\rangle) \models \tau.$$

The basis of induction follows from the definition of $\mathfrak{V}'$, the cases of intersection and complement are trivial, and those of temporal operators follow by IH.

Suppose that $(\mathfrak{M}, \langle w, x\rangle) \models \mathbf{I}\tau$ and $[x]S[y]$. Then there is a moment $w_i$ such that $(\mathfrak{M}, \langle w, z\rangle) \models \tau$ iff $(\mathfrak{M}, \langle w_i, z\rangle) \models \tau$, for all $\tau \in term\varphi$ and $z \in V$. By the definition of $S$, we have $(\mathfrak{M}, \langle w_i, y\rangle) \models \tau$, and so $(\mathfrak{M}, \langle w, y\rangle) \models \tau$. Finally, by IH, $(\mathfrak{M}', \langle w, [y]\rangle) \models \tau$, and since $y$ was arbitrary, we obtain $(\mathfrak{M}', \langle w, [x]\rangle) \models \mathbf{I}\tau$.

Conversely, let $(\mathfrak{M}', \langle w, [x]\rangle) \models \mathbf{I}\tau$ and $xRy$. Then $[x]S[y]$, and so $(\mathfrak{M}', \langle w, [y]\rangle) \models \tau$, from which, by IH, $(\mathfrak{M}, \langle w, y\rangle) \models \tau$. Thus, $(\mathfrak{M}, \langle w, x\rangle) \models \mathbf{I}\tau$.

Finally, by a straightforward induction on the structure of $\varphi$, one can show that

$$(\mathfrak{M}, w) \models \psi \qquad \text{iff} \qquad (\mathfrak{M}', w) \models \psi,$$

for all $\psi \in sub\varphi$ and $w \in W$. It follows that $\varphi$ is satisfied in $\mathfrak{M}'$. ❑

### B.1 Temporalisations of $\mathcal{S}4_u$

**Lemma B.2.** *Let $\Gamma$ be a finite set of $\mathcal{S}4_u$-formulas. Then there is a finite quasi-order $\mathfrak{G}$ such that every satisfiable subset $\Phi \subseteq \Gamma$ is satisfied in some Aleksandrov model based on $\mathfrak{G}$.*

**Proof.** For every satisfiable $\Phi \subseteq \Gamma$, fix a model based on a finite quasi-order $\mathfrak{G}_\Phi = \langle V_\Phi, R_\Phi\rangle$ and satisfying $\Phi$. Let $n = \max\{|V_\Phi| : \Phi \subseteq \Gamma,\ \Phi \text{ is satisfiable}\}$ and let $\mathfrak{G}$ be the disjoint union of $n$ full $n$-ary (transitive) trees of depth $n$ whose nodes are clusters of cardinality $n$. It is not difficult to see that every $\mathfrak{G}_\Phi$ is a p-morphic image of $\mathfrak{G}$. Therefore, every satisfiable $\Phi \subseteq \Gamma$ is satisfied in an Aleksandrov model based on $\mathfrak{G}$. ❑

**Proof of Theorem 3.1.** PSPACE-hardness follows from Theorem 2.1 or 2.5. We show the matching upper bound.

Let $\varphi$ be a $\mathcal{PTL}[\mathcal{S}4_u]$-formula. Since $\varphi$ is a $\mathcal{PTL} \circ \mathcal{S}4_u$-formula, by Lemma B.1 (ii), it is satisfiable in a tt-model iff it is satisfiable in an Aleksandrov tt-model based on the same flow of time. With every (spatial) subformula $\boxdot\tau$ of $\varphi$ we associate a fresh propositional variable $\mathsf{p}_\tau$ and denote by $\varphi^*$ the $\mathcal{PTL}$-formula that results from $\varphi$ by replacing all its subformulas of the form $\boxdot\tau$ with $\mathsf{p}_\tau$. We claim that $\varphi$ is satisfiable in an Aleksandrov tt-model over a flow of time $\mathfrak{F} = \langle W, < \rangle$ iff

- there exists a temporal model $\mathfrak{N} = \langle \mathfrak{F}, \mathfrak{U} \rangle$ satisfying $\varphi^*$ and,
- for every $w \in W$, the set $\Phi_w = \{\boxdot\tau \mid (\mathfrak{N}, w) \models \mathsf{p}_\tau\} \cup \{\neg\boxdot\tau \mid (\mathfrak{N}, w) \models \neg\mathsf{p}_\tau\}$ of spatial formulas is satisfiable.

The implication ($\Rightarrow$) is obvious. Conversely, suppose that we have a temporal model $\mathfrak{N}$ satisfying the conditions above. Let $\Gamma = \bigcup_{w \in W} \Phi_w$. By Lemma B.2, there is a finite quasi-order $\mathfrak{G}$ such that, for every $w \in W$, we have $\langle \mathfrak{G}, \mathfrak{V}_w \rangle \models \Phi_w$ for some valuation $\mathfrak{V}_w$. It should be clear that $\varphi$ is satisfied in the Aleksandrov tt-model $\langle \mathfrak{F}, \mathfrak{G}, \mathfrak{V} \rangle$, where $\mathfrak{V}(p, w) = \mathfrak{V}_w(p)$, for every spatial variable $p$ and every $w \in W$.

Now, to devise a decision procedure for $\mathcal{PTL}[\mathcal{S}4_u]$ which uses polynomial space in the length of the input formula, one can take the corresponding nondeterministic PSPACE algorithm for $\mathcal{PTL}$ (Sistla & Clarke, 1985; Reynolds, 2004, 2003) and modify it as follows. The algorithm constructs a 'pure' temporal model $\mathfrak{N} = \langle \mathfrak{F}, \mathfrak{U} \rangle$ for $\varphi^*$ and every time it produces a state for a time instant $w \in W$, it additionally checks whether the set $\Phi_w$ of spatial formulas is satisfiable. By Theorem 2.1, this extra test can also be performed by a PSPACE algorithm, which does not increase the complexity of the 'combined' algorithm. ❑

**Proof of Theorem 3.2.** The proof is essentially the same as that of Theorem 3.1, but now nondeterministic polynomial-time algorithms for the component logics are available. ❑

### B.2 Undecidability of $\mathcal{PTL} \circ \mathcal{S}4_u$ and $\mathcal{PTL}_\Box \times \mathcal{S}4_u$

Note that although our spatio-temporal languages contain no propositional variables, we still can simulate them: for a spatial variable $p$, formula $\boxdot p$ can be regarded as a proposition. Thus, in what follows by a *propositional variable* $\mathsf{p}$ we mean the formula $\boxdot p$, for a spatial variable $p$ (note the different typefaces used to denote propositional and spatial variables).

**Proof of Theorem 3.6.** The proof is by reduction of the undecidable Post's (1946) correspondence problem or PCP, for short. It is formulated as follows. Given a finite alphabet $A$ and a finite set $P$ of pairs $\langle v_1, u_1 \rangle, \dots, \langle v_k, u_k \rangle$ of nonempty finite words

$$v_i = \langle b_1^i, \dots, b_{l_i}^i \rangle, \qquad u_i = \langle c_1^i, \dots, c_{r_i}^i \rangle \qquad (i = 1, \dots, k)$$

over $A$, an instance of PCP, decide whether there exist an $N \geq 1$ and a sequence $i_1, \dots, i_N$ of indices such that

$$v_{i_1} * \dots * v_{i_N} = u_{i_1} * \dots * u_{i_N}, \tag{14}$$

where $*$ is the concatenation operation. We will construct (using only future-time temporal operators) a $\mathcal{PTL} \circ \mathcal{S}4_u$-formula $\varphi_{A,P}$ such that

(i) the length of $\varphi_{A,P}$ is a polynomial function in the size of both $A$ and $P$;

(ii) if $\varphi_{A,P}$ is satisfiable in a tt-model based on $\langle \mathbb{N}, < \rangle$ then there exist an $N \geq 1$ and a sequence $i_1, \dots, i_N$ of indices such that (14) holds;

(iii) if there exist an $N \geq 1$ and a sequence $i_1, \dots, i_N$ of indices such that (14) holds then $\varphi_{A,P}$ is satisfiable in a tt-model with **FSA** and based on $\langle \mathbb{N}, < \rangle$;

(iv) $\varphi_{A,P}$ is satisfiable in a tt-model based on $\langle \mathbb{N}, < \rangle$ iff $\varphi_{A,P}$ is satisfiable in a tt-model based on a finite flow of time.

The case of $\langle \mathbb{Z}, < \rangle$ follows immediately. By Lemma B.1 (ii), it suffices to consider only Aleksandrov tt-models for $\varphi_{A,P}$.

We build $\varphi_{A,P}$ using spatial variables $left_a$ and $right_a$ ($a \in A$), $left$, $right$ and $stripe$, as well as propositional variables $\mathsf{pair}_i$, for every pair $\langle v_i, u_i \rangle$, $1 \leq i \leq k$, and $\mathsf{range}$.

The variable $\mathsf{range}$ is required to 'relativise' temporal operators $\Box_F$ and $\Diamond_F$ in order to ensure that we can construct a model based on a finite flow of time. The variable $stripe$ is used to introduce a new 'strict closure' operator in Aleksandrov spaces by taking, for every spatio-temporal term $\tau$,

$$\mathbf{S}\tau = \big(stripe \sqsubset \mathbf{C}(\overline{stripe} \sqcap \mathbf{C}\tau)\big) \sqcap \big(\overline{stripe} \sqsubset \mathbf{C}(stripe \sqcap \mathbf{C}\tau)\big).$$

Denote by $\mathbf{S}^n$ a sequence of $n$ operators $\mathbf{S}$. Other abbreviations we need are $\tau_1 \equiv \tau_2$ which stands for $(\tau_1 \sqsubset \tau_2) \sqcap (\tau_2 \sqsubset \tau_1)$ and $\Box_F^+\varphi$ which replaces $\varphi \wedge \Box_F\varphi$.

The formula $\varphi_{A,P}$ is defined as the conjunction

$$\varphi_{A,P} = \varphi_{range} \ \wedge \ \varphi_{stripe} \ \wedge \ \varphi_{pair} \ \wedge \ \varphi_{eq} \ \wedge \ \varphi_{left} \ \wedge \ \varphi_{right},$$

where

$$\begin{aligned}
\varphi_{range} &= \mathsf{range} \wedge \Diamond_F \neg\mathsf{range} \wedge \Box_F(\neg\mathsf{range} \to \Box_F\neg\mathsf{range}),\\
\varphi_{pair} &= \Box_F^+\Big(\Diamond_F\mathsf{range} \to \bigvee_{1 \leq i \leq k} \mathsf{pair}_i \ \wedge \bigwedge_{1 \leq i < j \leq k} \neg(\mathsf{pair}_i \wedge \mathsf{pair}_j)\Big),\\
\varphi_{stripe} &= \Box_F^+\big(\Diamond_F\mathsf{range} \to \boxdot(stripe \equiv \bigcirc stripe)\big),\\
\varphi_{eq} &= \Diamond_F\Big(\mathsf{range} \wedge \bigwedge_{a \in A} \boxdot(left_a \equiv right_a)\Big),
\end{aligned}$$

$\varphi_{left}$ is the conjunction of (15)–(21), for all $i$ with $1 \leq i \leq k$,

$$\bigwedge_{\substack{a \neq b\\ a,b \in A}} \Box_F^+ \neg\diamondsuit(left_a \sqcap left_b) \ \wedge \ \Box_F^+\boxdot\big(left \equiv \bigsqcup_{a \in A} left_a\big), \tag{15}$$

$$\bigwedge_{a \in A} \Box_F^+\big(\mathsf{pair}_i \to \boxdot(left_a \sqsubset \bigcirc left_a)\big), \tag{16}$$

$$\boxdot\overline{left} \ \wedge \ \Box_F^+\boxdot(\overline{left} \sqsubset \overline{\mathbf{S}left}), \tag{17}$$

$$\Box_F^+\big(\mathsf{pair}_i \to \boxdot(\overline{left} \sqsubset \bigcirc\overline{\mathbf{S}^{l_i}left})\big), \tag{18}$$

$$\Box_F^+\big(\mathsf{pair}_i \to \bigwedge_{j < l_i} \bigcirc\boxdot((\mathbf{S}^j left \sqcap \overline{\mathbf{S}^{j+1}left}) \sqsubset left_{b^i_{l_i - j}})\big), \tag{19}$$

$$\mathsf{pair}_i \to \bigcirc\diamondsuit\tau_i^{left}, \tag{20}$$

$$\Box_F\big(\mathsf{pair}_i \to \boxdot((left \sqcap \overline{\mathbf{S}left}) \sqsubset \bigcirc\mathbf{S}\tau_i^{left})\big), \tag{21}$$

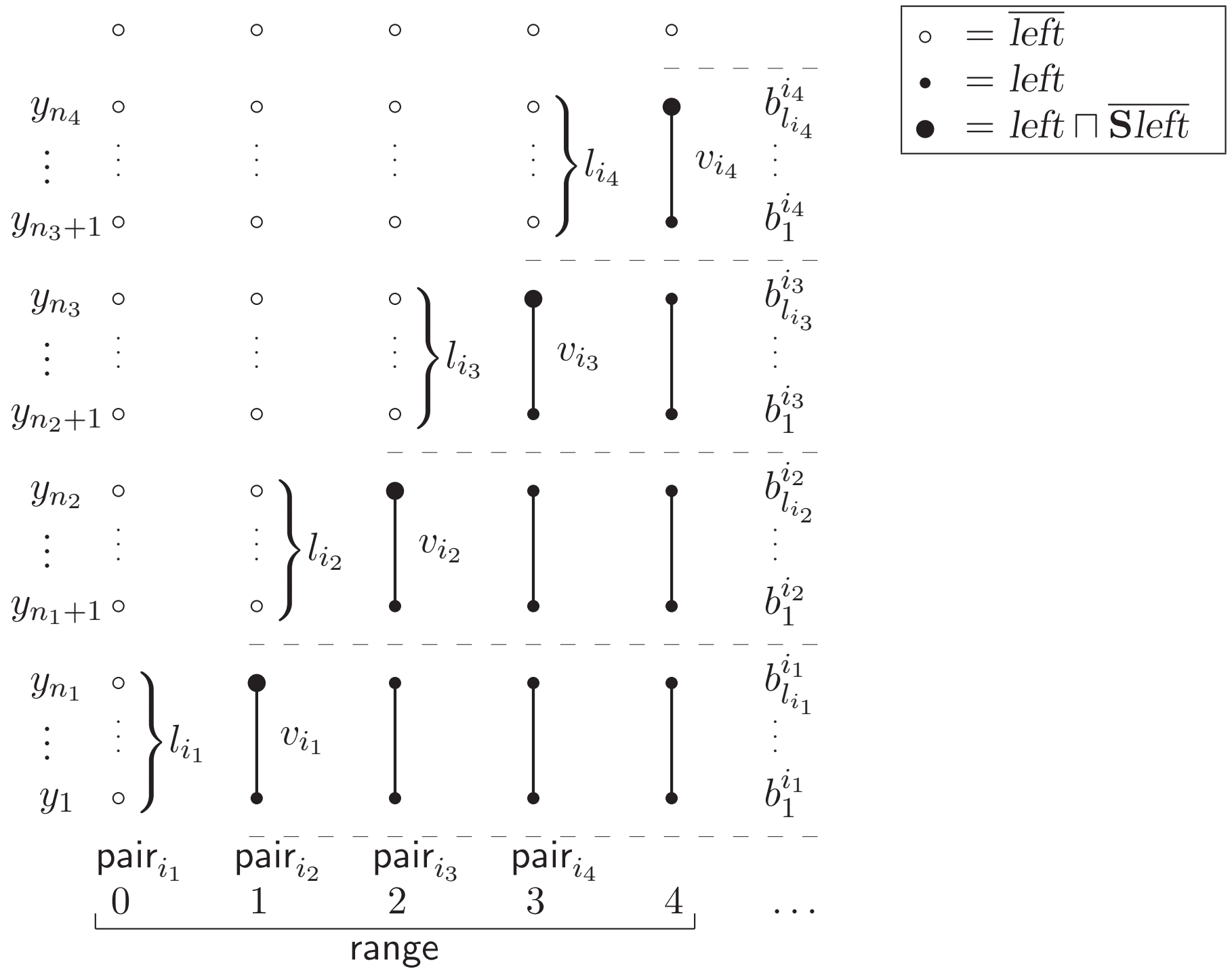

Figure 7: Model satisfying $\varphi_{left}$, for $N = 4$.

where

$$\tau_i^{left} = left_{b_1^i} \sqcap \mathbf{S}\big(left_{b_2^i} \sqcap \mathbf{S}(left_{b_3^i} \sqcap \cdots \sqcap \mathbf{S}left_{b_{l_i}^i})\dots\big)$$

(remember that $l_i$ is the length of the word $v_i$). The conjunct $\varphi_{right}$ is defined by replacing in $\varphi_{left}$ all occurrences of *left* with *right*, $left_a$ with $right_a$ (for $a \in A$), $l_i$ with $r_i$ and $\tau_i^{left}$ with $\tau_i^{right}$, which is defined similarly. (Note that $\mathsf{pair}_i$ occurs in both $\varphi_{left}$ and $\varphi_{right}$.)

Let us prove that $\varphi_{A,P}$ is as required. Suppose that $(\mathfrak{M}, 0) \models \varphi_{A,P}$, for an Aleksandrov tt-model $\mathfrak{M} = \langle \langle \mathbb{N}, < \rangle, \mathfrak{G}, \mathfrak{V} \rangle$ with $\mathfrak{G} = \langle V, R \rangle$. Since $(\mathfrak{M}, 0) \models \varphi_{eq}$, we can find an $N$, $1 \leq N < \omega$, such that

$$(\mathfrak{M}, N) \models \mathsf{range} \wedge \bigwedge_{a \in A} \boxdot(left_a \equiv right_a). \tag{22}$$

In view of $\varphi_{range}$, we have $(\mathfrak{M}, j) \models \mathsf{range}$ for all $j$, $0 \leq j \leq N$. Let $i_1, \dots, i_N$ be the sequence of indices such that, for $1 \leq j \leq N$, we have $(\mathfrak{M}, j-1) \models \mathsf{pair}_{i_j}$ ($\varphi_{pair}$ ensures that there is a unique sequence of this sort). We claim that (14) holds for this sequence.

Since $\varphi_{stripe}$ holds in $\mathfrak{M}$ at 0, we have, for every $y \in V$, $(\mathfrak{M}, \langle 0, y \rangle) \models stripe$ iff $(\mathfrak{M}, \langle j, y \rangle) \models stripe$ for all $j$, $0 \leq j \leq N$. Denote by $R_s$ the transitive binary relation on $V$ defined by taking $xR_sy$ if there is $z \in V$ such that $xRzRy$ and $(\mathfrak{M}, \langle 0, x \rangle) \models stripe$ holds iff $(\mathfrak{M}, \langle 0, z \rangle) \not\models stripe$. Then we clearly have that, for every $j$, $0 \leq j \leq N$, and every $x \in V$,

$$(\mathfrak{M}, \langle j, x \rangle) \models \mathbf{S}\tau \quad \text{iff} \quad \text{there is } y \in V \text{ such that } xR_sy \text{ and } (\mathfrak{M}, \langle j, y \rangle) \models \tau.$$

Call a sequence $\langle y_1, \dots, y_l \rangle$ of (not necessarily distinct) points from $V$ an *$R_s$-path in $\mathfrak{V}(left, j)$ of length $l$* if $y_1, \dots, y_l \in \mathfrak{V}(left, j)$ and $y_1R_sy_2R_s \dots R_sy_l$. For every sequence

$z_1, \dots, z_l$ of points from $\mathfrak{V}(\mathit{left}, j)$ we define

$$\mathit{leftword}_j(z_1, \dots, z_l) = \langle a_1, \dots, a_l \rangle\,,$$

where the $a_i$ are the (uniquely determined by (15)) symbols from $A$ with $(\mathfrak{M}, \langle j, z_i\rangle) \models \mathit{left}_{a_i}$.

We will show now that, for every $j$, $1 \le j \le N$, the following holds:

(a) there exists an $R_s$-path $\langle y_1, \dots, y_{n_j} \rangle$ in $\mathfrak{V}(\mathit{left}, j)$ of length $n_j = l_{i_1} + \cdots + l_{i_j}$ such that

$$\mathit{leftword}_j(y_1, \dots, y_{n_j}) = v_{i_1} * \dots * v_{i_j};$$

(b) every $R_s$-path in $\mathfrak{V}(\mathit{left}, j)$ is of length $\le n_j$;

(c) for every $R_s$-path $\langle y_1, \dots, y_{n_j} \rangle$ in $\mathfrak{V}(\mathit{left}, j)$, we have

$$\mathit{leftword}_j(y_1, \dots, y_{n_j}) = v_{i_1} * \dots * v_{i_j}.$$

Indeed, for $j = 1$, we have (a) by $(\mathfrak{M}, 0) \models \mathsf{pair}_{i_1}$ and (20), (b) by (17) and (18), and (c) by (19). Now assume inductively that (a)–(c) hold for some $j$, $1 \le j < N$. Let $\langle y_1, \dots, y_{n_j} \rangle$ be a maximal $R_s$-path in $\mathfrak{V}(\mathit{left}, j)$. First, by (16), $y_1, \dots, y_{n_j} \in \mathfrak{V}(\mathit{left}, j+1)$. Second, since $(\mathfrak{M}, \langle j, y_{n_j}\rangle) \models \mathit{left} \sqcap \overline{\mathbf{S}\mathit{left}}$ and $(\mathfrak{M}, j) \models \mathsf{pair}_{i_{j+1}}$, (21) now implies that there exist $y_{n_j+1}, \dots, y_{n_j+l_{i_{j+1}}}$ such that $\langle y_1, \dots, y_{n_j+l_{i_{j+1}}} \rangle$ is an $R_s$-path in $\mathfrak{V}(\mathit{left}, j+1)$, as required in (a). For (b) and (c), observe first that for every $R_s$-path $\langle y_1, \dots, y_l \rangle$ in $\mathfrak{V}(\mathit{left}, j+1)$, $\langle y_1, \dots, y_{l-l_{i_{j+1}}} \rangle$ is an $R_s$-path in $\mathfrak{V}(\mathit{left}, j)$, by (18). So $l \le n_{j+1}$ must hold. If $l = n_{j+1}$ then $\mathit{leftword}_j(y_1, \dots, y_{l-l_{i_{j+1}}}) = v_{i_1} * \dots * v_{i_j}$ by the induction hypothesis, and so $\mathit{leftword}_{j+1}(y_1, \dots, y_{l-l_{i_{j+1}}}) = v_{i_1} * \dots * v_{i_j}$ by (16). On the other hand, $\mathit{leftword}_{j+1}(y_{l-l_{i_{j+1}}+1}, \dots, y_l) = v_{i_{j+1}}$ by (19), and so we have $\mathit{leftword}_{j+1}(y_1, \dots, y_l) = v_{i_1} * \dots * v_{i_j} * v_{i_{j+1}}$, as required.

We can repeat the argument above for the 'right side' as well. For every sequence $z_1, \dots, z_l$ of points from $\mathfrak{V}(\mathit{right}, j)$, define

$$\mathit{rightword}_j(z_1, \dots, z_l) = \langle a_1, \dots, a_l \rangle\,,$$

where the $a_i$ are the uniquely determined elements of $A$ such that $(\mathfrak{M}, \langle j, z_i\rangle) \models \mathit{right}_{a_i}$. We then have, for every $1 \le j \le N$:

(a′) there is an $R_s$-path $\langle y_1, \dots, y_{m_j} \rangle$ in $\mathfrak{V}(\mathit{right}, j)$ of length $m_j = r_{i_1} + \cdots + r_{i_j}$ such that

$$\mathit{rightword}_j(y_1, \dots, y_{m_j}) = u_{i_1} * \dots * u_{i_j};$$

(b′) every $R_s$-path in $\mathfrak{V}(\mathit{right}, j)$ is of length $\le m_j$;

(c′) for every $R_s$-path $\langle y_1, \dots, y_{m_j} \rangle$ in $\mathfrak{V}(\mathit{right}, j)$, we have

$$\mathit{rightword}_j(y_1, \dots, y_{m_j}) = u_{i_1} * \dots * u_{i_j}.$$

Now, by (15) and (22), we have $\mathfrak{V}(\mathit{left}, N) = \mathfrak{V}(\mathit{right}, N)$. By (a), there exists an $R_s$-path $\langle y_1, \dots, y_l\rangle$ in $\mathfrak{V}(\mathit{left}, N)$ such that $l = n_N$ and $\mathit{leftword}_N(y_1, \dots, y_l) = v_{i_1} * \dots * v_{i_N}$. By (b′), we have $n_N \le m_N$. Similarly, using (a′) and (b), we obtain $m_N \le n_N$, from which $n_N = m_N$. Hence, by (c′), $\mathit{rightword}_N(y_1, \dots, y_l) = u_{i_1} * \dots * u_{i_N}$. Since, by (22),

$$\mathit{leftword}_N(y_1, \dots, y_l) = \mathit{rightword}_N(y_1, \dots, y_l),$$

we finally obtain $v_{i_1} * \dots * v_{i_N} = u_{i_1} * \dots * u_{i_N}$, as required.

Conversely, suppose there is an $N \ge 1$ and a sequence $i_1, \dots, i_N$ for which (14) holds. We will show that $\varphi_{A,P}$ is satisfiable in an Aleksandrov tt-model $\mathfrak{M} = \langle\langle\mathbb{N}, <\rangle, \langle\mathbb{N}, \le\rangle, \mathfrak{V}\rangle$ with **FSA**. Let $n_j = l_{i_1} + \dots + l_{i_j}$ and $m_j = r_{i_1} + \dots + r_{i_j}$ for every $j$, $1 \le j \le N$. By our assumption, $n_N = m_N$ and we have

$$v_{i_1} * \dots * v_{i_N} = \langle a_1, \dots, a_{n_N}\rangle = u_{i_1} * \dots * u_{i_N}.$$

Define a valuation $\mathfrak{V}$ by taking

- $\mathfrak{V}(\mathsf{range}, j)$ is true iff $0 \le j \le N$,
- $\mathfrak{V}(\mathit{stripe}, j) = \{2m \mid m < \omega,\ 0 \le j \le N\}$,
- $\mathfrak{V}(\mathsf{pair}_i, j-1)$ is true iff $i = i_j$ and $1 \le j \le N$,
- $\mathfrak{V}(\mathit{left}_a, j) = \{k \mid 1 \le k \le n_j,\ a_k = a\}$ for $a \in A$ and $1 \le j \le N$,
- $\mathfrak{V}(\mathit{right}_a, j) = \{k \mid 1 \le k \le m_j,\ a_k = a\}$ for $a \in A$ and $1 \le j \le N$,
- $\mathfrak{V}(\mathit{left}, j) = \bigcup_{a \in A} \mathfrak{V}(\mathit{left}_a, j)$ and $\mathfrak{V}(\mathit{right}, j) = \bigcup_{a \in A} \mathfrak{V}(\mathit{right}_a, j)$.

One can easily check that under this valuation we have $(\mathfrak{M}, 0) \models \varphi_{A,P}$ and $\mathfrak{M}$ satisfies **FSA**. It is also readily seen that $\varphi_{A,P}$ is satisfiable in a tt-model based on $\langle\mathbb{N}, <\rangle$ iff it is satisfiable in a tt-model based on a finite flow of time. ❑

**Proof of Theorem 3.7.** We show this by modifying formulas from the proof of Theorem 3.6. First, we replace $\varphi_{\mathit{stripe}}$ with

$$\Box_F^+ \boxdot(\mathit{stripe} \sqsubset \Box_F \mathit{stripe}) \ \wedge\ \Box_F^+ \boxdot(\overline{\mathit{stripe}} \sqsubset \Box_F \overline{\mathit{stripe}}).$$

Then, $\varphi_{\mathit{left}}$ is the conjunction of (15′)–(21′), for all $i$ with $1 \le i \le k$,

$$\bigwedge_{\substack{a \ne b\\ a,b \in A}} \Box_F^+ \neg \diamondplus\big(\mathit{left}_a \sqcap \mathit{left}_b\big) \ \wedge\ \Box_F^+ \boxdot\big(\mathit{left} \equiv \bigsqcup_{a \in A} \mathit{left}_a\big), \qquad (15')$$

$$\bigwedge_{a \in A} \Box_F^+\big(\mathsf{pair}_i \to \boxdot(\mathit{left}_a \sqsubset \Box_F \mathit{left}_a)\big), \qquad (16')$$

$$\boxdot\overline{\mathit{left}} \ \wedge\ \Box_F^+ \boxdot(\overline{\mathit{left}} \sqsubset \overline{\mathbf{S}\mathit{left}}), \qquad (17')$$

$$\Box_F^+\big(\mathsf{pair}_i \to \boxdot(\overline{\mathit{left}} \sqsubset \Diamond_F \overline{\mathbf{S}^{l_i}\mathit{left}})\big), \qquad (18')$$

$$\Box_F^+\big(\mathsf{pair}_i \to \bigwedge_{j < l_i} \boxdot\big((\overline{\mathit{left}} \sqcap \Box_F \mathit{left}) \sqsubset \Box_F((\mathbf{S}^j \mathit{left} \sqcap \overline{\mathbf{S}^{j+1}\mathit{left}}) \sqsubset \mathit{left}_{b^i_{l_i - j}})\big)\big), \qquad (19')$$

$$\mathsf{pair}_i \to \Box_F \diamondplus \tau_i^{\mathit{left}}, \qquad (20')$$

$$\Box_F\big(\mathsf{pair}_i \to \boxdot((\mathit{left} \sqcap \overline{\mathbf{S}\mathit{left}}) \sqsubset \Box_F \mathbf{S}\tau_i^{\mathit{left}})\big), \qquad (21')$$

where the $\tau_i^{left}$ are defined exactly as in the proof of Theorem 3.6. Formula $\varphi_{right}$ is modified in a similar way. ❑

*Remark* B.3. In fact, the set of PCP instances without solutions is not recursively enumerable and therefore, the proofs above show that the sets of $\mathcal{PTL} \circ \mathcal{S}4_u$ and $\mathcal{PTL}_\Box \times \mathcal{S}4_u$-formulas which are true in all models based on $\langle \mathbb{N}, < \rangle$, $\langle \mathbb{Z}, < \rangle$ or finite flows of time are not recursively enumerable either. Therefore, these logics are not recursively axiomatisable.

## Appendix C. Spatio-Temporal Logics Based on $\mathcal{RC}$

In this appendix we establish lower and upper complexity bounds for a wide range of decidable spatio-temporal combinations and, in particular, prove Theorems 3.8–3.15. We begin with a straightforward generalisation of Lemma A.1 to the spatio-temporal case:

**Lemma C.1.** (i) *If a $\mathcal{PTL} \times \mathcal{RC}$-formula $\varphi$ is satisfiable in a tt-model with* **FSA** *and based on a flow of time $\mathfrak{F}$ then $\varphi$ is satisfiable in an Aleksandrov tt-model based on $\mathfrak{F}$ and a finite disjoint union of finite brooms.*

(ii) *If a $\mathcal{PTL} \circ \mathcal{RC}$-formula $\varphi$ is satisfiable in a tt-model based on a flow of time $\mathfrak{F}$ then $\varphi$ is satisfiable in an Aleksandrov tt-model based on $\mathfrak{F}$ and a (possibly infinite) disjoint union of $\omega$-brooms.*

**Proof.** (i) By Lemma B.1 (i), $\varphi$ is satisfiable in an Aleksandrov tt-model based on $\mathfrak{F}$ and a finite quasi-order $\mathfrak{G}$. The rest of the proof is similar to that of Lemma A.1. Further details are left to the reader.

(ii) By Lemma B.1 (ii), $\varphi$ is satisfiable in an Aleksandrov tt-model based on $\mathfrak{F}$ and a quasi-order $\mathfrak{G} = \langle V, R \rangle$. The rest of the proof again is similar to that of Lemma A.1. We only note that although $\mathfrak{G}$ can be infinite, still for every $x \in V$ there is a $y \in V_0$ such that $xRy$. This is guaranteed by the condition that the set $A_{w,x,\top}$ has a maximal point. ❑

Observe that Aleksandrov spaces are essentially infinite in case (ii) of Lemma C.1 and a generalisation of Lemma A.2 does not go through. First, we can easily enforce a topological space to be infinite using the $\mathcal{PTL} \circ \mathcal{RCC}$-8 formula

$$\Box_F^+ \mathsf{NTTP}(p, \bigcirc p).$$

Moreover, the formula

$$\diamondsuit\!\!\!\!\exists(\lceil p \rceil \sqcap \overline{\mathbf{I} \lceil p \rceil}) \quad \wedge \quad \Box_F^+ \boxed{\forall}(\lceil p \rceil \sqsubset \lceil \bigcirc p \rceil) \quad \wedge \quad \Box_F^+ \boxed{\forall}\big((\lceil p \rceil \sqcap \overline{\mathbf{I} \lceil p \rceil}) \sqsubset (\overline{\mathbf{I} \lceil \bigcirc p \rceil} \sqcap \lceil \overline{p} \sqcap \bigcirc p \rceil)\big)$$

is satisfied in an Aleksandrov tt-model based on a single $\omega$-broom, but cannot be satisfied in an Aleksandrov tt-model based on a union of $n$-brooms for any finite $n$.

On the other hand, Aleksandrov tt-models based on disjoint unions of $n$-brooms, where $n$ is bounded by the width of the formula, are enough for spatio-temporal logics based on $\mathcal{RC}^-$. Recall that spatial terms $\tau$ of $\mathcal{PTL} \times \mathcal{RC}^-$ (and $\mathcal{PTL} \circ \mathcal{RC}^-$) are defined as follows

$$\begin{aligned}
\delta \quad &::= \quad \varrho \quad | \quad \overline{\sigma}, \\
\sigma \quad &::= \quad \mathbf{I}\varrho \quad | \quad \overline{\delta} \quad | \quad \sigma_1 \sqcap \sigma_2, \\
\tau \quad &::= \quad \overline{\delta_1 \sqcap \cdots \sqcap \delta_m} \quad | \quad \overline{\delta \sqcap \sigma} \quad | \quad \overline{\sigma},
\end{aligned}$$

where the $\varrho$ are spatio-temporal Boolean region terms of $\mathcal{PTL} \times \mathcal{BRCC}$-8 (and $\mathcal{PTL} \circ \mathcal{BRCC}$-8, respectively). It is not hard to see that, for every tt-model $\mathfrak{M} = \langle \mathfrak{F}, \mathfrak{T}, \mathfrak{V} \rangle$ with $\mathfrak{F} = \langle W, < \rangle$, $\mathfrak{T} = \langle U, \mathbb{I} \rangle$ and every $w \in W$, we have

$$\mathfrak{V}(\delta, w) = \mathbb{CI}\mathfrak{V}(\delta, w) \qquad \text{and} \qquad \mathfrak{V}(\sigma, w) = \mathbb{IC}\mathfrak{V}(\sigma, w), \tag{23}$$

i.e., the $\delta$ are always interpreted by regular closed sets, whereas the $\sigma$ by regular open ones.

We define the *width* $w(\varphi)$ of a $\mathcal{PTL} \times \mathcal{RC}^-$-formula $\varphi$ as the maximal number $m$ of conjuncts in its subformulas of the form $\boxtimes(\overline{\delta_1 \sqcap \cdots \sqcap \delta_m})$, if such subformulas exist, and put $w(\varphi) = 1$ otherwise.

**Lemma C.2.** (i) *If a $\mathcal{PTL} \times \mathcal{RC}^-$-formula $\varphi$ is satisfiable in a tt-model with* **FSA** *and based on a flow of time $\mathfrak{F}$ then $\varphi$ is satisfiable in an Aleksandrov tt-model based on $\mathfrak{F}$ and a finite disjoint union of $w(\varphi)$-brooms.*

(ii) *If a $\mathcal{PTL} \circ \mathcal{RC}^-$-formula $\varphi$ is satisfiable in a tt-model based on a flow of time $\mathfrak{F}$ then $\varphi$ is satisfiable in an Aleksandrov tt-model based on $\mathfrak{F}$ and a (possibly infinite) disjoint union of $w(\varphi)$-brooms.*

**Proof.** By Lemma C.1, we may assume that $\varphi$ is satisfied in an Aleksandrov tt-model $\mathfrak{M} = \langle \mathfrak{F}, \mathfrak{G}, \mathfrak{V} \rangle$, where $\mathfrak{F} = \langle W, < \rangle$ and $\mathfrak{G} = \langle V, R \rangle$ is a disjoint union of brooms (in (i), the union and the brooms are finite). Without loss of generality we may assume that $\varphi$ is composed (using temporal operators and the Booleans) from formulas of the set $\Sigma_\varphi = \{\diamondsuit\tau_1, \ldots, \diamondsuit\tau_n\}$,[7] where every $\tau_i$ has one of the following forms

$$\delta_1 \sqcap \cdots \sqcap \delta_m, \qquad \delta \sqcap \sigma \qquad \text{or} \qquad \sigma, \tag{24}$$

and the $\delta_i$, $\delta$ and $\sigma$ are as defined above.

For every $\diamondsuit\tau \in \Sigma_\varphi$ and every $w \in W$ with $(\mathfrak{M}, w) \models \diamondsuit\tau$, we fix a point $x_{\tau,w} \in V$ such that $(\mathfrak{M}, \langle w, x_{\tau,w} \rangle) \models \tau$. We may assume that the $x_{\tau,w}$ are pairwise distinct and that the roots of all the brooms are the points of the form $x_{\tau,w}$ for some $w \in W$ and $\diamondsuit\tau \in \Sigma_\varphi$. Therefore, $\mathfrak{G}$ is a disjoint union of brooms $\mathfrak{b}_{\tau,w}$, for $\diamondsuit\tau \in \Sigma_\varphi$ and $w \in W$.

Let us construct a model $\mathfrak{M}' = \langle \mathfrak{F}, \mathfrak{G}', \mathfrak{V}' \rangle$ as follows. Given a broom $\mathfrak{b}_{\tau,w}$, we delete some of its leaves depending on the form of $\tau$. Three cases are possible:

*Case* $\tau = \delta_1 \sqcap \cdots \sqcap \delta_m$: take $m$ leaves $y_1, \ldots, y_m$ of $\mathfrak{b}_{\tau,w}$ such that $(\mathfrak{M}, \langle w, y_i \rangle) \models \delta_i$ and $x_{\tau,w} R y_i$ for $i = 1, \ldots, m$ and remove all leaves different from $y_1, \ldots, y_m$.

*Case* $\tau = \delta \sqcap \sigma$: take a leaf $y$ of $\mathfrak{b}_{\tau,w}$ such that $(\mathfrak{M}, \langle w, y \rangle) \models \delta$ and $x_{\tau,w} R y$ and remove all other leaves. Note that, by (23), we have $(\mathfrak{M}, \langle w, y \rangle) \models \sigma$, and therefore $(\mathfrak{M}, \langle w, y \rangle) \models \tau$.

*Case* $\tau = \sigma$: take a leaf $y$ of $\mathfrak{b}_{\tau,w}$ such that $x_{\tau,w} R y$ and remove all other leaves. By (23), we have $(\mathfrak{M}, \langle w, y \rangle) \models \tau$.

Denote by $\mathfrak{b}'_{\tau,w}$ the resulting broom. Clearly, it is a $w(\varphi)$-broom. Let $\mathfrak{G}' = \langle V', R' \rangle$ be the disjoint union of all $\mathfrak{b}_{\tau,w}$, for $\diamondsuit\tau \in \Sigma_\varphi$ and $w \in W$. It should be clear that $\mathfrak{G}'$ is as required. Finally, we define $\mathfrak{V}'$ by taking for every spatial variable $p$, every $w \in W$ and every $x \in V'$,

$$x \in \mathfrak{V}'(p, w) \quad \text{iff} \quad \text{there is } y \in V' \text{ of depth 0 such that } xR'y \text{ and } y \in \mathfrak{V}(p, w).$$

7. We treat $\diamondsuit$ as primitive and $\boxtimes$ as an abbreviation.

To show that $\varphi$ is satisfied in $\mathfrak{M}'$, we first prove that, for all $w \in W$ and all $\diamond\tau \in \Sigma_\varphi$,

$$(\mathfrak{M}', w) \models \diamond\tau \qquad \text{iff} \qquad (\mathfrak{M}, w) \models \diamond\tau.$$

It is readily proved by induction that we have $(\mathfrak{M}', \langle w, x\rangle) \models \tau$ iff $(\mathfrak{M}, \langle w, x\rangle) \models \tau$, for all points $x \in V'$ of depth 0, all $w \in W$ and *all spatio-temporal terms* $\tau$.

Then, by the construction, we also have that, for all formulas $\diamond\tau \in \Sigma_\varphi$ and all $w \in W$, $(\mathfrak{M}, w) \models \diamond\tau$ implies $(\mathfrak{M}', w) \models \diamond\tau$. So it remains to show that $(\mathfrak{M}, w) \models \neg\diamond\tau$ implies $(\mathfrak{M}', w) \models \neg\diamond\tau$ for all $\diamond\tau \in \Sigma_\varphi$ and all $w \in W$. Suppose that we have $(\mathfrak{M}', \langle w, x\rangle) \models \tau$ and $(\mathfrak{M}, w) \models \neg\diamond\tau$. Consider three possible cases for $\tau$:

*Case* $\tau = \delta_1 \sqcap \cdots \sqcap \delta_m$. Then, for every $i$, $1 \leq i \leq m$, there is $y_i \in V'$ of depth 0 such that $xR'y_i$ and $(\mathfrak{M}', \langle w, y_i\rangle) \models \delta_i$. But then $(\mathfrak{M}, \langle w, y_i\rangle) \models \delta_i$ and, by (23), $(\mathfrak{M}, \langle w, x\rangle) \models \delta_i$. Therefore, $(\mathfrak{M}, \langle w, x\rangle) \models \tau$, contrary to $(\mathfrak{M}, w) \models \neg\diamond\tau$.

*Case* $\tau = \delta \sqcap \sigma$. Then there is $y \in V'$ of depth 0 such that $xR'y$, $(\mathfrak{M}', \langle w, y\rangle) \models \delta$ and, by (23), $(\mathfrak{M}', \langle w, y\rangle) \models \sigma$. Thus $(\mathfrak{M}', \langle w, y\rangle) \models \tau$. But then $(\mathfrak{M}, \langle w, y\rangle) \models \tau$, contrary to $(\mathfrak{M}, w) \models \neg\diamond\tau$.

*Case* $\tau = \sigma$. Then there is $y \in V'$ of depth 0 with $xR'y$ and, by (23), $(\mathfrak{M}', \langle w, y\rangle) \models \tau$. But then $(\mathfrak{M}, \langle w, y\rangle) \models \tau$, contrary to $(\mathfrak{M}, w) \models \neg\diamond\tau$.

Now, by a straightforward induction we can easily show that, for all $w \in W$ and all formulas $\psi$ built from $\Sigma_\varphi$ using the temporal operators and the Booleans,

$$(\mathfrak{M}', w) \models \psi \qquad \text{iff} \qquad (\mathfrak{M}, w) \models \psi.$$

It follows that $\varphi$ is satisfied in $\mathfrak{M}'$. ❑

## C.1 Lower Complexity Bounds (I)

**Proof of Theorem 3.10, lower bound.** The proof is by reduction of an arbitrary problem in 2EXPSPACE to the satisfiability problem of $\mathcal{PTL} \circ \mathcal{RC}$. Let $\boldsymbol{A}$ be a (single-tape, deterministic) Turing machine such that $\boldsymbol{A}$ halts on every input (accepting or rejecting it), and $\boldsymbol{A}$ uses $\leq 2^{2^{f(n)}}$ cells of the tape on any input of length $n$, for some polynomial $f$. Given any such Turing machine $\boldsymbol{A}$ and an input $x$ for it, we will construct a $\mathcal{PTL} \circ \mathcal{RC}$-formula $\varphi_{\boldsymbol{A},x}$ (using only future-time temporal operators) such that

(i) the length of $\varphi_{\boldsymbol{A},x}$ is polynomial in the size of $\boldsymbol{A}$ and $x$;

(ii) if $\varphi_{\boldsymbol{A},x}$ is satisfiable in a tt-model based on $\langle\mathbb{N}, <\rangle$ then $\boldsymbol{A}$ accepts $x$; and

(iii) if $\boldsymbol{A}$ accepts $x$ then $\varphi_{\boldsymbol{A},x}$ is satisfiable in a tt-model with **FSA** and based on $\langle\mathbb{N}, <\rangle$.

The case of $\langle\mathbb{Z}, <\rangle$ as a flow of time (with or without **FSA**) follows immediately. The case of finite flows of time can be proved by relativising the temporal operators of $\varphi_{\boldsymbol{A},x}$ (say, by a propositional variable $\mathsf{range}$ as in the proof of Theorem 3.6 in Appendix B.2 and in the proof of the lower bound of Theorem 3.9 below): we can obtain a formula $\varphi'_{\boldsymbol{A},x}$ such that $\varphi'_{\boldsymbol{A},x}$ is satisfiable in an Aleksandrov tt-model based on $\langle\mathbb{N}, <\rangle$ iff it is satisfiable in an Aleksandrov tt-model based on the same quasi-order but on a finite flow of time. So this way all the lower bound results of this theorem follow.

Given a Turing machine $\boldsymbol{A}$, polynomial $f$, and input $x = \langle x_1, \dots, x_n \rangle$ as above, let $d = f(n)$,

$$exp(1,d) = d \cdot 2^d \quad \text{and} \quad exp(2,d) = exp(1,d) \cdot 2^{exp(1,d)}.$$

Then we have

$$2^{2^{f(n)}} \leq exp(2,d). \tag{25}$$

Our plan is as follows. First, we will show that 'yardsticks' of length $exp(2,d)$ (similar to those used by Stockmeyer, 1974 or Halpern and Vardi, 1989) can be encoded by $\mathcal{PTL} \circ \mathcal{RC}$-formulas of length polynomial in $d$. These yardsticks will be used to define a temporal operator $\bigcirc^{exp(2,d)}$. Then, using this operator, we will encode the computation of $\boldsymbol{A}$ on input $x$.

By Lemma C.1 (ii), if a $\mathcal{PTL} \circ \mathcal{RC}$-formula $\varphi_{\boldsymbol{A},x}$ is satisfied in a tt-model based on a flow of time $\langle \mathbb{N}, < \rangle$, then it is satisfied in an Aleksandrov tt-model $\mathfrak{M} = \langle \langle \mathbb{N}, < \rangle, \mathfrak{G}, \mathfrak{U} \rangle$, where $\mathfrak{G} = \langle V, R \rangle$ is a disjoint union of $\omega$-brooms. Take such a model $\mathfrak{M}$ and suppose that the $\mathcal{PTL} \circ \mathcal{RC}$-formula[8]

$$\Box_F^+ \boxtimes \big( \lceil aux \rceil \equiv \lceil \bigcirc aux \rceil \big) \tag{26}$$

is true in $\mathfrak{M}$ at moment 0. Since region $\lceil aux \rceil$ does not change over time, we can divide all points in $V$ into three disjoint sets: *external*, *boundary* and *internal* points with respect to $\lceil aux \rceil$—i.e., those satisfying

$$ep(aux) = \overline{\lceil aux \rceil}, \qquad bp(aux) = \lceil aux \rceil \sqcap \overline{\mathbf{I} \lceil aux \rceil} \qquad \text{and} \qquad \mathbf{I} \lceil aux \rceil,$$

respectively. Note that every boundary point has a non-boundary $R$-successor, so boundary points can only be of depth 1. In what follows we simply speak about external and boundary points not mentioning 'with respect to $\lceil aux \rceil$.'

We define the '$\bigcirc^{exp(2,d)}$ operator' by a $\mathcal{PTL} \circ \mathcal{RC}$-formula of length polynomial in $d$ as follows:

(a) First, we 'encode' yardsticks of length $d$. We will use different formulas for yardsticks on external points and for yardsticks on boundary points.

(b) Then, with the help of $d$-yardsticks, we 'encode' yardsticks of length $exp(1,d)$. We will again use different formulas for external and boundary points.

(c) Next, with the help of $exp(1,d)$-yardsticks on both boundary and external points, we 'encode' yardsticks of length $exp(2,d)$ on boundary points.

(d) Finally, with the help of $exp(2,d)$-yardsticks on boundary points, we define a polynomial-length '$\bigcirc^{exp(2,d)}$ operator' applicable to propositional variables.

*Step* (a). Suppose that (26) and the following formula hold in $\mathfrak{M}$ at 0:

$$\Box_F^+ \boxtimes \big( bp(aux) \sqsubset \delta_{0,d} \big) \ \wedge \ \Box_F^+ \boxtimes \big( ep(aux) \sqsubset \delta_{0,d}^{ext} \big) \tag{27}$$

where

$$\delta_{0,d} = \big( \lceil delim_0 \rceil \equiv \lceil \bigcirc^d delim_0 \rceil \big) \ \sqcap \ \Big( \lceil delim_0 \rceil \sqsubset \prod_{j=1}^{d-1} \overline{\lceil \bigcirc^j delim_0 \rceil} \Big),$$

8. Recall that $\tau_1 \equiv \tau_2$ stands for $(\tau_1 \sqsubset \tau_2) \sqcap (\tau_2 \sqsubset \tau_1)$. We assume that $\sqcap$ and $\sqcup$ bind stronger than $\equiv$.

and $\delta_{0,d}^{ext}$ results from $\delta_{0,d}$ by replacing each occurrence of $delim_0$ with $ext_delim_0$.

Take a boundary point $z$. Suppose that $v \in \mathbb{N}$ is such that $(\mathfrak{M}, \langle v, x\rangle) \models \lceil delim_0 \rceil$. By $\delta_{0,d}$, for every time moment $w \geq v$,

$$(\mathfrak{M}, \langle w, z\rangle) \models \lceil delim_0 \rceil \qquad \text{iff} \qquad w \equiv v \pmod{d},$$

that is, on $z$, $\lceil delim_0 \rceil$ holds once in every $d$ time instants, starting from $v$. By the second conjunct of (27), external points of $\lceil aux \rceil$ behave similarly with respect to $\lceil ext_delim_0 \rceil$.

*Step* (b). To encode yardsticks of length $exp(1, d)$, recall first that every number $a < 2^d$ can be represented in binary by a sequence $a_0 \dots a_{d-1}$ of bits. We will 'mark' the bits of binary numbers by a region term $\lceil bit_1 \rceil$ as follows. Given a boundary point $z$ and a time moment $v$ such that $(\mathfrak{M}, \langle v, z\rangle) \models \lceil delim_0 \rceil$, we say that an interval $[w, w + d - 1]$, for some $w = v + j \cdot d$, $j \in \mathbb{N}$, *encodes a number* $a < 2^d$ *on* $z$, if for every $i < d$,

$$(\mathfrak{M}, \langle w + i, z\rangle) \models \lceil bit_1 \rceil \qquad \text{iff} \qquad a_i = 1.$$

Recall that the binary representation $b_0 \dots b_{d-1}$ is the successor of $a_0 \dots a_{d-1}$ modulo $2^d$ if the following holds: for all $i$, $0 \leq i < d$, we have $a_i = b_i$ iff $a_j = 0$, for some $j$, $i < j < d$. We will use the $d$-intervals starting from $v$ to encode $< 2^d$ numbers in such a way that consecutive intervals encode consecutive (modulo $2^d$) numbers, starting from 0.

So, suppose that (26), (27) and the following formula hold in $\mathfrak{M}$ at 0:

$$\Box_F^+ \boxtimes \big(bp(aux) \sqsubset (\gamma_{1,d} \sqcap \delta_{1,d})\big) \wedge \Box_F^+ \boxtimes \big(ep(aux) \sqsubset (\gamma_{1,d}^{ext} \sqcap \delta_{1,d}^{ext})\big), \tag{28}$$

where

$$\begin{aligned}
\gamma_{1,d} \quad &= \quad \big(\lceil lwr_1 \rceil \;\equiv\; \lceil \bigcirc delim_0 \rceil \sqcup (\lceil \bigcirc bit_1 \rceil \sqcap \lceil \bigcirc lwr_1 \rceil)\big) \sqcap \\
&\quad\;\; \big(\lceil zr_1 \rceil \;\equiv\; \overline{\lceil bit_1 \rceil} \sqcap (\lceil \bigcirc delim_0 \rceil \sqcup \lceil \bigcirc zr_1 \rceil)\big) \sqcap \big(\lceil delim_1 \rceil \;\equiv\; \lceil delim_0 \rceil \sqcap \lceil zr_1 \rceil\big), \\
\delta_{1,d} \quad &= \quad \big(\overline{\lceil lwr_1 \rceil} \;\equiv\; (\lceil bit_1 \rceil \equiv \lceil \bigcirc^d bit_1 \rceil)\big),
\end{aligned}$$

and both $\gamma_{1,d}^{ext}$ and $\delta_{1,d}^{ext}$ result from $\gamma_{1,d}$ and $\delta_{1,d}$, respectively, by attaching prefix $ext_$ to all of their spatial variables (save $aux$).

Take a boundary point $z$. Suppose that $v \in \mathbb{N}$ is such that $(\mathfrak{M}, \langle v, z\rangle) \models \lceil delim_1 \rceil$. Then, by the last conjunct of $\gamma_{1,d}$, we have $(\mathfrak{M}, \langle v, z\rangle) \models \lceil delim_0 \rceil$. Since, by (a), $\lceil delim_0 \rceil$ holds once in every $d$ time instants on $z$, $\lceil delim_0 \rceil$ 'marks' the starting moment of each $d$-interval. Then, by the first conjunct of $\gamma_{1,d}$, for every $i$, $0 \leq i < d$, we have[9]

$$(\mathfrak{M}, \langle v + i, z\rangle) \models \lceil lwr_1 \rceil \qquad \text{iff} \qquad (\mathfrak{M}, \langle v + j, z\rangle) \models \lceil bit_1 \rceil, \text{ for all } j,\ i < j < d.$$

Therefore, $\delta_{1,d}$ says that consecutive $< 2^d$ numbers (starting with 0) are encoded by consecutive $d$-intervals (starting from $v$). Similarly to the first conjunct of $\gamma_{1,d}$, its second conjunct ensures that, for every $i$, $0 \leq i < d$,

$$(\mathfrak{M}, \langle v + i, z\rangle) \models \lceil zr_1 \rceil \qquad \text{iff} \qquad (\mathfrak{M}, \langle v + j, z\rangle) \models \overline{\lceil bit_1 \rceil}, \text{ for all } j,\ i \leq j < d.$$

9. Since we cannot apply the $\mathcal{U}$ operator to form spatio-temporal terms, auxiliary regions are used instead: $\lceil lwr_1 \rceil \equiv \lceil \bigcirc delim_0 \rceil \sqcup (\lceil \bigcirc bit_1 \rceil \sqcap \lceil \bigcirc lwr_1 \rceil)$ ensures that $\lceil lwr_1 \rceil$ behaves as $\lceil bit_1 \rceil \,\mathcal{U}\, \lceil delim_0 \rceil$. This equality term can indeed be regarded as a fixed point characterisation of the $\mathcal{U}$ operator. Note also that we do not need to require (as we should do for the fixed point characterisation) $\lceil lwr_1 \rceil \sqsubset \Diamond_F \lceil delim_0 \rceil$ to be true because the eventuality is already enforced by $\delta_{0,d}$.

So, by the last conjunct of $\gamma_{1,d}$, $\lceil delim_1 \rceil$ holds on $z$ once in every $exp(1,d) = d \cdot 2^d$ time instants, starting from $v$. By the second conjunct of (28), external points behave similarly with respect to $\lceil ext_delim_1 \rceil$.

*Step* (c). Now we construct yardsticks of length $exp(2,d)$, using the $exp(1,d)$-yardsticks constructed in (b). Suppose that (26)–(28) and the following formulas hold in $\mathfrak{M}$ at 0:

$$\Box_F^+ \boxdot \big(bp(aux) \sqsubset \lceil ext_delim_1 \rceil\big) \wedge \Box_F^+ \boxdot \big(\lceil ext_delim_1 \rceil \sqsubset (ep(aux) \sqcup bp(aux))\big), \quad (29)$$

$$\Box_F^+ \boxdot \big(ep(aux) \sqsubset \eta_{1,d}(bit_2)\big), \quad (30)$$

$$\Box_F^+ \boxdot \big(bp(aux) \sqsubset \gamma_{2,d} \sqcap \delta_{2,d}\big), \quad (31)$$

where $\gamma_{2,d}$ is defined similarly to $\gamma_{1,d}$ and

$$\begin{aligned}
\eta_{1,d}(bit_2) &= \big(\lceil jm_1_bit_2 \rceil \equiv \\
&\qquad \big(\lceil \bigcirc ext_delim_1 \rceil \sqcap \lceil \bigcirc bit_2 \rceil\big) \sqcup \big(\overline{\lceil \bigcirc ext_delim_1 \rceil} \sqcap \lceil \bigcirc jm_1_bit_2 \rceil\big)\big),\\
\delta_{2,d} &= \big(\lceil bit_2 \rceil \equiv \mathbf{I}\lceil bit_2 \rceil\big) \sqcap \big(\overline{\lceil lwr_2 \rceil} \equiv \big(\lceil bit_2 \rceil \equiv \mathbf{J}_{1,d}\, bit_2\big)\big),\\
\mathbf{J}_{1,d}\, bit_2 &= \mathbf{I}\lceil (\overline{aux} \sqcap ext_delim_1) \sqsubset jm_1_bit_2 \rceil.
\end{aligned}$$

Take a boundary point $z$. Suppose $v$ is a time moment such that $(\mathfrak{M}, \langle v, z\rangle) \models \lceil delim_2 \rceil$. Then, by the last conjunct of $\gamma_{2,d}$, $(\mathfrak{M}, \langle v, z\rangle) \models \lceil delim_1 \rceil$. We know from (b) that $\lceil delim_1 \rceil$ holds on $z$ once in every $exp(1,d)$ time instants starting from $v$. So, by $\delta_{2,d}$ and the first conjunct of $\gamma_{2,d}$ we intend to express that consecutive $< 2^{exp(1,d)}$ numbers (starting with 0) are encoded by consecutive $exp(1,d)$-intervals starting from $v$. If we could do this then, by the last conjunct of $\gamma_{2,d}$, $\lceil delim_2 \rceil$ would hold on $z$ once in every $exp(2,d)$ time instants starting from $v$. The only problem (and the only difference from step (b)) is that to 'mark' the bits of $< 2^{exp(1,d)}$ binary numbers by a term $\lceil bit_2 \rceil$, we need to show that the (polynomial length) term $\mathbf{J}_{1,d}\, bit_2$ actually defines '$\bigcirc^{exp(1,d)} \lceil bit_2 \rceil$' in the sense that, for every $w \geq v$,

$$(\mathfrak{M}, \langle w, z\rangle) \models \mathbf{J}_{1,d}\, bit_2 \quad \text{iff} \quad (\mathfrak{M}, \langle w + exp(1,d), z\rangle) \models \lceil bit_2 \rceil. \quad (32)$$

Suppose first that $(\mathfrak{M}, \langle w, z\rangle) \models \mathbf{J}_{1,d}\, bit_2$. Then, by (29), there is an external $R$-successor $y_w$ of $z$ (of depth 0) such that $(\mathfrak{M}, \langle w, y_w\rangle) \models \lceil ext_delim_1 \rceil$, and so $(\mathfrak{M}, \langle w, y_w\rangle) \models \lceil jm_1_bit_2 \rceil$. On the other hand, it is not hard to see that if $(\mathfrak{M}, \langle w, z\rangle) \not\models \mathbf{J}_{1,d}\, bit_2$, then there is an external $R$-successor $y'_w$ of $z$ (of depth 0) such that $(\mathfrak{M}, \langle w, y'_w\rangle) \models \lceil ext_delim_1 \rceil$ but $(\mathfrak{M}, \langle w, y'_w\rangle) \models \overline{\lceil jm_1_bit_2 \rceil}$.

In both cases, it is readily checked that if $(\mathfrak{M}, \langle w, y\rangle) \models \lceil ext_delim_1 \rceil$, for some external point $y$, then, by (30),

$$(\mathfrak{M}, \langle w, y\rangle) \models \lceil jm_1_bit_2 \rceil \quad \text{iff} \quad (\mathfrak{M}, \langle w + exp(1,d), y\rangle) \models \lceil bit_2 \rceil.$$

Now (32) follows by the first conjunct of $\delta_{2,d}$.

*Step* (d). We are now in a position to define a polynomial-length '$\bigcirc^{exp(2,d)}$ operator' $\mathbf{J}_{2,d}$ applicable to propositional variables. Recall that a propositional variable $\mathsf{p}$ stands for spatial formula $\boxdot p$, where $p$ is a spatial variable associated with $\mathsf{p}$. Now, for every propositional variable $\mathsf{p}$ we intend to apply the new operator to, we introduce a fresh spatial variable $jm_2_p$. Suppose that (26)–(31) and the following formulas hold in $\mathfrak{M}$ at 0:

$$\Box_F^+ \diamondsuit\big(bp(aux) \sqcap \lceil delim_2 \rceil\big), \tag{33}$$

$$\Box_F^+\big(\boxdot\lceil p\rceil \leftrightarrow \neg\boxdot\overline{\lceil p\rceil}\big) \;\wedge\; \Box_F^+\boxdot\, \eta_{2,d}(p), \tag{34}$$

where $\eta_{2,d}(p)$ is obtained by replacing $bit_2$, $jm_1_bit_2$ and $ext_delim_1$ in $\eta_{1,d}(bit_2)$ with $p$, $jm_2_p$ and $delim_2$, respectively. Let

$$\mathbf{J}_{2,d}\,\mathsf{p} = \boxdot\big((bp(aux) \sqcap \lceil delim_2\rceil) \;\sqsubset\; \lceil jm_2_p\rceil\big).$$

We claim that, for every time moment $w$ and every propositional variable $\mathsf{p}$,

$$(\mathfrak{M}, w) \models \mathbf{J}_{2,d}\,\mathsf{p} \qquad \text{iff} \qquad (\mathfrak{M}, w + exp(2,d)) \models \mathsf{p}. \tag{35}$$

Suppose first that $(\mathfrak{M}, w) \models \mathbf{J}_{2,d}\,\mathsf{p}$. Then, by (33), there is a boundary point $z$ such that $(\mathfrak{M}, \langle w, z\rangle) \models \lceil delim_2\rceil$, and therefore $(\mathfrak{M}, \langle w, z\rangle) \models \lceil jm_2_p\rceil$. On the other hand, if $(\mathfrak{M}, w) \not\models \mathbf{J}_{2,d}\,\mathsf{p}$, then there is a boundary point $z'$ with $(\mathfrak{M}, \langle w, z'\rangle) \models \lceil delim_2\rceil$ but $(\mathfrak{M}, \langle w, z'\rangle) \models \overline{\lceil jm_2_p\rceil}$. In both cases, it is readily checked that if $(\mathfrak{M}, \langle w, z\rangle) \models \lceil delim_2\rceil$, for some boundary point $z$, then, by the second conjunct of (34),

$$(\mathfrak{M}, \langle w, z\rangle) \models \lceil jm_2_p\rceil \qquad \text{iff} \qquad (\mathfrak{M}, \langle w + exp(2,d), z\rangle) \models \lceil p\rceil.$$

Now (35) follows by the first conjunct of (34).

Finally, we are in a position to define the $\mathcal{PTL} \circ \mathcal{RC}$-formula $\varphi_{\boldsymbol{A},x}$ that encodes the computation of Turing machine $\boldsymbol{A}$ on input $x$. Let $A$ be the tape alphabet (with the blank symbol $b \in A$) and $S$ the set of states (with two halt states $s_{yes}$ and $s_{no}$ in $S$) of $\boldsymbol{A}$. We use the symbol $£ \notin A$ to mark the left end of the tape. We know that the space used by $\boldsymbol{A}$ on input $x = \langle x_1, \dots, x_n\rangle$ is $\leq 2^{2^{f(n)}}$, which is $\leq exp(2,d)$ by (25). So we can represent each configuration of the computation of $\boldsymbol{A}$ on $x$ as a finite word

$$\langle £, a_1, \dots, a_{i-1}, \langle s, a_i\rangle, a_{i+1}, \dots, a_m, b, \dots, b\rangle$$

of length $exp(2,d)$, where $a_1, \dots, a_m \in A$ and $\langle s, a_i\rangle \in S \times A$ represents the current state and the active cell. The transition function $\delta$ of $\boldsymbol{A}$ takes triples of the form $\langle a_i, \langle s, a_j\rangle, a_k\rangle$ (for $a_i \in A \cup \{£\}$, $a_j, a_k \in A$, $s \in S - \{s_{yes}, s_{no}\}$) to similar triples. For instance,

$$\delta(a_i, \langle s, a_j\rangle, a_k) = \langle a_i, a_j, \langle s', a_k\rangle\rangle$$

means that, when being in state $s$ and reading symbol $a_j$, the new state should be $s'$ and the head should move one cell to the right. We also assume that, for all $a_i \in A \cup \{£\}$ and $a_j, a_k \in A$, we have $\delta(a_i, \langle s_{yes}, a_j\rangle, a_k) = \langle a_i, a_j, a_k\rangle$ and $\delta(a_i, \langle s_{no}, a_j\rangle, a_k) = \langle a_i, a_j, a_k\rangle$ meaning that the head is removed after $\boldsymbol{A}$ is being halted.

Now, for every $\alpha \in A \cup \{£\} \cup (S \times A)$, we introduce a fresh propositional variable $\mathsf{p}_\alpha$. Let $\varphi_{\boldsymbol{A},x}$ be the conjunction of (26)–(31), (33) and an instance of (34), for each $\mathsf{p}_\alpha$, as well

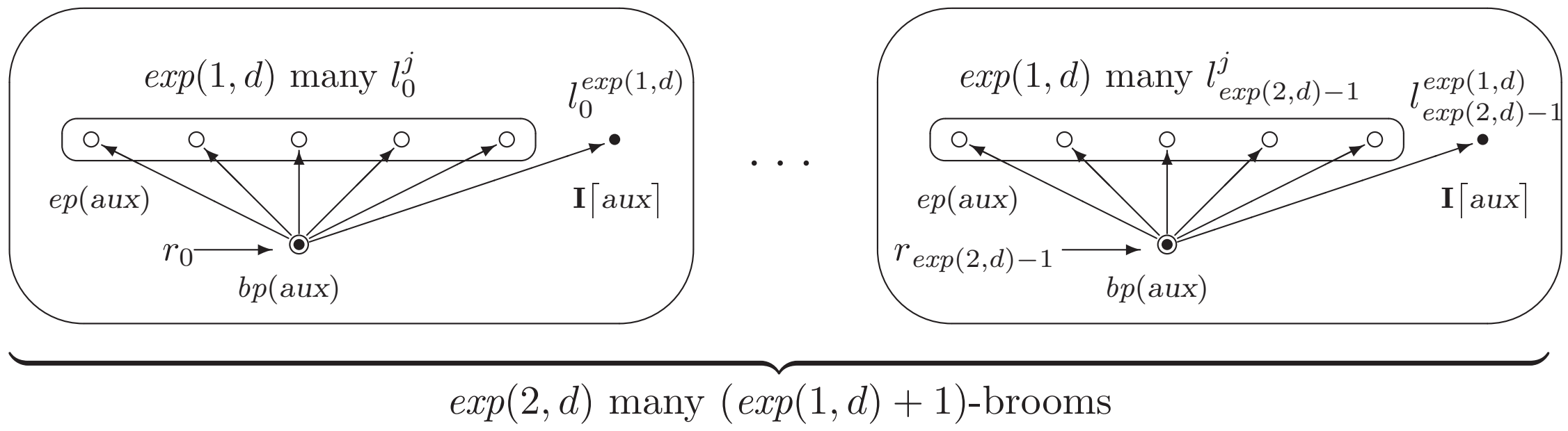

Figure 8: Structure of yardsticks.

as the following formulas:

$$\bigwedge_{\substack{\alpha,\beta\in A\cup\{\pounds\}\cup S\times A\\ \alpha\neq\beta}} \Box_F^+(\neg \mathsf{p}_\alpha \vee \neg \mathsf{p}_\beta), \tag{36}$$

$$\mathsf{p}_\pounds \wedge \bigcirc(\mathsf{p}_{\langle s_0,x_1\rangle} \wedge \bigcirc(\mathsf{p}_{x_2} \wedge \bigcirc(\cdots \wedge \bigcirc(\mathsf{p}_{x_n} \wedge \mathsf{p}_b \,\mathcal{U}\, \mathsf{p}_\pounds)\cdots))), \tag{37}$$

$$\Box_F^+\Big(\bigcirc\mathsf{af_head} \leftrightarrow \bigvee_{\langle s,a\rangle\in S\times A} \mathsf{p}_{\langle s,a\rangle}\Big), \tag{38}$$

$$\bigwedge_{\substack{\delta(\alpha,\beta,\gamma)=\\ \langle\alpha',\beta',\gamma'\rangle}} \Box_F^+\Big(\bigcirc\bigcirc\mathsf{af_head} \to \big(\mathsf{p}_\alpha \to \mathbf{J}_{2,d}\,\mathsf{p}_{\alpha'}\big) \wedge \bigcirc\big(\mathsf{p}_\beta \to \mathbf{J}_{2,d}\,\mathsf{p}_{\beta'}\big) \wedge \bigcirc\bigcirc\big(\mathsf{p}_\gamma \to \mathbf{J}_{2,d}\,\mathsf{p}_{\gamma'}\big)\Big), \tag{39}$$

$$\bigwedge_{a\in A\cup\{\pounds\}} \Box_F^+\Big(\neg\big(\bigcirc\bigcirc\mathsf{af_head} \vee \bigcirc\mathsf{af_head} \vee \mathsf{af_head}\big) \to \big(\mathsf{p}_a \to \mathbf{J}_{2,d}\,\mathsf{p}_a\big)\Big), \tag{40}$$

$$\Diamond_F \bigvee_{a\in A} \mathsf{p}_{\langle s_{yes},a\rangle} \;\wedge\; \neg\Diamond_F \bigvee_{a\in A} \mathsf{p}_{\langle s_{no},a\rangle}. \tag{41}$$

Suppose first that $\varphi_{\boldsymbol{A},x}$ holds in $\mathfrak{M}$ at time moment 0. By (36)–(40) and (35), the consecutive configurations of the computation of $\boldsymbol{A}$ on input $x$ are properly 'encoded' along the time axis. (For instance, $\mathsf{p}_\pounds$ holds once in every $exp(2,d)$ time moments.) Finally, (41) says that $\boldsymbol{A}$ accepts input $x$.

Conversely, suppose that $\boldsymbol{A}$ accepts input $x$. We will define an Aleksandrov tt-model $\mathfrak{M} = \langle\langle\mathbb{N},<\rangle, \mathfrak{G}, \mathfrak{U}\rangle$ with **FSA** that satisfies $\varphi_{\boldsymbol{A},x}$. Let the partial order $\mathfrak{G} = \langle V, R\rangle$ be a disjoint union of $exp(2,d)$ many $(exp(1,d)+1)$-brooms (see Fig. 8):

$$V = \{r_i \mid i < exp(2,d)\} \cup \{l_i^j \mid i < exp(2,d),\ j \leq exp(1,d)\},$$
$$zRy \qquad \text{iff} \qquad z = y \quad \text{or} \quad z = r_i,\ y = l_i^j, \text{ for some } i, j.$$

Suppose that the number of steps in the computation of $\boldsymbol{A}$ on $x$ is $m$. Then $\mathfrak{M}$ will have a prefix of length $N = m \cdot exp(2,d)$ after which the final configuration (without a halting state) repeats to infinity. For $w \in \mathbb{N}$, let

$$\mathfrak{U}(w, \mathrm{aux}) = \{l_i^{exp(1,d)} \mid i < exp(2,d)\}.$$

Then it is easy to see that the boundary points are the $r_i$, and the external points are the $l_i^j$, for $i < exp(1,d)$. Now put, for every $w \in \mathbb{N}$,

$$\begin{aligned}
\mathfrak{U}(w, delim_2) &= \{l_i^{exp(1,d)} \mid i \equiv w \pmod{exp(2,d)}\},\\
\mathfrak{U}(w, delim_1) &= \{l_i^{exp(1,d)} \mid i \equiv w \pmod{exp(1,d)}\},\\
\mathfrak{U}(w, delim_0) &= \{l_i^{exp(1,d)} \mid i \equiv w \pmod{d}\},\\
\mathfrak{U}(w, ext_delim_1) &= \{l_i^{v} \mid v \equiv w \pmod{exp(1,d)},\ i < exp(2,d)\},\\
\mathfrak{U}(w, ext_delim_0) &= \{l_i^{v} \mid v \equiv w \pmod{d},\ i < exp(2,d)\}.
\end{aligned}$$

The valuations for the other variables should be clear. We then have

$$\begin{aligned}
(\mathfrak{M}, \langle w, z\rangle) \models \lceil delim_2 \rceil \quad &\text{iff} \quad z = r_i \text{ or } z = l_i^{exp(1,d)} \text{ for some } i \equiv w \pmod{exp(2,d)},\\
(\mathfrak{M}, \langle w, z\rangle) \models \lceil delim_1 \rceil \quad &\text{iff} \quad z = r_i \text{ or } z = l_i^{exp(1,d)} \text{ for some } i \equiv w \pmod{exp(1,d)},\\
(\mathfrak{M}, \langle w, z\rangle) \models \lceil ext_delim_1 \rceil \quad &\text{iff} \quad z = r_i \text{ or } z = l_i^{v} \text{ for some } v \equiv w \pmod{exp(2,d)}\\
&\qquad\qquad \text{and } i < exp(2,d),
\end{aligned}$$

and so on, as required. It is not hard to see that $\mathfrak{M}$ satisfies **FSA** and $(\mathfrak{M}, 0) \models \varphi_{\boldsymbol{A},x}$. ❑

**Proof of Theorem 3.9, lower bound.** The proof is by reduction of the $2^n$-*corridor tiling problem* which is known to be EXPSPACE-complete (Chlebus, 1986; van Emde Boas, 1997). The problem can be formulated as follows: given an instance $\mathcal{T} = \langle T, t_0, t_1, n\rangle$, where $T$ is a finite set of tile types, $t_0, t_1 \in T$ and $n > 0$, decide whether there is an $m \in \mathbb{N}$ such that $\mathcal{T}$ *tiles* the $m \times 2^n$-grid (or corridor) in such a way that $t_0$ is placed onto $\langle 0, 0\rangle$, $t_1$ onto $\langle m-1, 0\rangle$, and the top and bottom sides of the corridor are of some fixed colour, say, *white*.

Suppose $\mathcal{T} = \langle T, t_0, t_1, n\rangle$ is given. Our aim is to construct (using only future-time temporal operators) a $\mathcal{PTL} \circ \mathcal{BRCC}$-8 formula $\varphi_{\mathcal{T}}$ such that

(i) the length of $\varphi_{\mathcal{T}}$ is a polynomial function of $|T|$ and $n$;

(ii) if $\varphi_{\mathcal{T}}$ is satisfiable in a tt-model based on $\langle \mathbb{N}, <\rangle$ then there is $m \in \mathbb{N}$ such that $\mathcal{T}$ tiles the $m \times 2^n$-corridor;

(iii) if there is $m \in \mathbb{N}$ such that $\mathcal{T}$ tiles the $m \times 2^n$-corridor, then $\varphi_{\mathcal{T}}$ is satisfiable in a tt-model with **FSA** and based on $\langle \mathbb{N}, <\rangle$;

(iv) $\varphi_{\mathcal{T}}$ is satisfiable in a tt-model based on $\langle \mathbb{N}, <\rangle$ iff it is satisfiable in a tt-model based on a finite flow of time.

The case of $\langle \mathbb{Z}, <\rangle$ follows immediately.

Recall that, by Lemma C.2 (ii), if $\varphi_{\mathcal{T}}$ is satisfied in a tt-model then it is satisfied in an Aleksandrov tt-model $\mathfrak{M} = \langle \langle \mathbb{N}, <\rangle, \mathfrak{G}, \mathfrak{V}\rangle$, where $\mathfrak{G} = \langle V, R\rangle$ is a disjoint union of $\omega$-brooms. To explain the meaning of $\varphi_{\mathcal{T}}$'s subformulas, we assume that such a model $\mathfrak{M}$ is given. Throughout the proof we use only a restricted subset of $\mathcal{RCC}$-8 predicates: for spatio-temporal terms $\tau_1$ and $\tau_2$ constructed from spatial variables using only the complement, the intersection and the next-time operator $\bigcirc$, we need $\mathsf{EQ}(\tau_1, \tau_2)$ as well as two abbreviations $\mathsf{P}(\tau_1, \tau_2) = \mathsf{EQ}(\tau_1, \tau_2) \vee \mathsf{TPP}(\tau_1, \tau_2) \vee \mathsf{NTPP}(\tau_1, \tau_2)$ and $\mathsf{E}\,\tau_1 = \neg\mathsf{DC}(\tau_1, \tau_1)$ standing for '$\tau_1$ is a part of $\tau_2$' and '$\tau_1$ is nonempty,' respectively. Clearly, this language forms a subset of $\mathcal{PTL} \circ \mathcal{BRCC}$-8 (in fact, as we show in Remark C.3 below, the proof goes through for an even more restricted subset of the langauge).

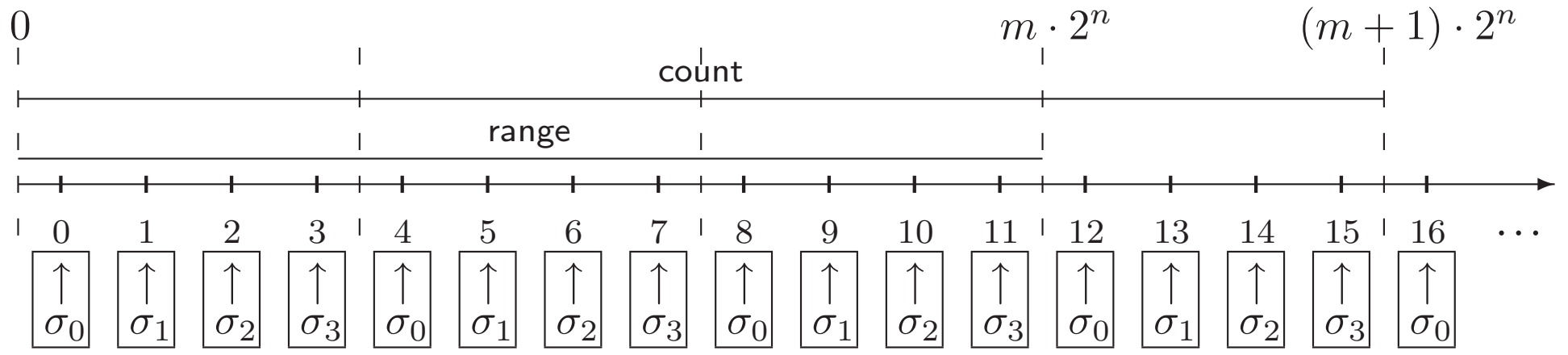

Figure 9: Counting formulas for $m = 3$ and $n = 2$.

Our first step in the construction of $\varphi_{\mathcal{T}}$ (which will contain, among many others, spatial variables $t$ for all $t \in T$) is to write down formulas forcing a sequence $y_0, y_1, \dots, y_{m\cdot 2^n-1}$ of distinct points (of depth 0) from $V$, for some $m \in \mathbb{N}$, such that, for each $i < m \cdot 2^n$, $(\mathfrak{M}, \langle i, y_i \rangle) \models t$ for a unique tile type $t \in T$. If $i = k \cdot 2^n + j$, for some $k < m$, $j < 2^n$, then we will use $y_i$ (at time $i$) to encode the pair $\langle k, j \rangle$ of the $m \times 2^n$-grid. Thus, the up neighbour $\langle k, j+1 \rangle$ of $\langle k, j \rangle$ will be coded by the point $y_{i+1}$ at time $i+1$, while its right neighbour $\langle k+1, j \rangle$ by $y_{i+2^n}$ at moment $i + 2^n$ (see Fig. 10).

Let $\mathsf{q}_0, \dots, \mathsf{q}_{n-1}$ be pairwise distinct propositional variables and

$$\sigma_j = \mathsf{q}_0^{d_0} \wedge \dots \wedge \mathsf{q}_{n-1}^{d_{n-1}},$$

where $d_{n-1} \dots d_0$ is the binary representation of $j < 2^n$, $\mathsf{q}_i^0 = \neg \mathsf{q}_i$ and $\mathsf{q}_i^1 = \mathsf{q}_i$, for each $i$. Suppose that the formula

$$\big(\mathsf{count} \wedge \sigma_0\big) \;\wedge\; \big(\mathsf{count}\, \mathcal{U}\, (\sigma_0 \wedge \Box_F^+ \neg \mathsf{count})\big) \;\wedge\; \Box_F^+\big(\mathsf{count} \to \chi\big) \tag{42}$$

is true in $\mathfrak{M}$ at 0, where $\mathsf{count}$ is a fresh propositional variable and $\chi$ is the following 'counting' formula (the length of which is polynomial in $n$)

$$\chi = \bigwedge_{k<n} \Big(\Big(\bigwedge_{i<k} \mathsf{q}_i \wedge \neg \mathsf{q}_k\Big) \to \Big(\bigwedge_{i<k} \bigcirc \neg \mathsf{q}_i \wedge \bigcirc \mathsf{q}_k \wedge \bigwedge_{k<i<n} (\mathsf{q}_i \leftrightarrow \bigcirc \mathsf{q}_i)\Big)\Big) \;\wedge\; \big(\sigma_{2^n-1} \to \bigcirc \sigma_0\big).$$

Then there is an $m \in \mathbb{N}$ such that $\mathsf{count}$ is true before moment $(m+1)\cdot 2^n$ and false starting from $(m+1)\cdot 2^n$. The sequence $\sigma_0, \sigma_1, \dots, \sigma_{2^n-1}$ is repeated $m+1$ times along the time-line, i.e., while $\mathsf{count}$ is true. Let

$$\mathsf{range} = \Diamond_F(\mathsf{count} \wedge \sigma_0).$$

Clearly, $\mathsf{range}$ is true before moment $m \cdot 2^n$ and then always false (see Fig. 9).

Let $equ$, $p_0, \dots, p_{n-1}$ and $e_0, \dots, e_{n-1}$ be fresh distinct spatial variables, and

$$\pi_j = p_0^{d_0} \sqcap \dots \sqcap p_{n-1}^{d_{n-1}},$$

where $d_{n-1} \dots d_0$ is the binary representation of $j < 2^n$, $p_i^0 = \overline{p_i}$ and $p_i^1 = p_i$, for each $i$. Suppose that (42) and

$$\Box_F^+ \mathsf{EQ}\big(equ, \bigsqcap_{i<n} e_i\big) \;\wedge\; \bigwedge_{i<n} \Box_F^+\big(\mathsf{q}_i \leftrightarrow \mathsf{EQ}(e_i,\ p_i)\big) \;\wedge\; \bigwedge_{i<n} \Box_F^+ \mathsf{EQ}\big(p_i, \bigcirc p_i\big) \tag{43}$$

are true in $\mathfrak{M}$ at 0. Then, by the first two conjuncts of (43), for all $i \in \mathbb{N}$ and $y \in V$ of depth 0, there is $j < 2^n$ such that $(\mathfrak{M}, \langle i, y\rangle) \models \mathit{equ}$ iff $(\mathfrak{M}, i) \models \sigma_j$ and $(\mathfrak{M}, \langle i, y\rangle) \models \pi_j$. By the last conjunct of (43), we then have

$$(\mathfrak{M}, \langle i, y\rangle) \models \mathit{equ} \quad \text{iff} \quad \exists j < 2^n \Big((\mathfrak{M}, i) \models \sigma_j \quad \text{and} \quad (\mathfrak{M}, \langle k, y\rangle) \models \pi_j, \ \text{for all } k \in \mathbb{N}\Big). \quad (*)$$

We can generate the required sequence of points $y_i$ using the formulas:

$$\mathsf{range} \ \wedge \ \Box_F^+(\mathsf{range} \to \mathsf{E}\, \mathit{tile}), \tag{44}$$

$$\Box_F^+\mathsf{EQ}\Big(\mathit{tile},\ \mathit{equ} \sqcap (\bigsqcup_{t \in T} t) \sqcap \bigsqcap_{t \in T} \mathit{no_t_in_future}\Big), \tag{45}$$

$$\bigwedge_{t \in T} \Box_F^+\mathsf{P}\big(\mathit{no_t_in_future},\ \bigcirc\bar{t} \sqcap \bigcirc \mathit{no_t_in_future}\big), \tag{46}$$

where *tile* and the *no_t_in_future* (for all $t \in T$) are fresh spatial variables. Indeed, suppose the conjunction of (42)–(46) holds at time 0 in $\mathfrak{M}$. Then, by the first conjunct of (44) and (42), $(\mathfrak{M}, 0) \models \mathsf{range} \wedge \sigma_0$ and, by the second conjunct of (44), $(\mathfrak{M}, \langle 0, y_0\rangle) \models \mathit{tile}$ for some $y_0 \in V$. We may assume that $y_0$ is of depth 0. Then, by (45), we have

($a_0$) $(\mathfrak{M}, \langle 0, y_0\rangle) \models \mathit{equ}$, and, by $(*)$, $(\mathfrak{M}, \langle k, y_0\rangle) \models \pi_0$ for all $k \in \mathbb{N}$;

($b_0$) for all $t \in T$, $(\mathfrak{M}, \langle 0, y_0\rangle) \models \mathit{no_t_in_future}$ and, by (46), $(\mathfrak{M}, \langle k, y_0\rangle) \models \bar{t}$ for all $k > 0$.

Next, by (42), we have $(\mathfrak{M}, 1) \models \mathsf{range} \wedge \sigma_1$ and, by (44), there is $y_1 \in V$ (again, of depth 0) such that $(\mathfrak{M}, \langle 1, y_1\rangle) \models \mathit{tile}$. In particular, we have:

($a_1$) $(\mathfrak{M}, \langle 1, y_1\rangle) \models \mathit{equ}$, and, by $(*)$, $(\mathfrak{M}, \langle k, y_1\rangle) \models \pi_1$ for all $k \in \mathbb{N}$;

($b_1$) for all $t \in T$, $(\mathfrak{M}, \langle 1, y_1\rangle) \models \mathit{no_t_in_future}$ and, by (46), $(\mathfrak{M}, \langle k, y_1\rangle) \models \bar{t}$ for all $k > 1$.

By ($b_0$), $(\mathfrak{M}, \langle 1, y_0\rangle) \models \bar{t}$, for all $t \in T$, and thus $y_1 \neq y_0$. Now we consider $y_1$ at moment 1 and use the same argument to find a point $y_2 \in V$ (which is different from $y_1$ by ($b_1$)), and so forth; see Fig. 10. This gives us points $y_0, y_1, \dots, y_{m \cdot 2^n - 1}$ (of depth 0) from $V$ we need.

Our next aim is to write down formulas that could serve as pointers to the up and right neighbours of a given pair in the corridor (at this moment we do not bother about its top border). Consider the formulas

$$\Box_F^+\mathsf{EQ}\big(\mathit{up},\ \bigcirc \mathit{tile}\big), \tag{47}$$

$$\Box_F^+\mathsf{EQ}\big(\mathit{right},\ \mathit{equ} \sqcap \mathit{no_equ_U_tile}\big), \tag{48}$$

$$\Box_F^+\mathsf{EQ}\big(\mathit{no_equ_U_tile},\ \bigcirc \mathit{tile} \sqcup \bigcirc\big(\overline{\mathit{equ}} \sqcap \mathit{no_equ_U_tile}\big)\big), \tag{49}$$

where *up*, *right* and *no_equ_U_tile* are fresh spatial variables. We claim that, for all $i, j < m \cdot 2^n$,

- $(\mathfrak{M}, \langle i, y_j\rangle) \models \mathit{up}$ iff $j = i + 1$;
- $(\mathfrak{M}, \langle i, y_j\rangle) \models \mathit{right}$ iff $j = i + 2^n$.

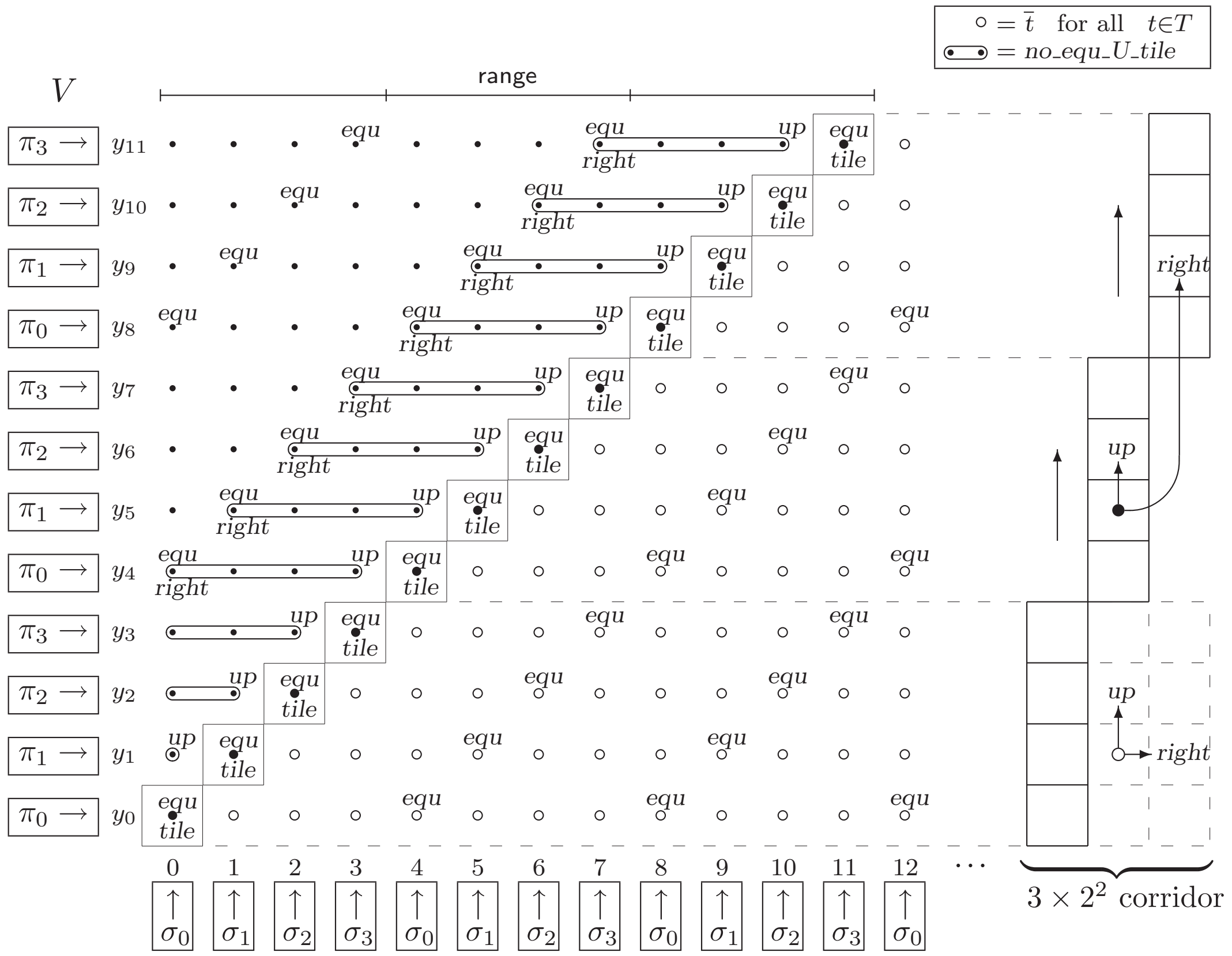

Figure 10: Satisfying $\varphi_{\mathcal{T}}$, $n = 2$, in a tt-model based on space with $3 \cdot 2^2$ points.

The former is obvious. Let us prove the latter. To show that $(\mathfrak{M}, \langle i, y_j\rangle) \models$ *right*, for $j = i + 2^n$, we first observe that $(\mathfrak{M}, \langle j, y_j\rangle) \models$ *equ* and $(\mathfrak{M}, \langle i, y_j\rangle) \models$ *equ* by $(*)$. It follows from $(\mathfrak{M}, \langle j, y_j\rangle) \models$ *tile* by (49) that $(\mathfrak{M}, \langle j-1, y_j\rangle) \models$ *no_equ_U_tile*. Then, applying (49) (from right to left) sufficiently many times, we obtain $(\mathfrak{M}, \langle i, y_j\rangle) \models$ *no_equ_U_tile*, $(\mathfrak{M}, \langle i-1, y_j\rangle) \not\models$ *no_equ_U_tile*, and so $(\mathfrak{M}, \langle i, y_j\rangle) \models$ *right*.

Conversely, suppose that $(\mathfrak{M}, \langle i, y_j\rangle) \models$ *right* for some $y_j$. Then $(\mathfrak{M}, \langle i, y_j\rangle) \models$ *equ* and, by $(*)$ (note that $i + 2^n < (m+1) \cdot 2^n$, and so count is still true at $i + 2^n$),

$$(\mathfrak{M}, \langle i + 2^n, y_j\rangle) \models \mathit{equ}. \qquad (**)$$

We have $(\mathfrak{M}, \langle i, y_j\rangle) \models$ *no_equ_U_tile*. Then applying (49) (from left to right) sufficiently many times we arrive at $(\mathfrak{M}, \langle i + 2^n - 1, y_j\rangle) \models$ *no_equ_U_tile* which together with $(**)$ gives $(\mathfrak{M}, \langle i + 2^n, y_j\rangle) \models$ *tile*. But then $j = i + 2^n$.

It should be noted that at every time point the extension of *no_equ_U_tile* coincides with the extension of the term $\overline{equ}\,\mathcal{U}\,\mathit{tile}$ on elements of the sequence $y_0, \ldots, y_{m\cdot 2^n}$, and that (49) is indeed the fixed point characterisation of this $\mathcal{U}$ operator.

Finally, the formulas below ensure that every point of the $m \times 2^n$-corridor is covered by at most one tile, $\langle 0, 0\rangle$ is covered by $t_0$, $\langle m-1, 0\rangle$ by $t_1$, the top and bottom sides are *white* and that the colours on adjacent edges of adjacent tiles match:

$$\bigwedge_{\substack{t,t'\in T\\ t\neq t'}} \Box_F^+ \neg \mathsf{E}\,(t \sqcap t'), \tag{50}$$

$$\mathsf{P}(\mathit{tile},\ t_0) \ \wedge\ \Box_F^+\Big((\sigma_0 \wedge \mathsf{range}) \wedge \neg\Diamond_F(\sigma_0 \wedge \mathsf{range}) \to \mathsf{P}\big(\mathit{tile},\ t_1\big)\Big), \tag{51}$$

$$\Box_F^+\Big(\sigma_{2^n-1} \to \bigvee_{\substack{t\in T\\ \mathit{up}(t)=\mathit{white}}} \mathsf{P}\big(\mathit{tile},\ t\big)\Big), \tag{52}$$

$$\Box_F^+\Big(\sigma_0 \to \bigvee_{\substack{t\in T\\ \mathit{down}(t)=\mathit{white}}} \mathsf{P}\big(\mathit{tile},\ t\big)\Big), \tag{53}$$

$$\bigwedge_{\substack{t,t'\in T\\ \mathit{up}(t)\neq \mathit{down}(t')}} \Box_F^+\Big(\neg\sigma_{2^n-1} \ \wedge\ \mathsf{E}\,t \ \to\ \mathsf{P}\big(\mathit{up},\ \mathit{no_t'_in_future}\big)\Big), \tag{54}$$

$$\bigwedge_{\substack{t,t'\in T\\ \mathit{right}(t)\neq \mathit{left}(t')}} \Box_F^+\Big(\mathsf{E}\,t \ \to\ \mathsf{P}\big(\mathit{right},\ \mathit{no_t'_in_future}\big)\Big). \tag{55}$$

Let $\varphi_{\mathcal{T}}$ be the conjunction of (42)–(55). Suppose that $\varphi_{\mathcal{T}}$ holds at 0 in $\mathfrak{M}$. Then there is $m \in \mathbb{N}$ such that $(\mathfrak{M}, m \cdot 2^n - 1) \models \mathsf{range}$ and, for every $i \geq m \cdot 2^n$, $(\mathfrak{M}, i) \models \neg\mathsf{range}$. Then we define a map $\boldsymbol{\tau}\colon m \times 2^n \to T$ by taking

$$\boldsymbol{\tau}(k,j) = t \qquad \text{iff} \qquad (\mathfrak{M}, \langle i, y_i\rangle) \models t \text{ and } i = k \cdot 2^n + j.$$

We leave it to the reader to check that $\boldsymbol{\tau}$ is a tiling of $m \times 2^n$, as required.

For the other direction, suppose that there is a tiling $\boldsymbol{\tau}$ of the $m \times 2^n$-corridor by $\mathcal{T}$, for some $m > 0$. Then $\varphi_{\mathcal{T}}$ is satisfied in the Aleksandrov tt-model $\mathfrak{M} = \langle \langle \mathbb{N}, < \rangle, \langle V, R\rangle, \mathfrak{V}\rangle$, where $V = \{y_0, \dots, y_{m\cdot 2^n - 1}\}$, $R$ is the minimal reflexive relation on $V$,

$$\mathfrak{V}(t,i) = \{y_i \in V \mid \boldsymbol{\tau}(k,j) = t \text{ and } i = k \cdot 2^n + j\},$$

and the other variables of $\varphi_{\mathcal{T}}$ are interpreted as shown in Fig. 10. Clearly, $\mathfrak{M}$ satisfies **FSA**. Moreover, $\varphi_{\mathcal{T}}$ is satisfiable in tt-models over finite flows of time iff it is satisfiable in tt-models over $\langle \mathbb{N}, <\rangle$. Details are left to the reader. ❑

*Remark* C.3. It may be of interest to note that the language used in the proof above is rather limited. In fact, it is enough to extend the PSPACE-complete logic $\mathcal{PTL} \circ \mathcal{RCC}$-8 with predicates of the form $\mathsf{EQ}(\varrho_1, \varrho_2 \sqcup \varrho_3)$ (where the $\varrho_i$ are *atomic* spatio-temporal region terms) to make it EXPSPACE-hard. To show this, we transform the $\mathcal{PTL}\circ\mathcal{BRCC}$-8 formula $\varphi_{\mathcal{T}}$ constructed above in the following way. First, we take a fresh spatial variable $u$ (denoting 'the universe') and add to $\varphi_{\mathcal{T}}$ the conjunct $\Box_F^+\mathsf{EQ}(u, \bigcirc u)$. Next, for every spatio-temporal Boolean region term $\varrho$ of $\varphi_{\mathcal{T}}$, we introduce a spatial variable $\mathit{neg_}\varrho$ ('the complement of $\varrho$ with respect to $u$'), add to $\varphi_{\mathcal{T}}$ conjuncts $\Box_F^+\mathsf{EQ}(u, \varrho \sqcup \mathit{neg_}\varrho) \wedge \Box_F^+\mathsf{DC}(\varrho, \mathit{neg_}\varrho)$, and replace every occurrence of $\overline{\varrho}$ in the resulting formula with $\mathit{neg_}\varrho$. Finally, for every spatio-temporal term of the form $\varrho_1 \sqcap \varrho_2$, we introduce a fresh spatial variable $\varrho_1\mathit{_and_}\varrho_2$, add the conjuncts

$$\Box_F^+\mathsf{P}(\varrho_1\mathit{_and_}\varrho_2,\ \varrho_1) \ \wedge\ \Box_F^+\mathsf{P}(\varrho_1\mathit{_and_}\varrho_2,\ \varrho_2) \ \wedge\ \Box_F^+\mathsf{P}(\varrho_1,\ \mathit{neg_}\varrho_2 \sqcup \varrho_1\mathit{_and_}\varrho_2)$$

and replace occurrences of $\varrho_1 \sqcap \varrho_2$ with $\varrho_1_and_\varrho_2$. One can readily see that (i) the length of the resulting formula $\varphi'_{\mathcal{T}}$ is linear in the length of $\varphi_{\mathcal{T}}$ and (ii) $\varphi'_{\mathcal{T}}$ is satisfiable in a tt-model based on $\langle \mathbb{N}, < \rangle$ (with **FSA**) iff $\varphi_{\mathcal{T}}$ is satisfiable in a tt-model based on $\langle \mathbb{N}, < \rangle$ (with **FSA**).

## C.2 Upper Complexity Bounds (I): Quasimodels for $\mathcal{PTL} \times \mathcal{RC}$

In this appendix we define quasimodels for $\mathcal{PTL} \times \mathcal{RC}$ in the spirit of the paper (Hodkinson et al., 2000) in order to establish the upper complexity bounds of Theorems 3.10 and 3.13. We remind the reader that spatio-temporal terms of $\mathcal{PTL} \times \mathcal{RC}$ are of the form:

$$\begin{aligned} \tau \quad &::= \quad \varrho \quad | \quad \mathbf{I}\varrho \quad | \quad \overline{\tau} \quad | \quad \tau_1 \sqcap \tau_2, \\ \varrho \quad &::= \quad \mathbf{CI}p \quad | \quad \mathbf{CI}\overline{\varrho} \quad | \quad \mathbf{CI}(\varrho_1 \sqcap \varrho_2) \quad | \quad \mathbf{CI}(\varrho_1 \,\mathcal{U}\, \varrho_2) \quad | \quad \mathbf{CI}(\varrho_1 \,\mathcal{S}\, \varrho_2), \end{aligned}$$

and that $\mathcal{PTL} \circ \mathcal{RC}$ forms a sublanguage of $\mathcal{PTL} \times \mathcal{RC}$—it differs from the latter only in the definition of spatio-temporal *region* terms:

$$\varrho \quad ::= \quad \mathbf{CI}p \quad | \quad \mathbf{CI}\overline{\varrho} \quad | \quad \mathbf{CI}(\varrho_1 \sqcap \varrho_2) \quad | \quad \mathbf{CI}\bigcirc\varrho.$$

Let $\varphi$ be a $\mathcal{PTL} \times \mathcal{RC}$-formula. Recall from p. 200 that by $sub\,\varphi$ we denote the set of all subformulas of $\varphi$ and by $term\,\varphi$ the set of all its spatio-temporal terms including those of the form $\tau$ and $\varrho$. A *type* $\boldsymbol{t}$ for $\varphi$ is a subset of $term\,\varphi$ such that

- for every $\tau_1 \sqcap \tau_2 \in term\,\varphi$, $\quad \tau_1 \sqcap \tau_2 \in \boldsymbol{t}$ iff $\tau_1 \in \boldsymbol{t}$ and $\tau_2 \in \boldsymbol{t}$;
- for every $\overline{\tau} \in term\,\varphi$, $\quad \overline{\tau} \in \boldsymbol{t}$ iff $\tau \notin \boldsymbol{t}$.

Clearly, the number $\flat(\varphi)$ of different types for $\varphi$ is bounded by $2^{|term\,\varphi|}$.

A *broom type* $\mathfrak{b}$ for $\varphi$ is a pair $\langle \langle T, \leq \rangle, \boldsymbol{t} \rangle$, where $\langle T, \leq \rangle$ is a broom (with $T^0$ being its leaves) and $\boldsymbol{t}$ a labelling function associating with each $x \in T$ a type $\boldsymbol{t}(x)$ for $\varphi$ such that the following conditions hold:

**(bt0)** $\boldsymbol{t}(x) \neq \boldsymbol{t}(y)$, for each pair of distinct points $x, y \in T^0$;

**(bt1)** for every $x \in T^0$,

- for every $\mathbf{CI}(\varrho_1 \sqcap \varrho_2) \in term\,\varphi$, $\quad \mathbf{CI}(\varrho_1 \sqcap \varrho_2) \in \boldsymbol{t}(x)$ iff $\varrho_1 \in \boldsymbol{t}(x)$ and $\varrho_2 \in \boldsymbol{t}(x)$,
- and for every $\mathbf{CI}\overline{\varrho} \in term\,\varphi$, $\quad \mathbf{CI}\overline{\varrho} \in \boldsymbol{t}(x)$ iff $\varrho \notin \boldsymbol{t}(x)$;

**(bt2)** for every $\mathbf{I}\varrho \in term\,\varphi$, $\quad \mathbf{I}\varrho \in \boldsymbol{t}(x)$ iff $\varrho \in \boldsymbol{t}(y)$ for every $y \in T$, $x \leq y$;

**(bt3)** for every $\varrho \in term\,\varphi$, $\quad \varrho \in \boldsymbol{t}(x)$ iff $\exists y \in T^0$ with $x \leq y$ and $\varrho \in \boldsymbol{t}(y)$.

Broom types $\mathfrak{b}_1 = \langle \langle T_1, \leq_1 \rangle, \boldsymbol{t}_1 \rangle$ and $\mathfrak{b}_2 = \langle \langle T_2, \leq_2 \rangle, \boldsymbol{t}_2 \rangle$ for $\varphi$ are said to be *isomorphic* if

- for every $x_1 \in T_1^0$, there is $x_2 \in T_2^0$ such that $\boldsymbol{t}_1(x_1) = \boldsymbol{t}_2(x_2)$ and
- for every $x_2 \in T_2^0$, there is $x_1 \in T_1^0$ such that $\boldsymbol{t}_1(x_1) = \boldsymbol{t}_2(x_2)$.

Clearly, given two isomorphic broom types $\mathfrak{b}_1$ and $\mathfrak{b}_2$, we also have $\boldsymbol{t}_1(r_1) = \boldsymbol{t}_2(r_2)$, where $r_1$ and $r_2$ are the roots of $\mathfrak{b}_1$ and $\mathfrak{b}_2$, respectively.

A *quasistate* for $\varphi$ is a pair $\langle \boldsymbol{s}, \mathfrak{m} \rangle$, where $\boldsymbol{s}$ is a Boolean-saturated subset of $sub\,\varphi$ and $\mathfrak{m}$ a disjoint union $\langle \langle T, \leq \rangle, \boldsymbol{t} \rangle$ of broom types $\mathfrak{b}_1, \ldots, \mathfrak{b}_n$ for $\varphi$ such that the following conditions hold:

**(qs0)** $\mathfrak{b}_i$ and $\mathfrak{b}_j$ are not isomorphic, for $i \neq j$;

**(qs1)** for every $\boxplus\tau \in sub\,\varphi$, $\quad \boxplus\tau \in \boldsymbol{s}$ iff $\quad \tau \in \boldsymbol{t}(x)$ for every $x \in T$.

Clearly, the number $\sharp(\varphi)$ of quasistates for $\varphi$ is bounded by $2^{2^{\flat(\varphi)}} \cdot 2^{|sub\,\varphi|}$.

Fix a flow of time $\mathfrak{F} = \langle W, < \rangle$. A *basic structure* for $\varphi$ is a pair $\langle \mathfrak{F}, \boldsymbol{q} \rangle$, where $\boldsymbol{q}$ is a function associating with each $w \in W$ a quasistate $\boldsymbol{q}(w) = \langle \boldsymbol{s}_w, \mathfrak{m}_w \rangle$ for $\varphi$ such that, for each $w \in W$,

- for every $\psi_1 \,\mathcal{U}\, \psi_2 \in sub\,\varphi$, $\psi_1 \,\mathcal{U}\, \psi_2 \in \boldsymbol{s}_w$ iff there is $v > w$ such that $\psi_2 \in \boldsymbol{s}_v$ and $\psi_1 \in \boldsymbol{s}_u$ for all $u \in (w, v)$;
- for every $\psi_1 \,\mathcal{S}\, \psi_2 \in sub\,\varphi$, $\psi_1 \,\mathcal{S}\, \psi_2 \in \boldsymbol{s}_w$ iff there is $v < w$ such that $\psi_2 \in \boldsymbol{s}_v$ and $\psi_1 \in \boldsymbol{s}_u$ for all $u \in (v, w)$.

Let $\langle \mathfrak{F}, \boldsymbol{q} \rangle$ be a basic structure for $\varphi$, where $\boldsymbol{q}(w) = \langle \boldsymbol{s}_w, \mathfrak{m}_w \rangle$ and $\mathfrak{m}_w = \langle \langle T_w, \leq_w \rangle, \boldsymbol{t}_w \rangle$ for $w \in W$. Denote by $T_w^0$ the set of all leaves in $\langle T_w, \leq_w \rangle$ and by $T_w^1$ the set of all roots of brooms in it. A 1*-run through* $\langle \mathfrak{F}, \boldsymbol{q} \rangle$ is a function $r$ giving for each $w \in W$ a point $r(w) \in T_w^1$; a *coherent and saturated* 0*-run through* $\langle \mathfrak{F}, \boldsymbol{q} \rangle$ is a function $r$ giving for each $w \in W$ a point $r(w) \in T_w^0$ such that the following conditions hold:

- for every $\mathbf{CI}(\varrho_1 \,\mathcal{U}\, \varrho_2) \in term\,\varphi$, $\mathbf{CI}(\varrho_1 \,\mathcal{U}\, \varrho_2) \in \boldsymbol{t}_w(r(w))$ iff there is $v > w$ such that $\varrho_2 \in \boldsymbol{t}_v(r(v))$ and $\varrho_1 \in \boldsymbol{t}_u(r(u))$ for all $u \in (w, v)$;
- for every $\mathbf{CI}(\varrho_1 \,\mathcal{S}\, \varrho_2) \in term\,\varphi$, $\mathbf{CI}(\varrho_1 \,\mathcal{S}\, \varrho_2) \in \boldsymbol{t}_w(r(w))$ iff there is $v < w$ such that $\varrho_2 \in \boldsymbol{t}_v(r(v))$ and $\varrho_1 \in \boldsymbol{t}_u(r(u))$ for all $u \in (v, w)$.

Say that a quadruple $\mathfrak{Q} = \langle \mathfrak{F}, \boldsymbol{q}, \mathfrak{R}, \lhd \rangle$ is a *quasimodel for* $\varphi$ *based on* $\mathfrak{F}$ if $\langle \mathfrak{F}, \boldsymbol{q} \rangle$ is a basic structure for $\varphi$, $\mathfrak{R} = \mathfrak{R}^0 \cup \mathfrak{R}^1$, with $\mathfrak{R}^1$ being a set of 1-runs and $\mathfrak{R}^0$ a set of coherent and saturated 0-runs through $\langle \mathfrak{F}, \boldsymbol{q} \rangle$, and $\lhd$ the reflexive closure of a subset of $\mathfrak{R}^1 \times \mathfrak{R}^0$ such that

**(qm2)** $\exists w_0 \in W \ \varphi \in \boldsymbol{s}_{w_0}$;

**(qm3)** for every $w \in W$ and every $x \in T_w$, there is $r \in \mathfrak{R}$ with $r(w) = x$;

**(qm4)** for all $r, r' \in \mathfrak{R}$, if $r \lhd r'$ then $r(w) \leq_w r'(w)$ for all $w \in W$;

**(qm5)** for all $r \in \mathfrak{R}$, $w \in W$ and $x \in T_w^0$, if $r(w) \leq_w x$ then there is $r' \in \mathfrak{R}^0$ such that $r'(w) = x$ and $r \lhd r'$.

A quasimodel $\mathfrak{Q}$ is said to be *finitary* if the set $\mathfrak{R}$ of runs is finite.

**Lemma C.4.** *A* $\mathcal{PTL} \times \mathcal{RC}$*-formula* $\varphi$ *is satisfiable in an Aleksandrov tt-model based on a flow of time* $\mathfrak{F}$ *and a (finite) disjoint union of (finite) brooms iff there is a (finitary) quasimodel for* $\varphi$ *based on* $\mathfrak{F}$.

**Proof.** ($\Leftarrow$) Let $\varphi$ be a $\mathcal{PTL} \times \mathcal{RC}$-formula and $\mathfrak{Q} = \langle \mathfrak{F}, \boldsymbol{q}, \mathfrak{R}, \lhd \rangle$ a quasimodel for $\varphi$, where $\mathfrak{F} = \langle W, < \rangle$ and $\boldsymbol{q}(w) = \langle \boldsymbol{s}_w, \langle \langle T_w, \leq_w \rangle, \boldsymbol{t}_w \rangle \rangle$ for $w \in W$. We construct an Aleksandrov tt-model $\mathfrak{M} = \langle \mathfrak{F}, \mathfrak{G}, \mathfrak{V} \rangle$ by taking $\mathfrak{G} = \langle \mathfrak{R}, \lhd \rangle$ and, for each spatial variable $p$ and $w \in W$,

$$\mathfrak{V}(p, w) = \{r \mid \mathbf{CI}p \in \boldsymbol{t}_w(r(w))\}.$$

Clearly, if $\mathfrak{Q}$ is finitary then $\mathfrak{G}$ is finite. Thus, it remains to prove that $\varphi$ is satisfied in $\mathfrak{M}$.

First, we show by induction on the construction of a *region* term $\varrho \in term\varphi$ that, for every $w \in W$ and every $r \in \mathfrak{R}$,

$$(\mathfrak{M}, \langle w, r\rangle) \models \varrho \qquad \text{iff} \qquad \varrho \in \boldsymbol{t}_w(r(w)). \tag{56}$$

The *basis* of induction: $\varrho = \mathbf{CI}p$. Let $(\mathfrak{M}, \langle w, r\rangle) \models \varrho$. Then there is $r' \in \mathfrak{R}$ such that $r \lhd r'$ and $(\mathfrak{M}, \langle w, r'\rangle) \models \mathbf{I}p$. By **(qm4)**, $r(w) \leq_w r'(w)$. Take any $y \in T_w^0$, $r'(w) \leq_w y$. By **(qm5)**, there is a run $r'' \in \mathfrak{R}^0$ such that $r' \lhd r''$ and $r''(w) = y$. Then $(\mathfrak{M}, \langle w, r''\rangle) \models p$ and, by the definition of $\mathfrak{V}$, $\mathbf{CI}p \in \boldsymbol{t}_w(r''(w))$ and, by **(bt3)**, $\varrho \in \boldsymbol{t}_w(r(w))$.

Conversely, if $\varrho \in \boldsymbol{t}_w(r(w))$ then, by **(bt3)**, there is $y \in T_w^0$ with $r(w) \leq_w y$ and $\varrho \in \boldsymbol{t}_w(y)$. By **(qm5)**, there is $r'' \in \mathfrak{R}^0$, $r \lhd r''$, such that $r''(w) = y$. Then $\mathbf{CI}p \in \boldsymbol{t}_w(r''(w))$ and, by the definition of $\mathfrak{V}$, $(\mathfrak{M}, \langle w, r''\rangle) \models p$. Therefore, $(\mathfrak{M}, \langle w, r\rangle) \models \varrho$.

The induction *steps* for $\varrho = \mathbf{CI}\overline{\varrho_1}$, $\mathbf{CI}(\varrho_1 \sqcap \varrho_2)$, $\mathbf{CI}(\varrho_1 \,\mathcal{U}\, \varrho_2)$ and $\mathbf{CI}(\varrho_1 \,\mathcal{S}\, \varrho_2)$ are similar, but instead of the definition of $\mathfrak{V}$, we use **(bt1)** for the cases of the Booleans and coherence and saturatedness of $r''$ for the cases of temporal operators.

Next, we extend (56) to arbitrary spatio-temporal terms $\tau \in term\varphi$.

*Case* $\tau = \mathbf{I}\varrho$. Suppose that $(\mathfrak{M}, \langle w, r\rangle) \models \mathbf{I}\varrho$. Take any $y \in T_w$, $r(w) \leq_w y$. If $y \in T_w^0$ then, by **(qm5)**, there is $r' \in \mathfrak{R}^0$ such that $r \lhd r'$ and $r'(w) = y$. If $y \notin T_w^0$ then clearly $y = r(w)$ and take $r' = r$. We have $(\mathfrak{M}, \langle w, r'\rangle) \models \varrho$, which, by IH, implies $\varrho \in \boldsymbol{t}_w(r'(w))$. Therefore, $\varrho \in \boldsymbol{t}_w(y)$ for every $y \geq_w r(w)$ and, by **(bt2)**, $\mathbf{I}\varrho \in \boldsymbol{t}_w(r(w))$.

Conversely, if $\mathbf{I}\varrho \in \boldsymbol{t}_w(r(w))$ then, by **(bt2)**, $\varrho \in \boldsymbol{t}_w(y)$, for every $y \geq_w r(w)$. Take any run $r' \in \mathfrak{R}$ such that $r \lhd r'$. By **(qm4)**, $r(w) \leq_w r'(w)$, and so $\varrho \in \boldsymbol{t}_w(r'(w))$, from which, by IH, $(\mathfrak{M}, \langle w, r'\rangle) \models \varrho$. Hence, $(\mathfrak{M}, \langle w, r\rangle) \models \mathbf{I}\varrho$.

*Cases* $\tau = \tau_1 \sqcap \tau_2$ and $\overline{\tau_1}$ follow from IH by the definition of type.

Finally, we show by induction on the construction of $\psi \in sub\varphi$ that, for every $w \in W$,

$$(\mathfrak{M}, w) \models \psi \qquad \text{iff} \qquad \psi \in \boldsymbol{s}_w. \tag{57}$$

*Case* $\psi = \boxtimes\tau$. Suppose $(\mathfrak{M}, w) \models \boxtimes\tau$. Take any $x \in T_w$. By **(qm3)**, there is $r \in \mathfrak{R}$ such that $r(w) = x$. Then $(\mathfrak{M}, \langle w, r\rangle) \models \tau$ and, by IH, $\tau \in \boldsymbol{t}_w(r(w))$. Therefore, by **(qs1)**, $\boxtimes\tau \in \boldsymbol{s}_w$. Conversely, let $\boxtimes\tau \in \boldsymbol{s}_w$. Take any run $r \in \mathfrak{R}$. By **(qs1)**, we have $\tau \in \boldsymbol{t}_w(r(w))$, from which, by IH, $(\mathfrak{M}, \langle w, r\rangle) \models \tau$. Hence, $(\mathfrak{M}, w) \models \boxtimes\tau$.

*Cases* $\psi = \psi_1 \wedge \psi_2$ and $\neg\psi_1$ follow from IH by the Boolean-saturatedness of the $\boldsymbol{s}_w$.

It follows from (57) and **(qm2)** that $\varphi$ is satisfiable in $\mathfrak{M}$.

($\Rightarrow$) Let $\varphi$ be a $\mathcal{PTL} \times \mathcal{RC}$-formula and suppose that $\varphi$ is satisfied in an Aleksandrov tt-model $\mathfrak{M} = \langle \mathfrak{F}, \mathfrak{G}, \mathfrak{V}\rangle$, where $\mathfrak{F} = \langle W, <\rangle$ and $\mathfrak{G} = \langle \Delta, \leq\rangle$ is a disjoint union of brooms.

Denote by $\Delta^0$ and $\Delta^1$ the leaves and the roots of brooms in $\mathfrak{G}$, respectively. With every pair $\langle w, x\rangle \in W \times \Delta$ we associate the type

$$\boldsymbol{t}(w, x) = \{\tau \in term\varphi \mid (\mathfrak{M}, \langle w, x\rangle) \models \tau\}.$$

Fix a $w \in W$ and define a binary relation on $\Delta$ as follows. For $x, x' \in \Delta^0$, let $x \sim_w x'$ iff $\boldsymbol{t}(w, x) = \boldsymbol{t}(w, x')$ and, for $z, z' \in \Delta^1$, let $z \sim_w z'$ iff the brooms generated by $z$ and $z'$ are isomorphic, i.e.,

$$\forall x \in \Delta^0 \,(z \leq x \;\rightarrow\; \exists x' \in \Delta^0 \,(z' \leq x' \;\wedge\; x \sim_w x')) \;\wedge$$
$$\forall x' \in \Delta^0 \,(z' \leq x' \;\rightarrow\; \exists x \in \Delta^0 \,(z \leq x \;\wedge\; x \sim_w x')).$$

Clearly, $\sim_w$ is an equivalence relation on $\Delta$. Denote by $[x]_w$ the $\sim_w$-equivalence class of $x$ and define a map $f_w$ by taking, for each $x \in \Delta$,

$$f_w(x) = \begin{cases} [x]_w, & x \in \Delta^1, \\ \langle [z]_w, [x]_w \rangle, & x \in \Delta^0 \quad \text{and} \quad z \in \Delta^1 \quad \text{such that } z \leq x. \end{cases}$$

Since $\mathfrak{G}$ is a disjoint union of brooms, $f_w$ is well-defined. Now put

$$T_w = \{f_w(x) \mid x \in \Delta\},$$
$$u \leq_w v \quad \text{iff} \quad \exists x, y \in \Delta \quad \text{such that } x \leq y,\ u = f_w(x) \ \text{and } v = f_w(y),$$
$$\boldsymbol{t}_w(f_w(x)) = \boldsymbol{t}(w, x), \quad \text{for } x \in \Delta.$$

By definition of $f_w$, $\langle T_w, \leq_w \rangle$ is a union of brooms and $\boldsymbol{t}_w$ is well-defined. Consider the structure $\langle \boldsymbol{s}_w, \mathfrak{m}_w \rangle$, where

$$\mathfrak{m}_w = \langle \langle T_w, \leq_w \rangle, \boldsymbol{t}_w \rangle \qquad \text{and} \qquad \boldsymbol{s}_w = \{\psi \in sub\,\varphi \mid (\mathfrak{M}, w) \models \psi\}.$$

It is readily seen that for each of the brooms of $\mathfrak{m}_w$ we have **(bt0)** and that $\mathfrak{m}_w$ satisfies **(qs0)**. Moreover, as $f_w$ is a p-morphism from $\langle \Delta, \leq \rangle$ onto $\langle T_w, \leq_w \rangle$, we also have **(bt1)**–**(bt3)** and **(qs1)**. So, by taking $\boldsymbol{q}(w) = \langle \boldsymbol{s}_w, \mathfrak{m}_w \rangle$ for each $w \in W$ we obtain a basic structure $\langle \mathfrak{F}, \boldsymbol{q} \rangle$ for $\varphi$ satisfying **(qm2)**.

It remains to define appropriate runs through $\langle \mathfrak{F}, \boldsymbol{q} \rangle$. For $k = 0, 1$, let $\mathfrak{R}^k$ be the set of all maps $r\colon w \mapsto f_w(x)$ for $x \in \Delta^k$. Clearly, $\mathfrak{R}^1$ and $\mathfrak{R}^0$ are sets of 1- and coherent and saturated 0-runs, respectively. Put $\mathfrak{R} = \mathfrak{R}^0 \cup \mathfrak{R}^1$ and for $r, r' \in \mathfrak{R}$, $r \lhd r'$ iff $r(w) \leq_w r'(w)$ for all $w \in W$. Then **(qm4)** holds by definition. Let $v \in W$ and $y \in T_v$. Then there is $x \in \Delta$ such that $f_v(x) = y$. Clearly, $\mathfrak{R}$ contains the run $r\colon w \mapsto f_w(x)$, which proves **(qm3)**. Finally, let $r \in \mathfrak{R}$, $v \in W$ and $y \in T_v^0$ be such that $r(v) \leq_v y$. There are some $z, x \in \Delta$ such that $f_w(z) = r(w)$, for every $w \in W$, and $f_v(x) = y$. We clearly have $z \leq x$ and $x \in \Delta^0$. Then take the run $r'\colon w \mapsto f_w(x)$. By definition, $r \lhd r'$, which proves **(qm5)**. Thus, $\mathfrak{Q} = \langle \mathfrak{F}, \boldsymbol{q}, \mathfrak{R}, \lhd \rangle$ is a quasimodel for $\varphi$. Note that if $\mathfrak{G}$ is finite then $\mathfrak{R}$ is finite as well and therefore, $\mathfrak{Q}$ is finitary. ❑

We are now in a position to establish the upper complexity bounds of the satisfiability problem for $\mathcal{PTL} \times \mathcal{RC}$- and $\mathcal{PTL} \circ \mathcal{RC}$-formulas in tt-models based on $\langle \mathbb{N}, < \rangle$, $\langle \mathbb{Z}, < \rangle$ or arbitrary finite flows of time.

**Proof of Theorem 3.10, upper bound.** We consider the cases of $\langle \mathbb{N}, < \rangle$ and $\langle \mathbb{Z}, < \rangle$. The case of arbitrary finite flows of time and that of tt-models with **FSA** and based on $\langle \mathbb{N}, < \rangle$ and $\langle \mathbb{Z}, < \rangle$ will follow from Theorem 3.13.

One can readily check that as for the propositional temporal logic $\mathcal{PTL}$, we have the following polynomial reductions for $\mathcal{PTL} \circ \mathcal{RC}$:

- satisfiability in tt-models based on $\langle \mathbb{Z}, < \rangle$ can be polynomially reduced to satisfiability in tt-models based on $\langle \mathbb{N}, < \rangle$;
- satisfiability in tt-models based on $\langle \mathbb{N}, < \rangle$ can be polynomially reduced to satisfiability of formulas without past-time temporal operators.

So, in what follows we consider the simplest case of the satisfiability problem, that is for $\mathcal{PTL} \circ \mathcal{RC}$-formulas without past-time temporal operators in tt-models based on $\langle \mathbb{N}, < \rangle$.

We present a nondeterministic 2EXPSPACE satisfiability checking algorithm which is similar to that of Sistla and Clarke (1985). First, one can prove (with the help of Lemmas C.1 (ii) and C.4) an analogue of (Hodkinson et al., 2000, Theorem 24) which states that a $\mathcal{PTL} \circ \mathcal{RC}$-formula $\varphi$ is satisfiable in tt-model based on $\langle \mathbb{N}, < \rangle$ iff there are $l_1, l_2 \in \mathbb{N}$ such that

$$l_1 \leq \sharp(\varphi), \qquad 0 \; < \; l_2 \; \leq \; |term\,\varphi| \cdot \left(2^{\flat(\varphi)}\right)^2 \cdot \sharp(\varphi) + \sharp(\varphi)$$

and a 'balloon'-like quasimodel $\mathfrak{Q} = \langle \langle \mathbb{N}, < \rangle, \boldsymbol{q}, \mathfrak{R}, \lhd \rangle$ for $\varphi$ with $\boldsymbol{q}(l_1 + n) = \boldsymbol{q}(l_1 + l_2 + n)$ for every $n \in \mathbb{N}$. Although Theorem 24 of (Hodkinson et al., 2000) was proved for the monodic fragment of first-order temporal logic, the basic idea of extracting a 'balloon'-like quasimodel from an arbitrary one works for $\mathcal{PTL} \circ \mathcal{RC}$ as well. The only difference is that quasistates now are more complex: they can be regarded as sets of sets of types for $\varphi$ (not just sets of types) and thus, both $l_1$ and $l_2$ are triple exponential in the length $\ell(\varphi)$ of $\varphi$. Then a quasimodel $\mathfrak{Q}$ can be guessed in 2EXPSPACE by an algorithm which is very similar to that in the proof of (Hodkinson et al., 2003, Theorem 4.1). ❑

**Proof of Theorem 3.13, upper bound.** The proof is similar to that of Theorem 3.10. Again, one can show that all the cases are polynomially reducible to the case of satisfiability of $\mathcal{PTL} \times \mathcal{RC}$-formulas without past-time temporal operators in tt-models with **FSA** and based on $\langle \mathbb{N}, < \rangle$. To take the **FSA** into account, we can prove (using Lemmas C.1 (i) and C.4) analogues of Theorems 29 and 35 of (Hodkinson et al., 2000) which state that a $\mathcal{PTL} \times \mathcal{RC}$-formula $\varphi$ is satisfiable in a tt-model with **FSA** and based on $\langle \mathbb{N}, < \rangle$ iff there is a *finitary* 'balloon'-like quasimodel for $\varphi$ based on $\langle \mathbb{N}, < \rangle$. The condition of finiteness for the set of runs can also be ensured by an algorithm similar to that of Theorem 3.10. ❑

### C.3 Upper Complexity Bounds (II): Embedding into First-Order Temporal Logic

In this appendix we introduce the *first-order temporal language* $\mathcal{QTL}$ and use some known complexity results for fragments of $\mathcal{QTL}$ to obtain upper complexity bounds for spatio-temporal logics based on $\mathcal{RC}^-$ (and therefore, on $\mathcal{BRCC}$-8).

The alphabet of $\mathcal{QTL}$ consists of individual variables $x_1, x_2, \dots$, predicate symbols $P_1, P_2, \dots$, each of which is of some fixed arity, the Booleans, the universal $\forall x$ and existential $\exists x$ quantifiers for each variable $x$, and the temporal operators $\mathcal{U}$, $\mathcal{S}$ (with their derivatives $\bigcirc$, $\Diamond_F$, $\Box_F$, etc.). Note that our language contains neither constant symbols nor equality (we simply do not need them to obtain our complexity results).

$\mathcal{QTL}$ is interpreted in *first-order temporal models* of the form $\mathfrak{M} = \langle \mathfrak{F}, D, I \rangle$, where $\mathfrak{F} = \langle W, < \rangle$ is a flow of time, $D$ a nonempty set, the *domain* of $\mathfrak{M}$, and $I$ a function associating with every moment of time $w \in W$ a first-order structure

$$I(w) = \left\langle D, P_0^{I(w)}, P_1^{I(w)}, \dots \right\rangle,$$

the *state* of $\mathfrak{M}$ at moment $w$, where each $P_i^{I(w)}$ is a relation on $D$ of the same arity as $P_i$. An *assignment* in $D$ is a function $\mathfrak{a}$ from the set of individual variables to $D$. Given such an assignment and a $\mathcal{QTL}$-formula $\varphi$, we define the *truth-relation* $(\mathfrak{M}, w) \models^{\mathfrak{a}} \varphi$ by taking

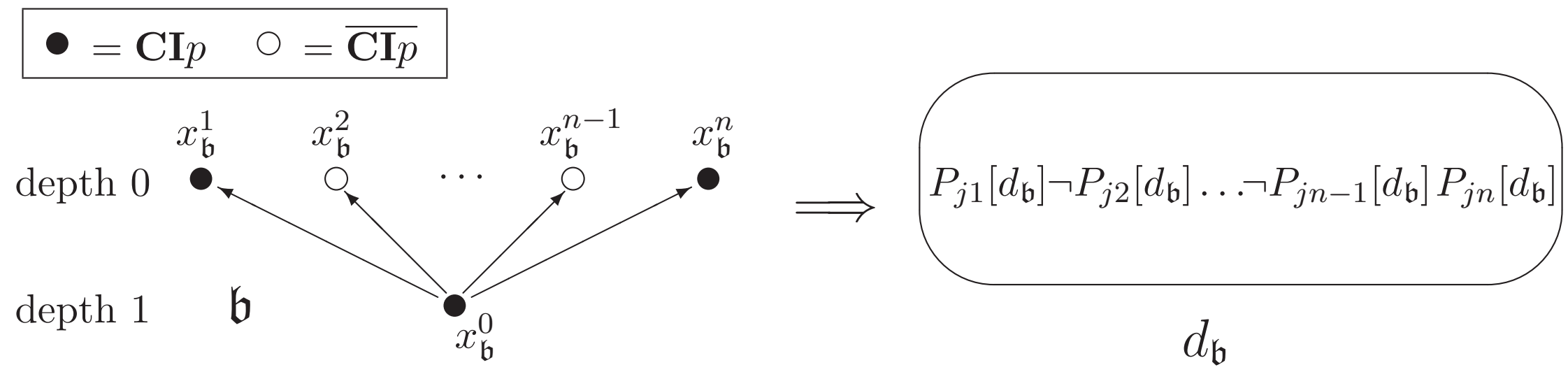

Figure 11: Representing $n$-broom $\mathfrak{b}$ with region $\mathbf{CI}p_j$ by a point in a first-order model.

- $(\mathfrak{M}, w) \models^{\mathfrak{a}} P_i(x_1, \dots, x_m)$ iff $\langle \mathfrak{a}(x_1), \dots, \mathfrak{a}(x_m)\rangle \in P_i^{I(w)}$,
- $(\mathfrak{M}, w) \models^{\mathfrak{a}} \forall x\, \psi$ iff $(\mathfrak{M}, w) \models^{\mathfrak{a}'} \psi$, for every assignment $\mathfrak{a}'$ in $D$ which differ from $\mathfrak{a}$ only on $x$, plus the standard clauses for the Booleans and temporal operators. We say that a $\mathcal{QTL}$-formula $\varphi$ is *satisfied* in $\mathfrak{M}$ if $(\mathfrak{M}, w) \models^{\mathfrak{a}} \varphi$ for some $w \in W$ and some assignment $\mathfrak{a}$ in $D$. If all free variables of $\varphi$ are among $x_1, \dots, x_m$, then instead of $(\mathfrak{M}, w) \models^{\mathfrak{a}} \varphi$ we often write $(\mathfrak{M}, w) \models \varphi[d_1, \dots, d_m]$, where $d_i = \mathfrak{a}(x_i)$ for all $i$, $1 \le i \le m$.

Denote by $\mathcal{QTL}^1$ the *one-variable fragment of* $\mathcal{QTL}$, i.e., the set of all $\mathcal{QTL}$-formulas which contain at most one individual variable, say, $x$. Without loss of generality we may assume that all predicate symbols of $\mathcal{QTL}^1$ are at most unary.

Now we define an embedding of spatio-temporal languages based on $\mathcal{RC}^-$ into $\mathcal{QTL}^1$. Recall that, by Lemma C.2 (i), if a $\mathcal{PTL} \times \mathcal{RC}^-$-formula $\varphi$ of width $n$ is satisfied in a tt-model with **FSA** then $\varphi$ is also satisfiable in an Aleksandrov tt-model based on the same flow of time and a finite *disjoint union of $n$-brooms*. Similarly, if a $\mathcal{PTL} \circ \mathcal{RC}^-$-formula $\varphi$ of width $n$ is satisfiable then $\varphi$ is also satisfiable in an Aleksandrov tt-model based on the same flow of time and possibly infinite *disjoint union of $n$-brooms*.

To cover both cases, let $\varphi$ be a $\mathcal{PTL} \times \mathcal{RC}^-$-formula of width $n$. We show how to construct a $\mathcal{QTL}^1$-formula $\varphi^{\dagger_n}$ of length linear in $\ell(\varphi)$ such that every Aleksandrov tt-model based on a (finite) union of $n$-brooms satisfying $\varphi$ gives rise to a first-order temporal model (with finite domain, respectively) satisfying $\varphi^{\dagger_n}$ and vice versa. Thus, we polynomially reduce the satisfiability problem for spatio-temporal languages to that for $\mathcal{QTL}^1$.

Suppose that $\varphi$ is satisfied in an Aleksandrov tt-model $\mathfrak{M} = \langle \mathfrak{F}, \mathfrak{G}, \mathfrak{V}\rangle$, where $\mathfrak{F} = \langle W, <\rangle$ and $\mathfrak{G}$ is a (finite or infinite) disjoint union of $n$-brooms. With every $n$-broom $\mathfrak{b}$ of $\mathfrak{G}$ we associate an element $d_{\mathfrak{b}}$ of the first-order domain $D$. Then, for every spatial variable $p_j$ in $\varphi$, we fix $n$ different unary predicate symbols $P_{j1}(x), \dots, P_{jn}(x)$ with the following meaning: $P_{ji}(x)$ is true on $d_{\mathfrak{b}} \in D$ at moment $w \in W$ iff the $i$-th leaf of $\mathfrak{b}$ ($x^i_{\mathfrak{b}}$ in Fig. 11) belongs to region $\mathbf{CI}p$ at $w$. Define $n$ distinct translations $\cdot^{\dagger^i_n}$, $1 \le i \le n$, encoding the truth values of spatio-temporal *region* terms of $\varphi$ on leaves of $\mathfrak{G}$ by taking, for a spatial variable $p_j$ and terms $\varrho_1$ and $\varrho_2$,

$$(\mathbf{CI}p_j)^{\dagger^i_n} = P_{ji}(x), \qquad (\mathbf{CI}\overline{\varrho_1})^{\dagger^i_n} = \neg(\varrho_1)^{\dagger^i_n}, \qquad (\mathbf{CI}(\varrho_1 \sqcap \varrho_2))^{\dagger^i_n} = (\varrho_1)^{\dagger^i_n} \wedge (\varrho_2)^{\dagger^i_n},$$
$$(\mathbf{CI}(\varrho_1 \,\mathcal{U}\, \varrho_2))^{\dagger^i_n} = (\varrho_1)^{\dagger^i_n} \,\mathcal{U}\, (\varrho_2)^{\dagger^i_n}, \qquad (\mathbf{CI}(\varrho_1 \,\mathcal{S}\, \varrho_2))^{\dagger^i_n} = (\varrho_1)^{\dagger^i_n} \,\mathcal{S}\, (\varrho_2)^{\dagger^i_n}.$$

Next we extend these $n$ translations to *arbitrary* spatio-temporal terms of $\varphi$. First we introduce a translation $\cdot^{\dagger_n^0}$ to encode the truth value of arbitrary spatio-temporal terms in the roots of the $n$-brooms of $\mathfrak{G}$: for a region term $\varrho$, let

$$(\varrho)^{\dagger_n^0} = \bigvee_{k=1}^{n} (\varrho)^{\dagger_n^k}.$$

The formula above shows, in particular, that $\cdot^{\dagger_n^0}$ is redundant for *region* terms since their truth values in the roots can be 'computed' as defined by $\cdot^{\dagger_n^0}$. For a spatio-temporal term of the form $\mathbf{I}\varrho$, where $\varrho$ is a region term, we take

$$(\mathbf{I}\varrho)^{\dagger_n^0} = \bigwedge_{k=1}^{n} (\varrho)^{\dagger_n^k} \qquad \text{and} \qquad (\mathbf{I}\varrho)^{\dagger_n^i} = (\varrho)^{\dagger_n^i} \quad \text{for all } i,\ 1 \le i \le n,$$

and then, for spatio-temporal terms $\tau_1$ and $\tau_2$,[10]

$$(\overline{\tau_1})^{\dagger_n^i} = \neg(\tau_1)^{\dagger_n^i} \qquad \text{and} \qquad (\tau_1 \sqcap \tau_2)^{\dagger_n^i} = (\tau_1)^{\dagger_n^i} \wedge (\tau_2)^{\dagger_n^i} \quad \text{for all } i,\ 0 \le i \le n.$$

Finally, we define the translation $\cdot^{\dagger_n}$ of subformulas of $\varphi$: for a spatio-temporal term $\tau$,

$$(\boxed{\forall}\tau)^{\dagger_n} = \forall x\, (\tau)^{\dagger_n^0} \quad \wedge \quad \bigwedge_{k=1}^{n} \forall x\, (\tau)^{\dagger_n^k}$$

and, for spatio-temporal formulas $\psi_1$ and $\psi_2$,

$$(\neg\psi_1)^{\dagger_n} = \neg\psi_1^{\dagger_n}, \qquad (\psi_1 \wedge \psi_2)^{\dagger_n} = \psi_1^{\dagger_n} \wedge \psi_2^{\dagger_n},$$
$$(\psi_1\, \mathcal{U}\, \psi_2)^{\dagger_n} = \psi_1^{\dagger_n}\, \mathcal{U}\, \psi_2^{\dagger_n}, \qquad (\psi_1\, \mathcal{S}\, \psi_2)^{\dagger_n} = \psi_1^{\dagger_n}\, \mathcal{S}\, \psi_2^{\dagger_n},$$

Clearly, the length of $\varphi^{\dagger_n}$ is linear in both $n$ and $\ell(\varphi)$.

**Lemma C.5.** *A $\mathcal{PTL}\times\mathcal{RC}^-$-formula $\varphi$ of width $n$ is satisfiable in an Aleksandrov tt-model based on a (finite) disjoint union of $n$-brooms iff $\varphi^{\dagger_n}$ is satisfiable in a first-order temporal model (with a finite domain) based on the same flow of time.*

**Proof.** ($\Rightarrow$) Suppose that $\varphi$ is satisfied in an Aleksandrov tt-model $\mathfrak{M} = \langle \mathfrak{F}, \mathfrak{G}, \mathfrak{V}\rangle$, where $\mathfrak{F} = \langle W, <\rangle$, $\mathfrak{G} = \langle V, R\rangle$ is a disjoint union of $n$-brooms $\mathfrak{b} = \langle W_\mathfrak{b}, R_\mathfrak{b}\rangle$, $W_\mathfrak{b} = \{x_\mathfrak{b}^0, x_\mathfrak{b}^1, \dots, x_\mathfrak{b}^n\}$ and $R_\mathfrak{b}$ is the reflexive closure of $\{\langle x_\mathfrak{b}^0, x_\mathfrak{b}^1\rangle, \dots, \langle x_\mathfrak{b}^0, x_\mathfrak{b}^n\rangle\}$ (see Fig. 11).

Construct a first-order temporal model $\mathfrak{N} = \langle \mathfrak{F}, D, I\rangle$ by taking $D$ to be the set of all $d_\mathfrak{b}$ for $n$-brooms $\mathfrak{b}$ in $\mathfrak{G}$ and, for every $w \in W$,

$$I(w) = \left\langle D, P_{11}^{I(w)}, \dots, P_{1n}^{I(w)}, P_{21}^{I(w)}, \dots, P_{2n}^{I(w)}, \dots \right\rangle,$$

where for each spatial variable $p_j$ in $\varphi$ and each $i$, $1 \le i \le n$,

$$P_{ji}^{I(w)} = \{d_\mathfrak{b} \in D \mid (\mathfrak{M}, \langle w, x_\mathfrak{b}^i\rangle) \models p_j\}.$$

10. For brevity, in this definition we follow the syntax of $\mathcal{PTL}\times\mathcal{RC}$ rather than $\mathcal{PTL}\times\mathcal{RC}^-$.

Note that $D$ is finite whenever $\mathfrak{G}$ is finite.

Now, by induction on the construction of a spatio-temporal *region* term $\varrho$ of $\varphi$, it can easily be shown that for every $w \in W$, every $n$-broom $\mathfrak{b}$ in $\mathfrak{G}$ and every $i$, $1 \leq i \leq n$,

$$(\mathfrak{N}, w) \models (\varrho)^{\dagger_n^i}[d_{\mathfrak{b}}] \quad \text{iff} \quad (\mathfrak{M}, \langle w, x_{\mathfrak{b}}^i \rangle) \models \varrho. \tag{58}$$

Next, (58) can be extended to *arbitrary* spatio-temporal terms $\tau$ of $\varphi$ and $i$, $0 \leq i \leq n$:

$$(\mathfrak{N}, w) \models (\tau)^{\dagger_n^i}[d_{\mathfrak{b}}] \quad \text{iff} \quad (\mathfrak{M}, \langle w, x_{\mathfrak{b}}^i \rangle) \models \tau. \tag{59}$$

The cases for $i$, $1 \leq i \leq n$, trivially follow from (58) and the fact that leaves have no successors but themselves. Consider now $i = 0$. The case $\tau = \varrho$ holds simply because region terms are interpreted by regular closed sets:

$$(\mathfrak{M}, \langle w, x_{\mathfrak{b}}^0 \rangle) \models \varrho \qquad \text{iff} \qquad (\mathfrak{M}, \langle w, x_{\mathfrak{b}}^k \rangle) \models \varrho, \quad \text{for some } k,\ 1 \leq k \leq n, \tag{60}$$

If $\tau = \mathbf{I}\varrho$ then, on one hand,

$$(\mathfrak{M}, \langle w, x_{\mathfrak{b}}^0 \rangle) \models \mathbf{I}\varrho \qquad \text{iff} \qquad (\mathfrak{M}, \langle w, x_{\mathfrak{b}}^k \rangle) \models \varrho, \quad \text{for all } k,\ 0 \leq k \leq n,$$

and on the other, by the definition of $\cdot^{\dagger_n^0}$,

$$(\mathfrak{N}, w) \models (\mathbf{I}\varrho)^{\dagger_n^0}[d_{\mathfrak{b}}] \qquad \text{iff} \qquad (\mathfrak{N}, w) \models (\varrho)^{\dagger_n^k}[d_{\mathfrak{b}}], \quad \text{for all } k,\ 1 \leq k \leq n,$$

which together with (60) and IH yields (59). The cases of the Booleans are trivial.

Finally, we show that for every $\psi \in \mathit{sub}\,\varphi$, we have

$$(\mathfrak{N}, w) \models \psi^{\dagger_n} \qquad \text{iff} \qquad (\mathfrak{M}, w) \models \psi.$$

Case $\psi = \boxtimes\tau$:

$$\begin{aligned} (\mathfrak{N}, w) \models (\boxtimes\tau)^{\dagger_n} \quad & \text{iff} \quad \forall d_{\mathfrak{b}} \in D\ \forall k \in \{0, 1, \dots, n\}\ (\mathfrak{N}, w) \models (\tau)^{\dagger_n^k}[d_{\mathfrak{b}}] \\ & \text{iff} \quad \forall \mathfrak{b} \text{ in } \mathfrak{G}\ \forall k \in \{0, 1, \dots, n\}\ (\mathfrak{M}, \langle w, x_{\mathfrak{b}}^k \rangle) \models \tau \\ & \text{iff} \quad (\mathfrak{M}, w) \models \boxtimes\tau. \end{aligned}$$

The remaining cases are trivial. It follows that $\varphi^{\dagger_n}$ is satisfied in $\mathfrak{N}$.

($\Leftarrow$) Assume that $\varphi^{\dagger}$ is satisfied in a first-order temporal model $\mathfrak{N} = \langle \mathfrak{F}, D, I \rangle$, where $\mathfrak{F} = \langle W, < \rangle$ and, for every $w \in W$,

$$I(w) = \left\langle D, P_{11}^{I(w)}, \dots, P_{1n}^{I(w)}, P_{21}^{I(w)}, \dots, P_{2n}^{I(w)}, \dots \right\rangle.$$

With every point $d \in D$ we associate an $n$-broom $\mathfrak{b}_d = \langle W_{\mathfrak{b}_d}, R_{\mathfrak{b}_d} \rangle$ so that the sets $W_{\mathfrak{b}_d}$, for $d \in D$, are pairwise disjoint and each contains $n + 1$ distinct elements $x_{\mathfrak{b}_d}^0, \dots, x_{\mathfrak{b}_d}^n$. Construct an Aleksandrov tt-model $\mathfrak{M} = \langle \mathfrak{F}, \mathfrak{G}, \mathfrak{V} \rangle$ by taking

- $\mathfrak{G}$ to be the disjoint union of $n$-brooms $\{\mathfrak{b}_d \mid d \in D\}$,
- $\mathfrak{V}(p_j, w) = \left\{ x_{\mathfrak{b}_d}^i \;\mid\; (\mathfrak{N}, w) \models (\mathbf{CI}p_j)^{\dagger_n^i}[d], \quad 0 \leq i \leq n \ \text{ and } \ d \in D \right\}$.

Clearly, if $D$ is finite then $\mathfrak{G}$ is finite as well.

By a straightforward induction one can show that for all $w \in W$, $d \in D$, spatio-temporal region terms $\varrho$, spatio-temporal terms $\tau$, subformulas $\psi$ of $\varphi$, and all $i$, $0 \leq i \leq n$,

$$\begin{array}{rcll} (\mathfrak{N}, w) \models (\varrho)^{\dagger_n^i}[d] & \text{iff} & (\mathfrak{M}, \langle w, x^i_{\mathfrak{b}_d} \rangle) \models \varrho & (i > 0), \\ (\mathfrak{N}, w) \models (\tau)^{\dagger_n^i}[d] & \text{iff} & (\mathfrak{M}, \langle w, x^i_{\mathfrak{b}_d} \rangle) \models \tau, & \\ (\mathfrak{N}, w) \models \psi^{\dagger_n} & \text{iff} & (\mathfrak{M}, w) \models \psi. & \end{array}$$

For example,

$$\begin{array}{rcl} (\mathfrak{N}, w) \models (\mathbf{I}\varrho)^{\dagger_n^0}[d] & \text{iff} & (\mathfrak{N}, w) \models (\varrho)^{\dagger_n^k}[d], \text{ for all } k,\ 0 \leq k \leq n \\ & \text{iff} & (\mathfrak{M}, \langle w, x^k_{\mathfrak{b}_d} \rangle) \models \varrho, \text{ for all } k,\ 0 \leq k \leq n \\ & \text{iff} & (\mathfrak{M}, \langle w, x^0_{\mathfrak{b}_d} \rangle) \models \mathbf{I}\varrho. \end{array}$$

It follows that $\varphi$ is satisfied in $\mathfrak{M}$. ❑

Now we obtain the upper complexity bounds for combinations of $\mathcal{PTL}$ and $\mathcal{RC}^-$:

**Proof of Theorem 3.11, upper bound.** Follows from Lemmas C.2 (ii) and C.5 together with the results on the complexity of the one-variable fragment of $\mathcal{QTL}$ (Halpern & Vardi, 1989; Sistla & German, 1987; Hodkinson et al., 2000, 2003). ❑

**Proof of Theorem 3.12, upper bound.** Similar to the proof above. ❑

**Proof of Theorem 3.15, upper bound.** The proof follows from Lemmas C.2 (i) and C.5 together with the upper complexity bound of the guarded monodic (and so the one-variable) fragment of $\mathcal{QTL}$ (Hodkinson, 2004). ❑

### C.4 Lower Complexity Bounds (II): Embedding First-Order Temporal Logic

We are now in a position to prove Theorem 3.14 and establish the lower complexity bounds for spatio-temporal logics based on $\mathcal{BRCC}$-8 (and so for those based on $\mathcal{RC}^-$ as well). Denote by $\mathcal{QTL}^1_\Box$ the one-variable fragment of $\mathcal{QTL}$ with sole temporal operator $\Box_F$. We define a polynomial embedding of $\mathcal{QTL}^1_\Box$ into $\mathcal{PTL}_\Box \times \mathcal{BRCC}$-8. Note that a similar embedding of the full one-variable fragment $\mathcal{QTL}^1$ into $\mathcal{PTL} \times \mathcal{BRCC}$-8 can be regarded as an alternative way to prove the lower complexity bound of Theorem 3.12.

A $\mathcal{QTL}^1_\Box$-formula is said to be a *basic Q-formula* if it is of the form $\forall x\, \vartheta(x)$, where $\vartheta(x)$ is quantifier-free and contains no propositional variables. A $\mathcal{QTL}^1_\Box$-sentence $\varphi$ is in *Q-normal form* if it is built from basic Q-formulas using the Booleans and temporal operator $\Box_F$. In other words, sentences in Q-normal form do not contain nested quantifiers and use only unary predicate symbols. The following observation should not come as a surprise (see, e.g., Hughes & Cresswell, 1996):

**Lemma C.6.** *For every $\mathcal{QTL}^1_\Box$-sentence $\varphi$ one can effectively construct a $\mathcal{QTL}^1_\Box$-sentence $\widehat{\varphi}$ in Q-normal form such that $\varphi$ is satisfiable in a first-order temporal model with a flow of time $\mathfrak{F}$ (and having finite domain) iff $\widehat{\varphi}$ is satisfiable in a first-order temporal model based on $\mathfrak{F}$ (and having finite domain). Moreover, the length of $\widehat{\varphi}$ is linear in the length of $\varphi$.*

**Proof.** Without loss of generality we may assume that $\varphi$ contains no occurrences of $\exists$. To transform $\varphi$ into its Q-normal form, we first introduce a fresh unary predicate symbol $P_i(x)$ for every propositional variable $p_i$ in $\varphi$ and replace each occurrence of $p_i$ with $\forall x\, P_i(x)$. Denote the resulting formula by $\varphi_0$. For every subformula $\psi$ of $\varphi_0$ define a formula $\psi^\sharp$ by taking inductively

$$(P(x))^\sharp = P(x), \qquad (\forall x\, \psi)^\sharp = P_{\forall x\psi}(x),$$
$$(\neg\psi)^\sharp = \neg\psi^\sharp, \qquad (\psi_1 \wedge \psi_2)^\sharp = \psi_1^\sharp \wedge \psi_2^\sharp, \qquad (\Box_F\psi)^\sharp = \Box_F\psi^\sharp,$$

where $P_{\forall x\psi}(x)$ is a fresh unary predicate symbol. Let

$$\widehat{\varphi} \;=\; \neg\forall x\,\neg\varphi_0^\sharp \;\wedge \bigwedge_{\forall x\psi \in sub\varphi_0} \Box_F^+\Big[\big(\forall x\, P_{\forall x\psi}(x) \vee \forall x\,\neg P_{\forall x\psi}(x)\big) \;\wedge\; \big(\forall x\, P_{\forall x\psi}(x) \leftrightarrow \forall x\, \psi^\sharp\big)\Big].$$

One can readily show by induction that $\widehat{\varphi}$ is satisfiable in a first-order temporal model based on $\mathfrak{F}$ (and having finite domain) iff $\varphi$ is satisfiable in a first-order temporal model based on $\mathfrak{F}$ (and having finite domain). Moreover, $\widehat{\varphi}$ is in Q-normal form. ❑

Now, given a $\mathcal{QTL}^1_\Box$-formula $\varphi$ in Q-normal form, denote by $\varphi^*$ the result of replacing all occurrences of basic Q-formulas $\forall x\,\vartheta(x)$ in $\varphi$ with $\mathsf{EQ}(\vartheta^*, \top)$, where $\top$ is a region term representing the whole space (for instance, $\mathbf{CI}u \sqcup \mathbf{CI}\overline{u}$ for a fresh spatial variable $u$), and the translation $\vartheta^*$ of quantifier-free formulas $\vartheta(x)$ is defined by taking:

$$(P(x))^* = \mathbf{CI}p, \quad (\neg\psi)^* = \mathbf{CI}\;\overline{\psi^*}\;, \quad (\psi_1 \wedge \psi_2)^* = \mathbf{CI}(\psi_1^* \sqcap \psi_2^*), \quad (\Box_F\psi)^* = \mathbf{CI}\Box_F\psi^*,$$

where $P(x)$ is a unary predicate symbol and $p$ a spatial variable standing for $P(x)$. Clearly, $\varphi^*$ belongs to $\mathcal{PTL}_\Box \times \mathcal{BRCC}$-8.

**Lemma C.7.** *A $\mathcal{QTL}^1_\Box$-sentence $\varphi$ in Q-normal form is satisfiable in a first-order temporal model based on a flow of time $\mathfrak{F}$ and having finite domain iff $\varphi^*$ is satisfiable in a tt-model based on $\mathfrak{F}$ and satisfying* **FSA***.*

**Proof.** ($\Rightarrow$) Suppose that $\varphi$ is in Q-normal form and $\mathfrak{M} = \langle \mathfrak{F}, D, I\rangle$ is a first-order temporal model, where $\mathfrak{F} = \langle W, <\rangle$ and, for all $w \in W$, $I(w) = \big\langle D, P_0^{I(w)}, \dots, \big\rangle$. Let $(\mathfrak{M}, w_0) \models \varphi$ for some $w_0 \in W$. Construct an Aleksandrov tt-model $\mathfrak{M}' = \langle \mathfrak{F}, \mathfrak{G}, \mathfrak{V}\rangle$ by taking $\mathfrak{G} = \langle D, R\rangle$, where $R = \{\langle d, d\rangle \mid d \in D\}$ and $\mathfrak{V}(p_i, w) = \big\{\langle w, d\rangle \mid d \in P_i^{I(w)}\big\}$. Note that the topological space $\mathfrak{T}_\mathfrak{G} = \langle D, \mathbb{I}_\mathfrak{G}\rangle$ induced by $\mathfrak{G}$ is *discrete*, i.e., for all $X \subseteq D$,

$$\mathbb{I}_\mathfrak{G} X = \mathbb{C}_\mathfrak{G} X = X.$$

It follows by induction that for every quantifier-free $\mathcal{QTL}^1_\Box$-formula $\vartheta$, every $w \in W$ and every $d \in D$ we have

$$(\mathfrak{M}, w) \models \vartheta[d] \qquad \text{iff} \qquad (\mathfrak{M}', \langle w, d\rangle) \models \vartheta^*.$$

Therefore, for every basic Q-formula $\forall x\,\vartheta(x)$ and every $w \in W$, $(\mathfrak{M}, w) \models \forall x\,\vartheta(x)$ iff $(\mathfrak{M}', w) \models \mathsf{EQ}(\vartheta^*, \top)$. It follows by induction that $(\mathfrak{M}', w_0) \models \varphi^*$.

($\Leftarrow$) Suppose that $\varphi^*$ is satisfied in a tt-model based on $\mathfrak{F} = \langle W, < \rangle$. By Lemma C.1 (i), $\varphi^*$ is satisfied in an Aleksandrov tt-model $\mathfrak{M} = \langle \mathfrak{F}, \mathfrak{G}, \mathfrak{V} \rangle$, where $\mathfrak{G} = \langle V, R \rangle$ is a disjoint union of brooms. Denote by $V_0 \subseteq V$ the set of leaves of $\mathfrak{G}$ and define a first-order temporal model $\mathfrak{M}' = \langle \mathfrak{F}, V_0, I \rangle$ by taking, for each $w \in W$,

$$I(w) = \langle V_0, P_0^{I(w)}, \dots \rangle \qquad \text{and} \qquad P_i^{I(w)} = \mathfrak{V}(p_i, w) \cap V_0.$$

Clearly, for every $X \subseteq V$, we have $\mathbb{I}_{\mathfrak{G}} X \cap V_0 = \mathbb{C}_{\mathfrak{G}} X \cap V_0 = X \cap V_0$, where $\mathfrak{T}_{\mathfrak{G}} = \langle V, \mathbb{I}_{\mathfrak{G}} \rangle$ is the topological space induced by $\mathfrak{G}$. So we obtain by induction that for every quantifier-free $\mathcal{QTL}^1_\square$-formula $\vartheta$, all $w \in W$ and all $d \in V_0$

$$(\mathfrak{M}', w) \models \vartheta[d] \qquad \text{iff} \qquad (\mathfrak{M}, \langle w, d \rangle) \models \vartheta^*.$$

A regular closed set $X \subseteq V$ in $\mathfrak{T}_{\mathfrak{G}}$ coincides with $V$ iff it contains $V_0$. So, for all basic Q-formulas $\forall x\, \vartheta(x)$ and all $w \in W$, $(\mathfrak{M}', w) \models \forall x\, \vartheta(x)$ iff $(\mathfrak{M}, w) \models \mathsf{EQ}(\vartheta^*, \top)$. It follows by induction that $\varphi$ is satisfied in $\mathfrak{M}'$. ❑

**Proof of Theorem 3.14, lower bound.** By Lemmas C.6 and C.7 the satisfiability problem for $\mathcal{QTL}^1_\square$-formulas in first-order temporal models with finite domains and based on $\langle \mathbb{N}, < \rangle$, $\langle \mathbb{Z}, < \rangle$ or arbitrary finite flows of time is polynomially reducible to satisfiability of $\mathcal{PTL}_\square \times \mathcal{BRCC}$-8 formulas in tt-models with **FSA**. Since the former is known to be EXPSPACE-hard (Hodkinson et al., 2003) for $\langle \mathbb{N}, < \rangle$ and $\langle \mathbb{Z}, < \rangle$, the latter is also EXPSPACE-hard in these cases. It should be noted that the result of Hodkinson and his colleagues (2003) can readily be extended to the case of arbitrary finite flows of time (by reduction of a finite version of the corridor tiling problem). This gives us the lower complexity bound for $\mathcal{PTL}_\square \times \mathcal{BRCC}$-8 in the case of finite flows of time. ❑

### C.5 PSPACE-complete Spatio-Temporal Logic

In this appendix we prove Theorem 3.8. In fact, we show that the satisfiability problem for $\mathcal{PTL} \circ \mathcal{RC}_2$—an extension of $\mathcal{PTL} \circ \mathcal{RCC}$-8—is decidable in PSPACE, where $\mathcal{RC}_2$ is the sublanguage of $\mathcal{S}4_u$ with spatial terms $\tau$ restricted to the following:

$$\begin{aligned}
\varrho \quad &::= \quad \mathbf{CI}p, \\
\sigma \quad &::= \quad \varrho \quad | \quad \overline{\mathbf{I}\varrho}, \\
\delta \quad &::= \quad \mathbf{I}\varrho \quad | \quad \overline{\varrho}, \\
\tau \quad &::= \quad \sigma_1 \sqcup \sigma_2 \quad | \quad \delta_1 \sqcup \delta_2 \quad | \quad \sigma \sqcup \delta.
\end{aligned}$$

As before, we denote by $\sigma$ spatial terms representing regular closed sets (regions) and by $\delta$ those representing regular open sets (the interiors of regions). Clearly, this definition is equivalent to the definition on p. 190 (where we did not make an explicit distinction between $\sigma$ and $\delta$). It is easy to see that $\mathcal{RC}_2$ contains $\mathcal{RCC}$-8, but is less expressive than $\mathcal{BRCC}$-8. Spatio-temporal terms $\tau$ of $\mathcal{PTL} \circ \mathcal{RC}_2$ are constructed from region terms of the form

$$\varrho \quad ::= \quad \mathbf{CI}p \quad | \quad \mathbf{CI}\bigcirc\varrho$$

in the same way as spatial terms of $\mathcal{RC}_2$. Finally, $\mathcal{PTL} \circ \mathcal{RC}_2$-formulas are composed from atomic formulas of the form $\boxtimes\tau$ using the Booleans and the temporal operators.

We will reduce the satisfiability problem for $\mathcal{PTL}\circ\mathcal{RC}_2$ to that for $\mathcal{PTL}$. This reduction will be done in a number steps.

Let $\mathfrak{F} = \langle W, < \rangle$ be a flow of time (as in the formulation of Theorem 3.8) and $\varphi$ a $\mathcal{PTL}\circ\mathcal{RC}_2$-formula. We begin by removing the next-time operator from the subterms of $\varphi$. To this end, let $\psi_0 = \varphi$ and for each variable $p$ from the set

$$\Omega_1 \quad = \quad \{p \mid \mathbf{CI}\bigcirc\mathbf{CI}p \in \mathit{term}\,\psi_0\},$$

we introduce a fresh spatial variable $p'$, and then put

$$\varphi_1 \quad = \quad \psi_1 \quad \wedge \quad \bigwedge_{p\in\Omega_1} \Box_P^+\Box_F^+\big(\bigcirc\top \to \boxtimes(\mathbf{CI}\bigcirc\mathbf{CI}p \equiv \mathbf{CI}p')\big),$$

where $\psi_1$ is the result of replacing each occurrence of $\mathbf{CI}\bigcirc\mathbf{CI}p$ in $\psi_0$ with $\mathbf{CI}p'$ and $\boxtimes(\varrho_1 \equiv \varrho_2)$ stands for $\boxtimes(\overline{\varrho_1} \sqcup \varrho_2) \wedge \boxtimes(\varrho_1 \sqcup \overline{\varrho_2})$. Next, for each $p$ from

$$\Omega_2 \quad = \quad \{p \mid \mathbf{CI}\bigcirc\mathbf{CI}p \in \mathit{term}\,\psi_1\},$$

we introduce a fresh spatial variable $p'$, and set

$$\varphi_2 \quad = \quad \psi_2 \quad \wedge \quad \bigwedge_{p\in\Omega_1\cup\Omega_2} \Box_P^+\Box_F^+\big(\bigcirc\top \to \boxtimes(\mathbf{CI}\bigcirc\mathbf{CI}p \equiv \mathbf{CI}p')\big),$$

where $\psi_2$ is the result of replacing each occurrence of $\mathbf{CI}\bigcirc\mathbf{CI}p$ in $\psi_1$ with $\mathbf{CI}p'$. By repeating this process sufficiently many times we can obtain a formula

$$\widetilde{\varphi} \quad = \quad \psi_\varphi \quad \wedge \quad \bigwedge_{p\in\Omega_\varphi} \Box_P^+\Box_F^+\big(\bigcirc\top \to \boxtimes(\mathbf{CI}\bigcirc\mathbf{CI}p \equiv \mathbf{CI}p')\big), \tag{61}$$

where $\Omega_\varphi$ is a suitable set of spatial variables, and $\psi_\varphi$ contains no $\bigcirc$ in region terms, that is, $\psi_\varphi$ is a $\mathcal{PTL}[\mathcal{RC}_2]$-formula. (Note that $\Omega_\varphi$ is such that if a spatial variable $p$ occurs in $\psi_\varphi$ then either $\mathbf{CI}\bigcirc\mathbf{CI}p \notin \mathit{term}\,\varphi$ or $p \in \Omega_\varphi$.) It should be clear that the length of $\widetilde{\varphi}$ is linear in the length of $\varphi$, and $\varphi$ is satisfiable in a tt-model based on $\mathfrak{F}$ iff $\widetilde{\varphi}$ is satisfiable in a tt-model based on $\mathfrak{F}$.

Thus, it suffices to reduce the satisfiability problem for $\mathcal{PTL} \circ \mathcal{RC}_2$-formulas of the form (61) to the satisfiability problem for $\mathcal{PTL}$-formulas. Let us now recall the function $\cdot^*$ from Appendix B.1 which maps $\mathcal{PTL}[\mathcal{S}4_u]$-formulas (in particular, $\mathcal{PTL}[\mathcal{RC}_2]$-formulas) to $\mathcal{PTL}$-formulas. Namely, for every atomic $\mathcal{RC}_2$-formula $\boxtimes\tau$, let $(\boxtimes\tau)^* = \mathsf{p}_\tau$, where $\mathsf{p}_\tau$ is a fresh propositional variable. Then, given the $\mathcal{PTL}[\mathcal{RC}_2]$-formula $\psi_\varphi$, define $\psi_\varphi^*$ to be the result of replacing every occurrence of $\boxtimes\tau$ in it with $(\boxtimes\tau)^*$. As is shown in the proof of Theorem 3.1, $\psi_\varphi$ is satisfiable in a tt-model over $\mathfrak{F} = \langle W, < \rangle$ iff

(s1) there exists a temporal model $\mathfrak{N} = \langle \mathfrak{F}, \mathfrak{U} \rangle$ satisfying $\psi_\varphi^*$ and,

(s2) for every $w \in W$, the set

$$\Phi_w = \{\boxtimes\tau \mid (\mathfrak{N}, w) \models \mathsf{p}_\tau,\ \tau \in \mathit{term}\,\psi_\varphi\} \cup \{\neg\boxtimes\tau \mid (\mathfrak{N}, w) \models \neg\mathsf{p}_\tau,\ \tau \in \mathit{term}\,\psi_\varphi\} \tag{62}$$

of $\mathcal{RC}_2$-formulas is satisfiable.

To preserve satisfiability of not only $\psi_\varphi$ but the whole $\widetilde{\varphi}$, we have to ensure somehow that

(s3) the points satisfying $\Phi_w$ do have predecessors and successors satisfying $\Phi_{w-1}$ and $\Phi_{w+1}$, respectively.

In the remainder of the appendix we first describe an encoding of the satisfiability problem for sets of $\mathcal{RC}_2$-formulas of the form (62) in Boolean logic, which will be used as part of our final reduction. Then we prove a *completion property* of $\mathcal{RC}_2$ (cf. Balbiani & Condotta, 2002) in the class of *exhaustive models* that contain 'sufficiently many' points of every type. Roughly, the completion property says that, given a set $\Phi$ of the form (62) and an exhaustive model satisfying some subset of $\Phi$, one can extend the valuation of the model to satisfy the whole $\Phi$. This property will make it possible to solve problem (s3) above. It is worth noting that a similar construction works for stronger languages such as $\mathcal{BRCC}$-8, but then, to enjoy the completion property, sets (62) may need exponentially many formulas (in the number of spatial variables) and, therefore, the reduction to $\mathcal{PTL}$ will be exponential as well. For $\mathcal{RC}_2$ it suffices to consider sets (62) with a quadratic number of formulas, which results in a quadratic reduction.

### C.5.1 Properties of $\mathcal{RC}_2$-formulas

For any finite set $\Omega = \{p_1, \dots, p_n\}$ of spatial variables, let

$$AtFm_\Omega \quad = \quad \big\{ \boxdot\tau \ \mid \ \tau \text{ is an } \mathcal{RC}_2\text{-term with variables from } \Omega \big\}.$$

Clearly, every $\mathcal{RC}_2$-formula with spatial variables from $\Omega$ is a Boolean combination of spatial formulas from $AtFm_\Omega$. It should be also clear that $|AtFm_\Omega| \leq 16 \cdot |\Omega|^2$.

As the width of $\mathcal{RC}_2$-formulas is $\leq 2$ (see p. 209 for the definition), by Lemmas A.1 and C.2 (ii), an $\mathcal{RC}_2$-formula is satisfiable iff it is satisfiable in an Aleksandrov topological model based on a disjoint union of 2-brooms, alias *forks*. In what follows we will regard every such model $\mathfrak{M}$ as a disjoint union of *fork models* $\mathfrak{m} = \langle \mathfrak{f}, \mathfrak{v} \rangle$, where $\mathfrak{f} = \langle W, R \rangle$, $W = \{x^0, x^1, x^2\}$, $R$ is the reflexive closure of $\{\langle x^0, x^1 \rangle, \langle x^0, x^2 \rangle\}$ and $\mathfrak{v}$ a valuation of the spatial variables. Given $\Omega_0 \subseteq \Omega$, we say that fork models $\mathfrak{m}_1 = \langle \mathfrak{f}, \mathfrak{v}_1 \rangle$ and $\mathfrak{m}_2 = \langle \mathfrak{f}, \mathfrak{v}_2 \rangle$ are $\Omega_0$-*equivalent* and write $\mathfrak{m}_1 \sim_{\Omega_0} \mathfrak{m}_2$, if $\mathfrak{v}_1(\mathbf{CI}p) = \mathfrak{v}_2(\mathbf{CI}p)$ for every $p \in \Omega_0$.

Given some $\Phi \subseteq AtFm_\Omega$ and $\psi \in AtFm_\Omega$, we say that $\psi$ is an $\mathfrak{f}$-*consequence* of $\Phi$ and write $\Phi \models_\mathfrak{f} \psi$ if $\mathfrak{m} \models \Phi$ implies $\mathfrak{m} \models \psi$ for every fork model $\mathfrak{m}$ based on $\mathfrak{f}$. $\Phi$ is said to be *closed* (under $\mathfrak{f}$-consequences) if, for every $\psi \in AtFm_\Omega$, we have $\psi \in \Phi$ whenever $\Phi \models_\mathfrak{f} \psi$. Let $\Phi^c = \big\{ \neg\boxdot\tau \mid \boxdot\tau \in AtFm_\Omega - \Phi \big\}$. Then $\Phi \cup \Phi^c$ is satisfiable iff $\Phi$ is closed and satisfiable. This means, in particular, that to check whether the set $\Phi_w$ in (62) is satisfiable, it is enough to consider only the closure of $\{\boxdot\tau \mid (\mathfrak{N}, w) \models \mathsf{p}_\tau,\ \tau \in term\psi\}$.

Now we characterise $\models_\mathfrak{f}$ in terms of the Boolean consequence relation $\models$. As we know from Appendix C.3, spatial formulas can be embedded into the one-variable fragment of first-order logic. More precisely, it can easily be shown that first-order translations of formulas from $AtFm_\Omega$ are (equivalent to) formulas of the form (which are actually *Krom formulas*; see, e.g., Börger, Grädel, & Gurevich, 1997):

$$(\boxdot(\sigma_1 \sqcup \sigma_2))^{\dagger_2} = \forall x \left( \sigma_1^{\dagger_2^1} \vee \sigma_2^{\dagger_2^1} \right) \ \wedge \ \forall x \left( \sigma_1^{\dagger_2^2} \vee \sigma_2^{\dagger_2^2} \right), \tag{63}$$

$$(\boxdot(\sigma_1 \sqcup \delta_2))^{\dagger_2} = \forall x \left( \sigma_1^{\dagger_2^1} \vee \delta_2^{\dagger_2^1} \right) \ \wedge \ \forall x \left( \sigma_1^{\dagger_2^2} \vee \delta_2^{\dagger_2^2} \right), \tag{64}$$

$$(\boxed{\forall}(\delta_1 \sqcup \delta_2))^{\dagger_2} = \forall x \left(\delta_1^{\dagger_2^1} \vee \delta_2^{\dagger_2^1}\right) \;\wedge\; \forall x \left(\delta_1^{\dagger_2^1} \vee \delta_2^{\dagger_2^2}\right) \;\wedge\; \forall x \left(\delta_1^{\dagger_2^2} \vee \delta_2^{\dagger_2^1}\right) \;\wedge\; \forall x \left(\delta_1^{\dagger_2^2} \vee \delta_2^{\dagger_2^2}\right), \quad (65)$$

where

$$\sigma^{\dagger_2^i} = \begin{cases} P_{ji}(x), & \text{if } \sigma = \mathbf{CI}p_j, \\ \neg P_{ji}(x), & \text{if } \sigma = \overline{\mathbf{ICI}p_j} \end{cases} \quad \text{and} \quad \delta^{\dagger_2^i} = \begin{cases} P_{ji}(x), & \text{if } \delta = \mathbf{ICI}p_j, \\ \neg P_{ji}(x), & \text{if } \delta = \overline{\mathbf{CI}p_j}, \end{cases} \quad \text{for} \quad i = 1, 2.$$

It follows from the proof of Lemma C.5 that an $\mathcal{RC}_2$-formula $\psi$ is satisfied in an Aleksandrov model $\mathfrak{M}$ based on a disjoint union of forks iff its first-order translation $\psi^{\dagger_2}$ is satisfied in a first-order model where every fork $\mathfrak{f} = \langle W, R\rangle$ of $\mathfrak{M}$, $W = \langle x^0, x^1, x^2\rangle$, $x^0Rx^1$ and $x^0Rx^2$, is encoded by a domain element $d_\mathfrak{f}$ with $P_{ji}(x)$ being true on $d_\mathfrak{f}$ iff $(\mathfrak{M}, x^i) \models \mathbf{CI}p_j$, for $i = 1, 2$ (see Fig. 11). Since in the definition of closed sets we only consider Aleksandrov models based on a single fork $\mathfrak{f}$, the domains of respective first-order models contain a single element $d_\mathfrak{f}$. This means that (63)–(65) can be encoded by the Boolean formulas

$$(\boxed{\forall}(\sigma_1 \sqcup \sigma_2))^{\ddagger} = \left(\sigma_1^{\ddagger^1} \vee \sigma_2^{\ddagger^1}\right) \;\wedge\; \left(\sigma_1^{\ddagger^2} \vee \sigma_2^{\ddagger^2}\right),$$
$$(\boxed{\forall}(\sigma_1 \sqcup \delta_2))^{\ddagger} = \left(\sigma_1^{\ddagger^1} \vee \delta_2^{\ddagger^1}\right) \;\wedge\; \left(\sigma_1^{\ddagger^2} \vee \delta_2^{\ddagger^2}\right),$$
$$(\boxed{\forall}(\delta_1 \sqcup \delta_2))^{\ddagger} = \left(\delta_1^{\ddagger^1} \vee \delta_2^{\ddagger^1}\right) \;\wedge\; \left(\delta_1^{\ddagger^1} \vee \delta_2^{\ddagger^2}\right) \;\wedge\; \left(\delta_1^{\ddagger^2} \vee \delta_2^{\ddagger^1}\right) \;\wedge\; \left(\delta_1^{\ddagger^2} \vee \delta_2^{\ddagger^2}\right),$$

where

$$\sigma^{\ddagger^i} = \begin{cases} q_{ji}, & \text{if } \sigma = \mathbf{CI}p_j, \\ \neg q_{ji}, & \text{if } \sigma = \overline{\mathbf{ICI}p_j} \end{cases} \quad \text{and} \quad \delta^{\ddagger^i} = \begin{cases} q_{ji}, & \text{if } \delta = \mathbf{ICI}p_j, \\ \neg q_{ji}, & \text{if } \delta = \overline{\mathbf{CI}p_j}, \end{cases} \quad \text{for} \quad i = 1, 2.$$

Thus, with every $\Phi \subseteq AtFm_\Omega$ we can associate the conjunction $\Phi^\ddagger$ of the $\cdot^\ddagger$-translations of formulas in $\Phi$ such that the following holds:

**Claim C.8.** *For every $\psi \in AtFm_\Omega$, $\Phi \models_\mathfrak{f} \psi$ iff $\Phi^\ddagger \models \psi^\ddagger$.*

To construct the closure of $\Phi \subseteq AtFm_\Omega$ and to check whether $\Phi$ is satisfiable, we can use the following resolution-like inference rules:

$$(\sigma\sigma)\;\; \frac{\boxed{\forall}(\sigma_1 \sqcup \varrho) \qquad \boxed{\forall}(\overline{\mathbf{I}\varrho} \sqcup \sigma_2)}{\boxed{\forall}(\sigma_1 \sqcup \sigma_2)} \qquad (\sigma\delta)_1\;\; \frac{\boxed{\forall}(\mathbf{I}\varrho \sqcup \delta_1)}{\boxed{\forall}(\varrho \sqcup \delta_1)} \qquad (\sigma\delta)_2\;\; \frac{\boxed{\forall}(\overline{\varrho} \sqcup \delta_1)}{\boxed{\forall}(\overline{\mathbf{I}\varrho} \sqcup \delta_1)}$$

$$(\bot)\;\; \frac{\boxed{\forall}\varrho \qquad \boxed{\forall}\overline{\varrho}}{\bot} \qquad (\delta\delta)\;\; \frac{\boxed{\forall}(\delta_1 \sqcup \theta) \qquad \boxed{\forall}(\theta' \sqcup \delta_2)}{\boxed{\forall}(\delta_1 \sqcup \delta_2)} \quad \text{for} \quad \begin{cases} \theta = \overline{\varrho}, & \theta' = \mathbf{I}\varrho; \\ \theta = \varrho, & \theta' = \overline{\varrho}; \\ \theta = \overline{\mathbf{I}\varrho}, & \theta' = \mathbf{I}\varrho; \end{cases}$$

together with the equivalences:

$$\boxed{\forall}\varrho = \boxed{\forall}\mathbf{I}\varrho, \qquad \boxed{\forall}\overline{\varrho} = \boxed{\forall}\overline{\mathbf{I}\varrho}, \qquad \boxed{\forall}(\varrho \sqcup \sigma_1) = \boxed{\forall}(\mathbf{I}\varrho \sqcup \sigma_1), \qquad \boxed{\forall}(\overline{\varrho} \sqcup \sigma_1) = \boxed{\forall}(\overline{\mathbf{I}\varrho} \sqcup \sigma_1),$$

where $\varrho = \mathbf{CI}p$ for some $p \in \Omega$, $\sigma_1$ and $\sigma_2$ are of the form $\varrho$ or $\overline{\mathbf{I}\varrho}$, and $\delta_1$ and $\delta_2$ of the form $\mathbf{I}\varrho$ or $\overline{\varrho}$. It is readily checked that the above rules are sound, and so if $\bot$ is derivable from $\Phi$, then $\Phi$ is not satisfiable. On the other hand, if $\Phi$ is not satisfiable then $\Phi^\ddagger$ can be regarded as an unsatisfiable set of binary and unary propositional clauses and, using the standard resolution procedure, one can construct a derivation of the empty clause from $\Phi^{\ddagger}$—which, in turn, can be mimicked by applications of the above rules (and equivalences) to derive $\bot$ from $\Phi$. Moreover, since the propositional resolution is subsumption complete (see, e.g., Slagle, Chang, & Lee, 1969), we can also derive all consequences of $\Phi$, thereby obtaining its closure.

Now we encode the above rules and equivalences as Boolean formulas with variables $\mathsf{p}_\tau$, for $\boxtimes\tau \in AtFm_\Omega$. For instance, $(\bot)$ and $(\sigma\sigma)$ are encoded by

$$\big(\boxtimes\varrho \wedge \boxtimes\overline{\varrho}\big)^* \to \bot \qquad \text{and} \qquad \big(\boxtimes(\sigma_1 \sqcup \varrho) \wedge \boxtimes(\overline{\mathbf{I}\varrho} \sqcup \sigma_2) \to \boxtimes(\sigma_1 \sqcup \sigma_2)\big)^*,$$

respectively. Denote by $\Gamma_\Omega$ the conjunction of all such formulas for spatial variables from $\Omega$. Then we have the following:

**Claim C.9.** *For every $\Phi \subseteq AtFm_\Omega$, $\Phi$ is closed and satisfiable iff the Boolean formula*

$$\Gamma_\Omega \quad \wedge \quad \Big[\bigwedge_{\boxtimes\tau\in\Phi} \mathsf{p}_\tau \quad \wedge \quad \bigwedge_{\boxtimes\tau\in AtFm_\Omega - \Phi} \neg\mathsf{p}_\tau\Big] \tag{66}$$

*is satisfiable.*

Finally, to ensure (s3), we need the following *completion property* of $\mathcal{RC}_2$:

**Lemma C.10.** *Let $\Phi$ be a closed subset of $AtFm_\Omega$, $\Omega_0 \subseteq \Omega$ and $\Phi_0 = \Phi \cap AtFm_{\Omega_0}$. Then* (i) *$\Phi_0$ is closed and* (ii) *for every fork model $\mathfrak{m}_0$, if $\mathfrak{m}_0 \models \Phi_0$ then there is a fork model $\mathfrak{m}$ such that $\mathfrak{m}_0 \sim_{\Omega_0} \mathfrak{m}$ and $\mathfrak{m} \models \Phi$.*

**Proof.** Claim (i) is clear. To show (ii), we define the characteristic formula $\chi$ of $\mathfrak{m}_0$ on $\Omega_0$ by taking:

$$\chi \quad = \bigwedge_{p_j\in\Omega_0,\ i=1,2} l^*_{ji} \qquad \text{and} \qquad l^*_{ji} = \begin{cases} \neg q_{ji}, & \text{if } (\mathfrak{m}_0, x^i) \not\models \mathbf{CI}p_j, \\ q_{ji}, & \text{if } (\mathfrak{m}_0, x^i) \models \mathbf{CI}p_j. \end{cases}$$

If $\mathfrak{m}_0 \models \Phi_0$ then it follows immediately from the definitions that $\Phi_0^{\ddagger} \wedge \chi$ is satisfiable. Our aim is to show that $\Phi^{\ddagger} \wedge \chi$ is also satisfiable, which would mean that there is a fork model $\mathfrak{m}$ as required. Suppose otherwise. Then $\Phi^{\ddagger} \models \neg\chi$. We can regard $\Phi^{\ddagger}$ as a set of unary and binary clauses and $\neg\chi$ as a clause with $2\cdot|\Omega_0|$ literals $l_{ji}$, the negations of the $l^*_{ji}$. According to the subsumption theorem (Slagle et al., 1969), by applying the standard resolution rule to $\Phi^{\ddagger}$, we can derive a clause $l_{j_1i_1} \vee l_{j_2i_2}$ which subsumes $\neg\chi$ (i.e., its both literals occur in $\neg\chi$). Since $\Phi$ is closed, we have $l_{j_1i_1} \vee l_{j_2i_2}$ among the clauses of $\Phi^{\ddagger}$ and as the $l_{j_ki_k}$ are the $\cdot^{\ddagger^{i_k}}$-translations of spatial terms for spatial variables from $\Omega_0$, we conclude that $l_{j_1i_1} \vee l_{j_2i_2}$ is indeed among the clauses of $\Phi_0^{\ddagger}$, contrary to $\Phi_0^{\ddagger} \wedge \chi$ being satisfiable. ❑

### C.5.2 The Polynomial Translation of $\mathcal{PTL} \circ \mathcal{RC}_2$ into $\mathcal{PTL}$

Now we are in a position to define a polynomial (at most quadratic) translation $\cdot^\bullet$ of $\mathcal{PTL} \circ \mathcal{RC}_2$ into $\mathcal{PTL}$. Starting with a given formula $\varphi$, we construct the $\mathcal{PTL} \circ \mathcal{RC}_2$-formula $\widetilde{\varphi}$ of the form (61):

$$\widetilde{\varphi} \quad = \quad \psi_\varphi \quad \wedge \bigwedge_{p\in\Omega_\varphi} \Box_P^+\Box_F^+\big(\bigcirc\top \to \boxtimes(\mathbf{CI}\bigcirc\mathbf{CI}p \equiv \mathbf{CI}p')\big),$$

where $\psi_\varphi$ is a $\mathcal{PTL}[\mathcal{RC}_2]$-formula. Let $\Omega'_\varphi = \{p' \mid p \in \Omega_\varphi\}$ and let $\Omega$ denote the smallest set of spatial variables containing $\Omega_\varphi \cup \Omega'_\varphi$ and all spatial variables occurring in $\psi_\varphi$. Given $\boxtimes\tau \in AtFm_{\Omega_\varphi}$, denote by $\boxtimes\tau'$ the formula from $AtFm_{\Omega'_\varphi}$ obtained from $\boxtimes\tau$ by replacing every occurrence of $p \in \Omega_\varphi$ with $p' \in \Omega'_\varphi$. Consider the $\mathcal{PTL}$-formula

$$\varphi^\bullet \quad = \quad \psi^*_\varphi \quad \wedge \quad \Box^+_P\Box^+_F\,\Gamma_\Omega \quad \wedge \quad \Box^+_P\Box^+_F\Big(\bigcirc\top \to \bigwedge_{\boxtimes\tau \in AtFm_{\Omega_\varphi}} (\bigcirc\boxtimes\tau \leftrightarrow \boxtimes\tau')^*\Big).$$

**Lemma C.11.** *For every $\mathcal{PTL} \circ \mathcal{RC}_2$-formula $\varphi$, $\widetilde{\varphi}$ is satisfiable in a tt-model based on $\mathfrak{F} = \langle W, < \rangle$ iff $\varphi^\bullet$ is satisfiable in a temporal model based on $\mathfrak{F}$.*

**Proof.** ($\Rightarrow$) Let $(\mathfrak{M}, w_0) \models \widetilde{\varphi}$. Construct a temporal model $\mathfrak{N} = \langle \mathfrak{F}, \mathfrak{V} \rangle$ by taking, for $\boxtimes\tau \in AtFm_\Omega$,

$$\mathfrak{V}(\mathsf{p}_\tau) = \{w \in W \mid (\mathfrak{M}, w) \models \boxtimes\tau\}.$$

It is easy to see that $(\mathfrak{N}, w_0) \models \varphi^\bullet$.

($\Leftarrow$) Let $(\mathfrak{N}, w_0) \models \varphi^\bullet$ for some $w_0 \in W$. For every $w \in W$, set

$$\Phi_w = \{\boxtimes\tau \in AtFm_\Omega \mid (\mathfrak{N}, w) \models \mathsf{p}_\tau\}.$$

Let $\Lambda_w$, for $w \in W$, be a set of *all* non-$\Omega$-equivalent fork models $\mathfrak{m}$ with $\mathfrak{m} \models \Phi_w$. By Claim C.9, the $\Phi_w$ are closed and satisfiable, so the sets $\Lambda_w$ are nonempty. We use the elements of the $\Lambda_w$ as building blocks for exhaustive states in the tt-model we are going to construct in order to satisfy $\varphi$.

First we show that each element of $\Lambda_w$ has a successor in $\Lambda_{w+1}$ and a predecessor in $\Lambda_{w-1}$ (provided that $w$ has a successor and predecessor, respectively). More precisely, we say that a pair of fork models $\mathfrak{m} = \langle \mathfrak{f}, \mathfrak{v} \rangle$ and $\mathfrak{m}' = \langle \mathfrak{f}, \mathfrak{v}' \rangle$ is *suitable* and write $\mathfrak{m} \to \mathfrak{m}'$ if $\mathfrak{v}(\mathbf{CI}p') = \mathfrak{v}'(\mathbf{CI}p)$, for every $p \in \Omega_\varphi$.

(*succ*) Let $\mathfrak{m} \in \Lambda_w$, $\mathfrak{m} = \langle \mathfrak{f}, \mathfrak{v} \rangle$, and let $w \in W$ have a successor $w+1$. By the third conjunct of $\varphi^\bullet$, we have

$$\begin{aligned} \Phi_w \cap AtFm_{\Omega'_\varphi} &= \{\boxtimes\tau' \in AtFm_{\Omega'_\varphi} \mid (\mathfrak{N}, w) \models \mathsf{p}_{\tau'}\} \\ &= \{\boxtimes\tau' \in AtFm_{\Omega'_\varphi} \mid (\mathfrak{N}, w+1) \models \mathsf{p}_\tau\}. \end{aligned}$$

Therefore,

$$\Phi_{w+1} \cap AtFm_{\Omega_\varphi} \quad = \quad \{\boxtimes\tau \in AtFm_{\Omega_\varphi} \mid (\mathfrak{N}, w) \models \mathsf{p}_{\tau'}\}.$$

Now, by $\mathfrak{m} \in \Lambda_w$, we have $\mathfrak{m} \models \Phi_w \cap AtFm_{\Omega'_\varphi}$. So if we define a fork model $\mathfrak{m}' = \langle \mathfrak{f}, \mathfrak{v}' \rangle$ by taking $\mathfrak{v}'(p) = \mathfrak{v}(p')$, for all $p \in \Omega_\varphi$ (and arbitrary otherwise), then $\mathfrak{m}' \models \Phi_{w+1} \cap AtFm_{\Omega_\varphi}$ follows. Since $\Phi_{w+1}$ is closed, by Lemma C.10, we can find a fork model $\mathfrak{m}'' = \langle \mathfrak{f}, \mathfrak{v}'' \rangle$ such that $\mathfrak{m}'' \sim_{\Omega_\varphi} \mathfrak{m}'$ and $\mathfrak{m}'' \models \Phi_{w+1}$. It follows that $\mathfrak{m} \to \mathfrak{m}''$ and $\mathfrak{m}''$ is $\Omega$-equivalent to some fork model in $\Lambda_{w+1}$ (i.e., we may assume that $\mathfrak{m}'' \in \Lambda_{w+1}$).

(*pred*) Similarly, for every $\mathfrak{m} \in \Lambda_m$, $\mathfrak{m} = \langle \mathfrak{f}, \mathfrak{v} \rangle$, and every $w \in W$ with a predecessor $w-1$, there is $\mathfrak{m}'' = \langle \mathfrak{f}, \mathfrak{v}'' \rangle$ such that $\mathfrak{m}'' \in \Lambda_{w-1}$ and $\mathfrak{m}'' \to \mathfrak{m}$.

It should be clear that for every fork model $\mathfrak{m} \in \Lambda_w$ and every $w \in W$, we can define a function $r_{\mathfrak{m},w}$ that gives for *each* $u \in W$ a fork model $r_{\mathfrak{m},w}(u) \in \Lambda_u$ such that $r_{\mathfrak{m},w}(w) = \mathfrak{m}$ and $r_{\mathfrak{m},w}(u) \to r_{\mathfrak{m},w}(u+1)$, whenever $u+1$ is a successor of $u$. Let $\Delta$ be the set of all such functions $r_{\mathfrak{m},w}$, for $w \in W$ and $\mathfrak{m} \in \Lambda_w$.

We are now ready to define an Aleksandrov tt-model $\mathfrak{M} = \langle \mathfrak{F}, \mathfrak{G}, \mathfrak{V} \rangle$ satisfying $\widetilde{\varphi}$. Let $\mathfrak{G} = \langle W, R \rangle$ be a disjoint union of $|\Delta|$-many forks $\mathfrak{f}_r = \langle W_r, R_r \rangle$, $W_r = \{x_r^0, x_r^1, x_r^2\}$, $x_r^0 R_r x_r^1$ and $x_r^0 R_r x_r^2$, for each $r \in \Delta$, and let $\mathfrak{V}(p, w) = \{x_r^i \in W \mid (r(w), x_r^i) \models \mathbf{CI}p\}$, for all $p \in \Omega$ and $w \in W$. We show by induction on the construction of $\chi \in \mathit{sub}\,\psi_\varphi$ that, for every $w \in W$,

$$(\mathfrak{M}, w) \models \chi \qquad \text{iff} \qquad (\mathfrak{N}, w) \models \chi^*.$$

*Case* $\psi = \boxdot\tau$. Suppose that $(\mathfrak{M}, w) \models \boxdot\tau$ but $(\mathfrak{N}, w) \not\models \mathsf{p}_\tau$. Then $\boxdot\tau \notin \Phi_w$ and, since $\Phi_w$ is closed (by Claim C.9 and $\Gamma_\Omega$ being true at $w$), we have $\Phi_w \not\models_{\mathfrak{f}} \boxdot\tau$. It follows that there is a fork model $\mathfrak{m} \in \Lambda_w$ with $\mathfrak{m} \models \neg\boxdot\tau$, and so there is $r \in \Delta$ such that $r(w) = \mathfrak{m}$, contrary to $(\mathfrak{M}, w) \models \boxdot\tau$. Conversely, if $(\mathfrak{N}, w) \models \mathsf{p}_\tau$ then, by construction, $(\mathfrak{M}, w) \models \boxdot\tau$.

The cases of the Booleans and temporal operators are trivial.

As the second conjunct of $\widetilde{\varphi}$ is satisfied by construction, we obtain $(\mathfrak{M}, w_0) \models \widetilde{\varphi}$. ❑